\documentclass[10pt,twocolumn,letterpaper]{article}
\usepackage{iccv}
\usepackage{times}
\pdfoutput=1
\usepackage[ruled]{algorithm2e}

\usepackage{amsmath}
\usepackage{amssymb}
\usepackage{color}
\usepackage{graphicx}
\usepackage{tabularx}
\usepackage{subfigure}
\graphicspath{{./figures_flowchart/}{./figures_flowers/}{./figures_pascal/}}
\usepackage[pagebackref=true,breaklinks=true,letterpaper=true,colorlinks,bookmarks=false]{hyperref}

\iccvfinalcopy 

\def\iccvPaperID{1923} 

\ificcvfinal\fi
\begin{document}

\title{Deep Attributes from Context-Aware Regional Neural Codes}

\author{Jianwei Luo$^{1}${\thanks{This work was done when the first author was working as intern at Intel Labs China under the supervision of the second author.}},
 ~Jianguo Li$^{2}$, Jun Wang$^{3}$, Zhiguo Jiang$^{1}$, Yurong Chen$^{2}$\\
$^{1}$ Image Processing Center, Beihang University\\
$^{2}$ Intel Labs China\\
$^{3}$ Columbia University
}

\maketitle

\begin{abstract}
Recently, many researches employ middle-layer output of convolutional neural network models (CNN) as features for different visual recognition tasks. Although promising results have been achieved in some empirical studies, such type of representations still suffer from the well-known issue of semantic gap.
This paper proposes so-called deep attribute framework to alleviate this issue from three aspects.
First, we introduce object region proposals as intermedia to represent target images, and extract features from region proposals. Second, we study aggregating features from different CNN layers for all region proposals.
The aggregation yields a holistic yet compact representation of input images. Results show that cross-region max-pooling of soft-max layer output outperform all other layers. As soft-max layer directly corresponds to semantic concepts, this representation is named ``deep attributes".
Third, we observe that only a small portion of generated regions by object proposals algorithm are correlated to classification target.
Therefore, we introduce context-aware region refining algorithm to pick out contextual regions and build context-aware classifiers.

We apply the proposed deep attributes framework for various vision tasks.
Extensive experiments are conducted on standard benchmarks for three visual recognition tasks, i.e., image classification, fine-grained recognition and visual instance retrieval. Results show that deep attribute approaches achieve state-of-the-art results, and outperforms existing peer methods with a significant margin, even though some benchmarks have little overlap of concepts with the pre-trained CNN models.
\end{abstract}
\section{Introduction}

\begin{figure}[th]
\small
\centering
\includegraphics[width=3.2in, height=0.8in]{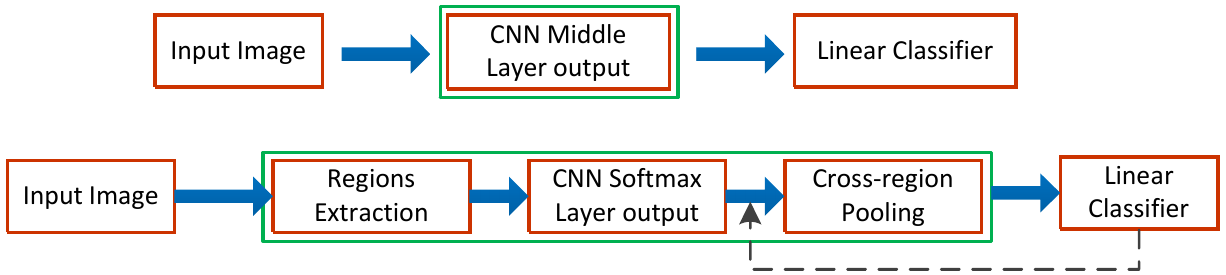}
\caption{Schematic comparison of traditional method (top row) with the proposed method (bottom row) on how to use CNN models on visual recognition tasks. Instead of representing images by neural codes from middle-layers, we adopt the semantic neural codes in combination with region proposals. Semantic neural codes from different regions are aggregating with cross-region-pooling. Classifier may feedback for context-aware region refining (the dot-line).}
\label{flowchart1:compare}
\vspace{-0.11in}
\end{figure}

Since the breakthrough work of Krizhevsky et al.~\cite{krizhevsky2012imagenet} on ImageNet \cite{deng2009imagenet}, researches on convolutional neural networks (CNN) have been exploding. Among them, a lot of researches adopt pre-trained CNN models as feature extractor for various visual recognition tasks like object detection \cite{girshick2013rich}, object recognition \cite{DeCaf,razavian2014cnn,Chatfield14}, image retrieval \cite{gong2014multi,razavian2014cnn}, etc. Features are usually from different CNN layer activations or outputs.
To achieve advanced and robust performance, people either fine-tune the pre-trained CNN models on their own tasks, or make extensively data augmentation to get robust classifiers. These developed techniques have shown promising results in comparison to conventional methods using standard feature representations like bag-of-words \cite{sivic2003video}, sparse-coding \cite{yang2009linear}, etc. However, there are two limitations of these kind of methods.
First, neural codes from middle-layer are difficult to explain. Second, neural code extraction from whole image definitely loss many context information.
These two are summarized to be the well-known {\em semantic gap}~\cite{Smeulders:2000PAMI}.
\begin{figure*}[t]
\small
\centering
\includegraphics[width=6.0in, height=2.3in]{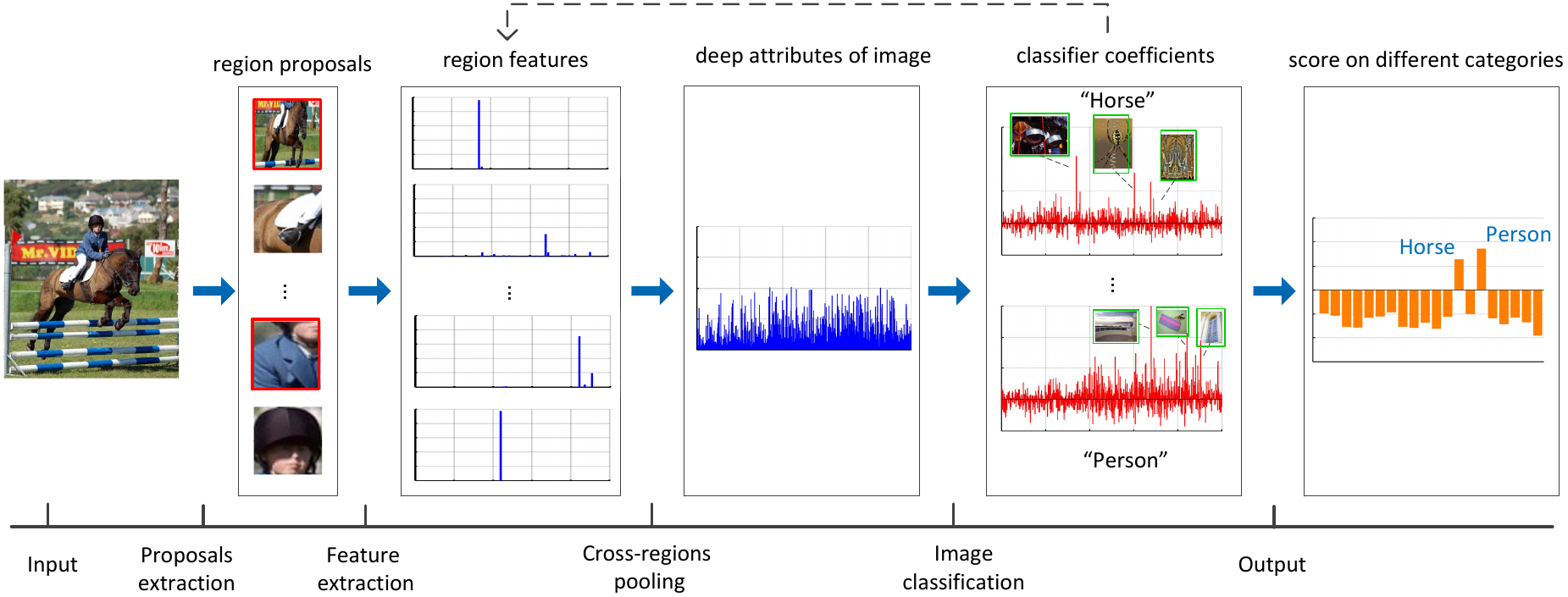}
\caption{Flowchart of deep attributes. The CNN models are pre-trained from 1000 categories ILSVRC.
Cross-region-pooling is performed over all the proposals within the image. Note that features in each region is sparse, after cross-region-pooling, the representation just has a portion of large elements, which corresponds to context categories.
The learned classifiers have relative large value on a few weight coefficients. This inspires us to use classifier feedback to pick context regions.}
\label{flowchart2:da}
\vspace{-0.12in}
\end{figure*}

Meanwhile, region features are appealing on recognition task. For instance, Gu \etal~\cite{gu2009recognition} employ mid-level features like contour shape, edge shape, color and texture to describe each region for visual recognition tasks. It is well known that region features can naturally preserve more mid-level semantic information like materials, textures, shapes, etc of objects~\cite{farhadi2009describing}.
However, traditional region representations either highly depend on segmentation algorithm~\cite{arbelaez2009contours,li2014attributes}, or lack of a generic semantic representation for regions for various visual recognition tasks.

To address these challenges and leverage the power from both richness and semantics of these two types of feature representations, in this paper, we propose to integrate the semantic output, i.e., the output from the soft-max layer of CNN models, as well as region proposals to achieve compact yet effective visual representations, namely {\em deep attribute} (DA). Since the soft-max layer neural codes are the probability response to the categories on which CNNs are trained, it is fairly compact, semantic, and sparse due to insignificant responses to most categories. Briefly, the proposed method contains four key components.
\begin{itemize}
\vspace{-0.04in}
\addtolength{\itemsep}{-0.05in}
\small
\item[(1)] We introduce region proposals using algorithm like selective search~\cite{uijlings2013selective} or edge-box~\cite{edgebox} from each input image.
\item[(2)] We feed each extracted regions to CNN models to compute neural codes from soft-max layer.
\item[(3)] We perform cross-region-pooling of regional neural codes to obtain a holistic yet compact representation of the image.
\item[(4)] We observe that only a small portion of region proposals are correlated to classification target, thus we impose learned classifier feedback to pick out a few contextual regions and re-pooling to get a context-aware classifier.
\vspace{-0.04in}
\end{itemize}
The major difference to traditional off-the-shelf CNN methods are illustrated in Figure~\ref{flowchart1:compare}, while the proposed approach is further illustrated in details in Figure~\ref{flowchart2:da}. To capture more context information related to scale, we also study different pooling layout schemes like multi-scale-pooling and spatial pyramid-pooling extensions.

To demonstrate the capacity and robustness of the proposed technique, we employ deep attributes on three visual recognition tasks: image classification, fine-grained object recognition, and visual instance retrieval. Experimental results on several standard benchmark datasets show that deep attributes can achieve state-of-the-arts performance. Especially, the context-aware region refining (CARR) algorithm clearly outperforms peer methods with a significant margin. In summary, the major contributions of this paper include:
\begin{itemize}
\vspace{-0.04in}
\addtolength{\itemsep}{-0.05in}
\small
  \item We propose a {\em deep attribute} approach for visual recognition tasks, which is equipped with the semantic power from the outputs of pre-trained CNN models, as
        well as the compactness of region proposals.
  \item We present schemes to utilize contextual information among region proposals.
        The context-aware region refining algorithm yield large performance gains.
  \item Experimental results on three different visual recognition tasks clearly show the superiority of the proposed method, as well as the generalization ability to different vision tasks.
\vspace{-0.04in}
\end{itemize}

In the rest of the paper, we will first give a brief survey on related works in Section 2, and then present the details of the proposed deep attribute approach in Section 3.
In Section 4, we conduct experiments to study different aspects of setting which will impact the performance of the algorithm.
We show experimental results on three visual recognition tasks on Section 5.
Conclusions and discussions are given in Section 6.

\section{Related Works}\label{sec:related_work}
In this section, we will briefly revisit related works from the following four aspects.

\textbf{CNN methods:}. Since the breakthrough success of CNN models on ImageNet Large Scale Visual Recognition Challenge (ILSVRC) 2012~\cite{krizhevsky2012imagenet}, employing CNN models to other vision tasks becomes popular in the computer vision community. Razavian \etal~\cite{razavian2014cnn} evaluate the performance of CNN features on several vision tasks, including object recognition, fine-grained object recognition, and image retrieval. Meanwhile, DeCAF~\cite{DeCaf} also shows that CNN features work surprisingly well on image classification. Subsequently, Babenko et al~\cite{babenko2014neural} present a similar idea on image retrieval with fine-tuning on self-collected datasets to further improve retrieval accuracy. In addition, they adopt PCA to compress neural-codes for efficient search. All these methods adopt the neural code activation from the first full-connected layer.

\textbf{Attribute methods:} Attribute methods adopt discriminative outputs of multi-class classifiers as mid-level features for visual recognition tasks. In face recognition, Kuma \etal~\cite{kumar2009attribute} consider the labels of reference faces and face-components as attributes to describe other faces. Farhadi \etal~\cite{farhadi2009describing} describe objects using $64$ explicitly semantic attribute classifiers. However, it remains as an open challenge to exploit the discriminative output of CNNs to solve generic vision tasks.

\textbf{Pooling strategy:} Pooling is a general strategy to augment features. As one of the most well known work, spatial pyramid matching performs pooling over pyramid of regular grids~\cite{lazebnik2006beyond,yang2009linear}. Gong \etal~\cite{gong2014multi} encodes the activations of CNN fully connected layer by VLAD~\cite{JDSP10}, and then concatenates the encoded features over windows at three scale levels. Most of these pooling methods simply concatenate features from different grids of scales. On the contrary, decision-level cross-region pooling has been applied when there are multiple region/patch candidates~\cite{sermanet-iclr-14, WeiXHNDZY14}. In our work, since we use the semantic output of CNNs as regional features, it is fairly straightforward to perform pooling across different region proposals.

\textbf{Region proposals:} Methods for detecting region proposal are used in object detection to avoid exhaustive sliding window search across images and speed up the detection without noticeable loss of recall rates~\cite{girshick2013rich}. In general, region proposal detection is based on low-level features and visual cues to measure objectness of local regions to generate relatively fewer candidate windows. In the past few years, there have been extensive studies on this topic and many techniques are invented, including selective search~\cite{uijlings2013selective}, edge-boxes~\cite{edgebox}, BING~\cite{cheng2014bing}, multiscale combinatorial grouping (MCG)~\cite{mcg}, and so on. Recently, Jan Hosang et al. \cite{hosang2014good} evaluates ten region proposal methods, in which selective search and edge-boxes achieved consistently better performance in terms of ground truth recall, repeatability, and detection speed. Hence, we may employ them to produce region proposals as the first step of our deep attribute method.

\section{Deep Attribute with Regional Neural Codes}
\label{sec:proposed_method}
The proposed deep attribute method consists of the following four steps.
\begin{itemize}
\vspace{-0.04in}
 \addtolength{\itemsep}{-0.05in}
 \small
  \item[(1)] {\em Region proposals extraction}: We use advanced techniques like selective search~\cite{uijlings2013selective} or edge-boxes~\cite{edgebox} to extract semantical regions, as both of which show satisfactory performance in benchmarks \cite{hosang2014good}.
  \item[(2)] {\em Computing of neural codes for region proposals}: We use CNN models trained from 1000 categories of ILSVRC 2012 to produce 1000 dimensional semantic output for each extracted region proposals. Such computed neural codes will serve as the semantic input for the next pooling stage.
  \item[(3)] {\em Cross-region-pooling}: We make pooling per dimension across extracted regions to get a 1000-dimensional holistic representations. Different pooling layout schemes are applicable for further possible performance improvement.
  \item[(4)] {\em Context-aware region refining and classifier build}: A linear classifier can be trained over deep attribute representations for visual recognition tasks. We observed that only a small portion of regions are correlated to classification targets. We thus
      impose classifier feedback to pick out a few contextual regions and re-pooling to get a context-aware classifier.
\vspace{-0.04in}
\end{itemize}
The first two steps are easy to understand. We will describe the latter two items in details below.

\subsection{Cross Region Pooling (CRP)}
Given image $I$, a set of object region proposals $\{R_1,R_2,...,R_N\}$ are extracted by region detection algorithms like \cite{uijlings2013selective,edgebox}.
Then the regions are wrapped and feed to CNN models. Each region $R_i$ is thus represented by the output from soft-max layer as $F_i = (f_{i1}, \cdots, f_{iK})$, where $f_{ik}$ is the $k$-th dimensional neural code of $F_i$. The computed neural codes of all the regions are then aggregated with a pooling operation to construct a holistic representation of the input image $I$. The pooling could be either max-pooling or average-pooling. In our practice, we find max-pooling works better than average-pooling when do pooling over all region proposals.
Thus, the final code for the $k$-th dimension $\hat f_k$ is obtained as $\hat f_k = \max_{i=1}^N \{f_{ik}\}$. Figure~\ref{flowchart2:da} illustrates the details of the pooling scheme. Since the CNN models are trained on 1000 categories, we derive a 1000 dimensional deep attribute after this pooling procedure.

As the single-scale cross-region-pooling does not consider the layout of regions across scales (like the {\em person} and the {\em horse} in Figure \ref{flowchart2:da}),
we may further enhance it with different pooling layout scheme.
This paper studies two layout schemes: multi-scale pooling scheme or spatial pyramid pooling scheme.
In multi-scale pooling, we divide the regions into different scale-interval groups, perform pooling over the regions within each group independently, and then concatenate the deep attributes from each group together, yielding a final holistical representation of the image.
Particularly, the scale of a region is defined as the ratio of its area proportional to the area of the whole image, or the area of the image bounding box when available.
In our experiments, we use five scale-intervals, which are $(0,\frac{1}{16}]$,$(\frac{1}{16},\frac{1}{8}]$,$(\frac{1}{8},\frac{1}{4}]$,$(\frac{1}{4},\frac{1}{2}]$ and $(\frac{1}{2},1]$. Therefore, the final representation is a $5000$-dimension deep attribute feature.
In contrast, spatial pyramid pooling (SPP) divides the whole image into 1x1, 2x2 and 4x4 grids, and make cross-region-pooling in each grid, and then concatenate features together for a holistic representation.

\begin{figure}[t]
\small
\centering
\includegraphics[width=0.65\linewidth]{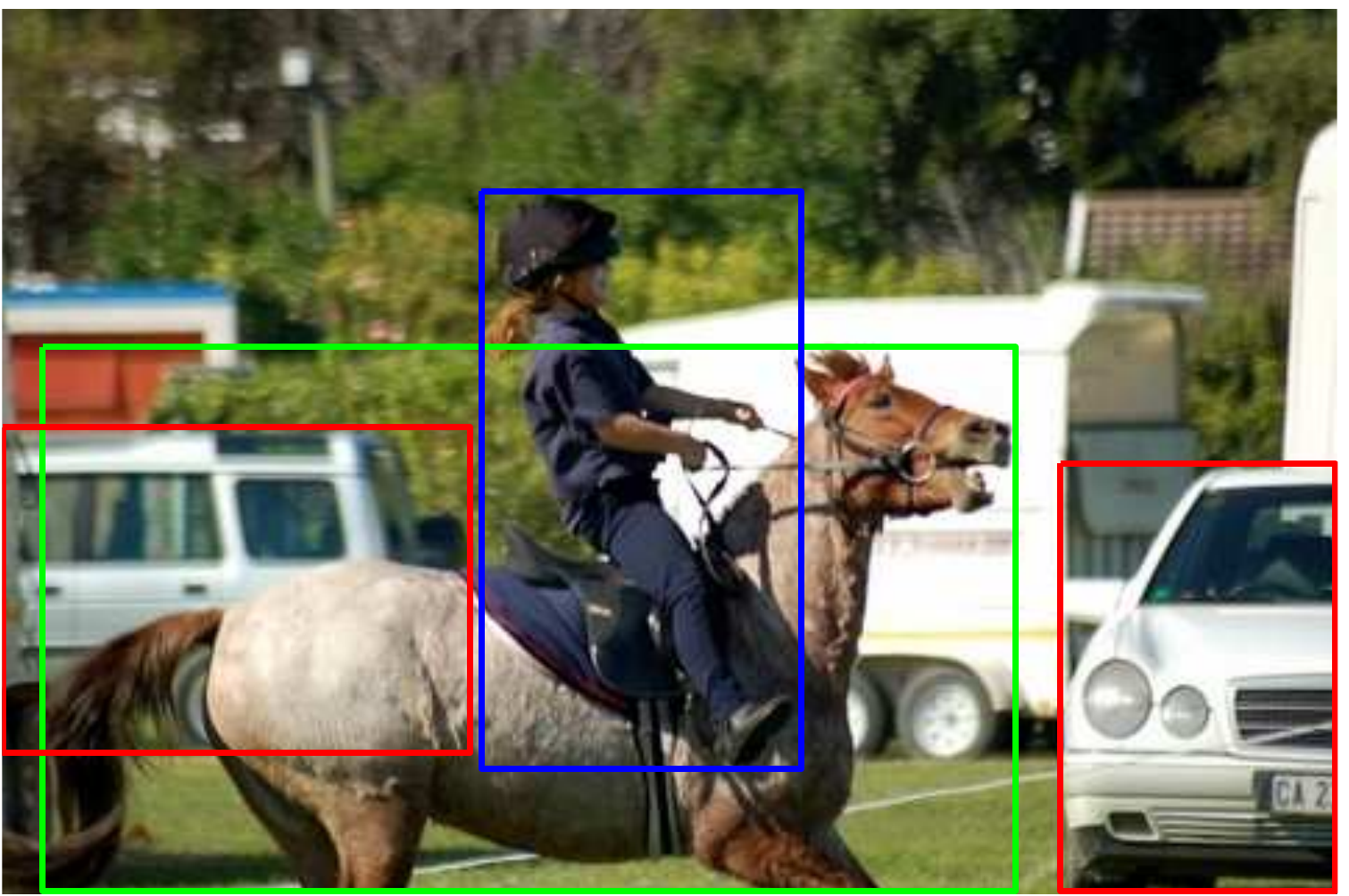}
\caption{Context region illustration. If we set ``hourse" as the classification target, the green box region is target region, the blue box region (human) is the context region, while the red regions (car) are background clutter/noise regions.}
\label{flowchart2:context}
\vspace{-0.15in}
\end{figure}

\subsection{Context Aware Region Refining (CARR)}
Region proposal algorithms usually produce thousands of regions for input images to ensure high recall rate \cite{hosang2014good}.
For a certain classification task, we observed that extracted regions could be divided into three categories.
First, some regions are directly related to the classification target.
Second, some regions can be viewed as useful context information.
Third, some regions belong to background clutter, which has little relationship to the classification task.
For instance, for a horse classification in Figure \ref{flowchart2:context}, the horse region is target region, and the human region can be viewed as a context region, while the car region is certainly background noise.
To improve the accuracy, we should maximally exploit the context information, while suppress the clutter information.
Note that the background clutter regions are category specific.
Again in Figure \ref{flowchart2:context}, for car classification, the horse region is background noise.
This inspires us imposing classifier feedback information to pick out category-specific context regions.
We therefore propose a context-aware region refining (CARR) algorithm based on this observation.

Suppose linear classifier $h^c(x)$ = $W^c * x$ is trained from features $x$ by cross-region-pooling for category $c$. Given image $I$,
region proposal algorithm extracted $N$ regions, and region $R_k$ is represented by semantical neural codes $F_k$. We define a score for region $R_k$ as
\begin{equation}
\label{eq1}
    S_k^c = h^c(F_k) = w^c \cdot F_k.
\end{equation}
The larger the score is, the higher the region correlated to category $c$.
A region is defined as category-$c$ {\em contextual region} if only if $S_k^c > \theta$.

In practice, for each image $I$ and category $c$, we rank the regions according to their score $S_k^c$.
We then pick top-$K$ regions as contextual regions, and apply cross-region-pooling on these top-$K$ regions again to get a new holistic representation for each category separately.
A refined linear classifier for category $c$ is then learned on these updated representations.
This procedure can be run iteratively to refine both context regions and linear classifiers.

Suppose we run the procedure with $T$ iterations, we thus have $T$ classifiers $\{h^c_t(x)\}_{t=1}^T$ for each category $c$.
Intuitively, we can directly use the final classifier $h^c_T(x)$. However, classifier ensemble may bring more accuracy gains.
We can fuse $h^c_t(x)$ together as
\begin{equation}
\label{eq2}
    H^c_T(x) = \sum\nolimits_{t=1}^T \alpha_t h^c_t(x) = \sum\nolimits_{t=1}^T \alpha_t  w_t^c \cdot x ,
\end{equation}
where $\alpha_t$ is a weight coefficient for $t$-th classifier.
Empirically, $\alpha_t$ can be estimated by grid search in a cross-validation manner. Since this fusion is similar to boosting,
we give $\alpha_t$ estimation followed by the AdaBoost rule as
\begin{equation}
\label{eq3}
\alpha_t = \log \frac{1 - E_t}{E_t},
\end{equation}
where $E_t$ is the classification error on training set at $t$-th iteration. Algorithm \ref{algo:carr} lists the training procedure.

In the testing phase, we apply $H^c_T(x)$ on each image, and do context-aware region refining based on classifier label.
Algorithm \ref{algo:test} lists the detailed testing procedure.

\begin{algorithm}
\small
\caption{Learning CARR Classifier}
\SetAlgoLined
\KwData{For each image $I$, given regions $\{R_k\}_I$ and corresponding soft-max output $\{F_k\}_I$}
\For{$t$ = $1$:$T$}{
    \For{each category $c$}{
        \For{each image $I$}{
            \For {each region $R_k$}{
                Compute $S_k^c(F_k)=H_{t-1}^c(F_k)$\;
            }
          Sort $S_k^c$ in descending order\;
          Pick top-$K$ regions from all $N$ regions\;
          Cross-region-pooling on top-$K$ regions\;
          Get new representation $F_I$\;
        }
      Train new linear classifier $h^c_t(x)$ based on $\{F_I\}$\;
      Compute error-rate $E_t$ for $h^c_t(x)$\;
      Update $H_t^c$ according to Eq. \ref{eq2}\;
  }
}
\KwResult{Output fused classifier $H^c_T(x)$.}
\label{algo:carr}
\end{algorithm}
\begin{algorithm}
\small
\caption{Prediction with CARR Classifier}
\SetAlgoLined
\KwData{Input testing image $I$ and region proposals $\{R_k\}$}
\For{each classifier $H^c(x)$}{
    \For{$t$ = $1$:$T$}{
        \For{each region $R_k$}{
             Compute score $S_k^c=H^c_{t}(F_k)$\;
        }
        Sort $S_k^c$ in descending order\;
        Pick top-$K$ regions from all $N$ regions\;
        Cross-region-pooling on top-$K$ regions\;
        Get new representation $F_I$\;
        Predict image score with $S^c_I = H^c_{t}(F_I)$\;
    }
}
\KwResult{Category label $\hat{c}= arg\max\limits_{c} S^c_I$.}
\label{algo:test}
\end{algorithm}
\section{Implementation and Performance Study}\label{sec:study}
This section will give implementation details and study performance impact of different choices.

Through this section, we study the performance on PASCAL VOC 2007 image classification benchmark with standard protocol.
PASCAL VOC \cite{everingham2010pascal} is a very challenging benchmark for object recognition. It contains images of $20$ categories including animals, handmade objects and natural objects. The objects are at different locations and scales with clutter background.
Even if objects are annotated with bounding box, this study does not use this information.
The accuracy is measured by mean average-precision (mAP) over 20 categories on the benchmark.

In this paper, we adopt the CNN models trained on ImageNet dataset~\cite{ILSVRCarxiv14}, which contains 1.2 million images associated with 1000 semantic categories.
We make experiments based on the Caffe deep learning framework \cite{jia2014caffe}.
It should be highlighted here that we do not make any data augmentation at all.
For each region, we only feed the single center cropping into CNN for computing neural-codes output.
Besides, we did not fine-tune the CNN model on the given datasets.
After deep attribute feature is extracted, we process the feature with a RootSIFT trick normalization as in \cite{Arandjelovic12}.
Linear SVM classifier is trained by choosing the best cost parameter $C$ in SVM on the train/val split.

\subsection{CNN Models: Alex's vs VGG's}
We first compare two CNN models: Alex's net \cite{krizhevsky2012imagenet} vs VGG's net \cite{Simonyan14c}.
There are several different VGG CNN models public available, in which we used the 16 layer VGG model.
In this study, we only apply the single-scale cross-proposal max-pooling without context refining.
We adopt selective search as region proposal generation algorithms.
Alex's net achieves 80.8\% mAP on 20 categories of VOC 2007, while VGG's net achieves 85.6\% mAP.
It is obvious that VGG's net outperforms Alex's net with a big margin.
This consistent with the factor that VGG's net performs better than Alex's net on ImageNet.
Hence, we adopt VGG's net in all the remaining studies.

\subsection{Selective Search vs Edge Box}
We then compare two best region proposal generation algorithms according to \cite{hosang2014good}, i.e., selective search \cite{uijlings2013selective} and edge-box \cite{edgebox}. Figure \ref{fig5a} illustrates recall rate with respect to number of top-regions on PASCAL VOC 2007 by both selective search and edge box. This graph is generated in the following way.
Since PASCAL VOC 2007 dataset gives object bounding box annotations, we employ selective search or edge-box to generate a set of region proposals and sort them according to region scores.
Then, we pick top-$K$ regions, and compute the intersection-over-union (IoU) rate (IoU) using top-$K$ regions with the ground truth.
One region is counted as a recall when the IoU is larger than 0.5. Changing the value of $K$, we got the graph. From this graph, we can see that edge-box works slightly better than selective search.

A single scale VGG's net is adopted in this study with max-pooling for region aggregation.
All the generated region proposals are used in cross-region-pooling. Selective search achieves 85.6\% mAP on VOC 2007,
while edge box achieves 86.1\% mAP.
This shows edge box works slightly better than selective search, and is consistent with results in Figure \ref{fig5a}.
Specially, edge box is more than 20 times faster than selective search. Therefore, we adopt edge box in all the remaining studies.

\begin{figure}[t]
  \footnotesize
 \centering
   \subfigure[]{
    \includegraphics[width=1.55in]{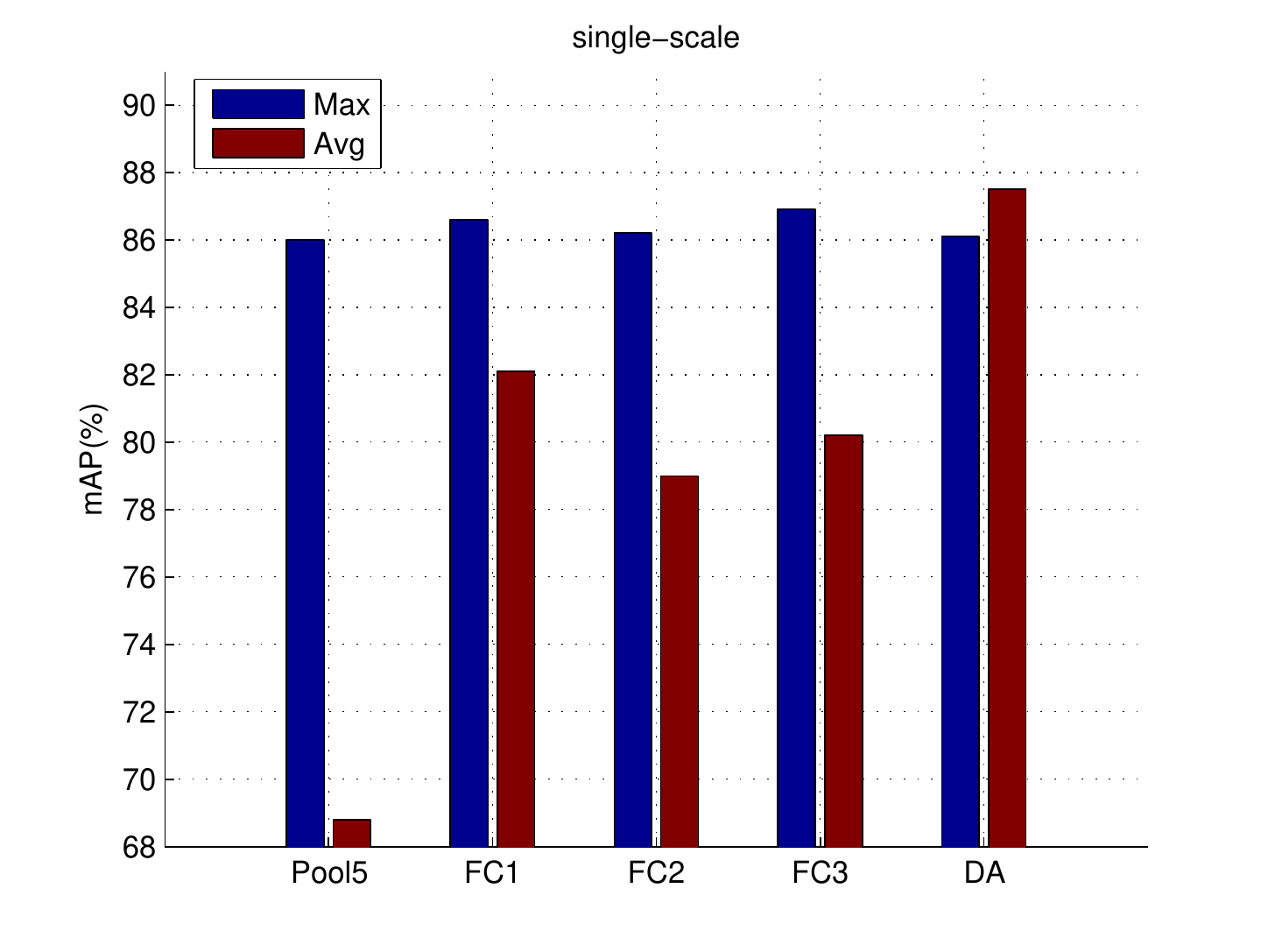}}
    \hspace{0.02in}
   \subfigure[]{
   \includegraphics[width=1.55in]{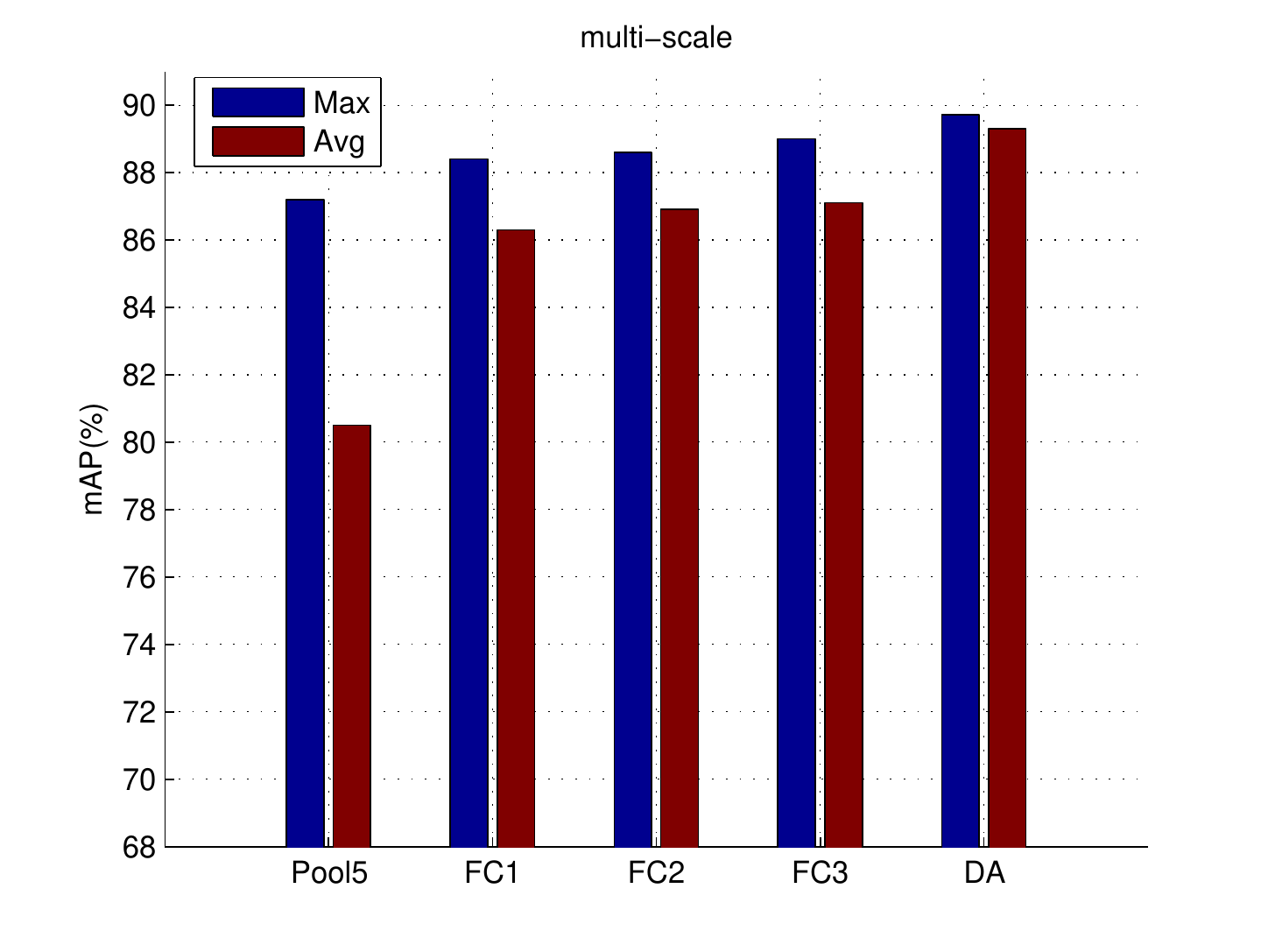}}
 \caption{Comparison of accuracy on cross-region-pooling with feature aggregating from different CNN layer on VOC 2007. (a) is single-scale pooling case,
  (b) is the multi-scale pooling case.}
 \label{fig4:difflayer}
 \vspace{-0.13in}
\end{figure}

\subsection{Different Pooling Schemes}
We also compare different pooling schemes. First, we compare pooling layout schemes under cross-proposal max-pooling.
Single-scale makes cross-region-pooling (CRP) over the whole image. Spatial-pyramid-pooling divides the whole image into 1x1 grids, 2x2 grids plus one center grid. It makes CRP on each grid, and then concatenates features from each grid together. Multi-scale-pooling makes CRP over five different region scales according to region size proportion to image size. Features from different scales are concatenated together to a holistic representation.
In this study, we adopt VGG's net with max-pooling. Experiments show that single-scale achieves 86.1\% mAP, spatial-pyramid-pooling achieves 87.2\% mAP,
and the multi-scale pooling achieves 89.7\% mAP. Hence, multi-scale pooling outperforms other pooling scheme with a big margin.
As spatial-pyramid-pooling does no show advantages over multi-scale case, we will not take it into consideration in future studies.

Second, we compare max-pooling to average-pooling under both single-scale layout and multi-scale layout.
In single-scale case, max-pooling achieves 86.1\% mAP, while average-pooling achieves 87.3\% mAP.
In the multi-scale case, max-pooling achieves 89.7\% mAP, while average pooling achieves 89.3\% mAP.
That means, max-pooling is worse than average-pooling in single-scale case, while it is better than average-pooling in multi-scale case.

\begin{figure}[t]
  \footnotesize
 \centering
   \subfigure[]{\label{fig5a}
    \includegraphics[width=1.55in, height=1.5in]{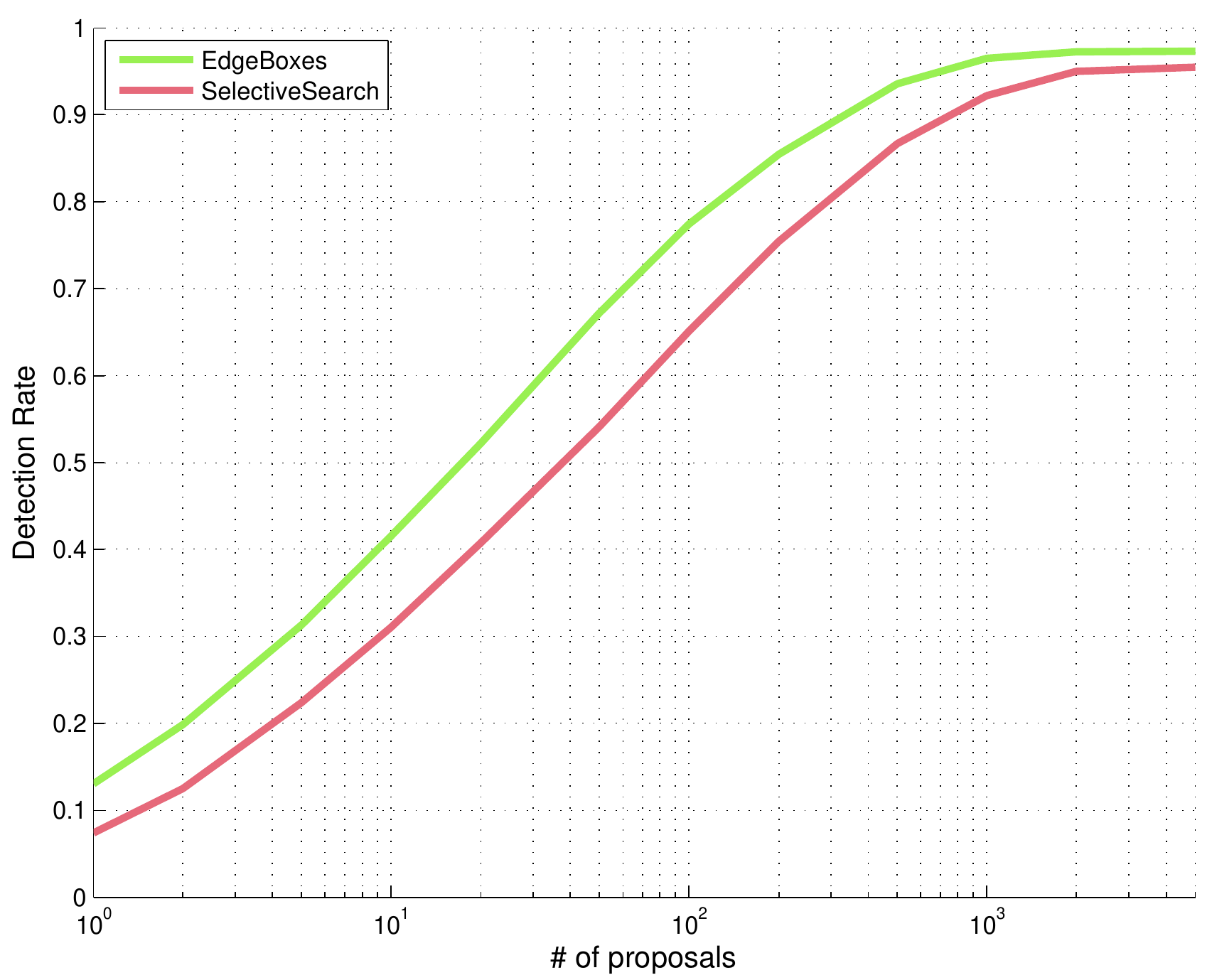}}
    \hspace{0.02in}
   \subfigure[]{\label{fig5b}
   \includegraphics[width=1.55in, height=1.5in]{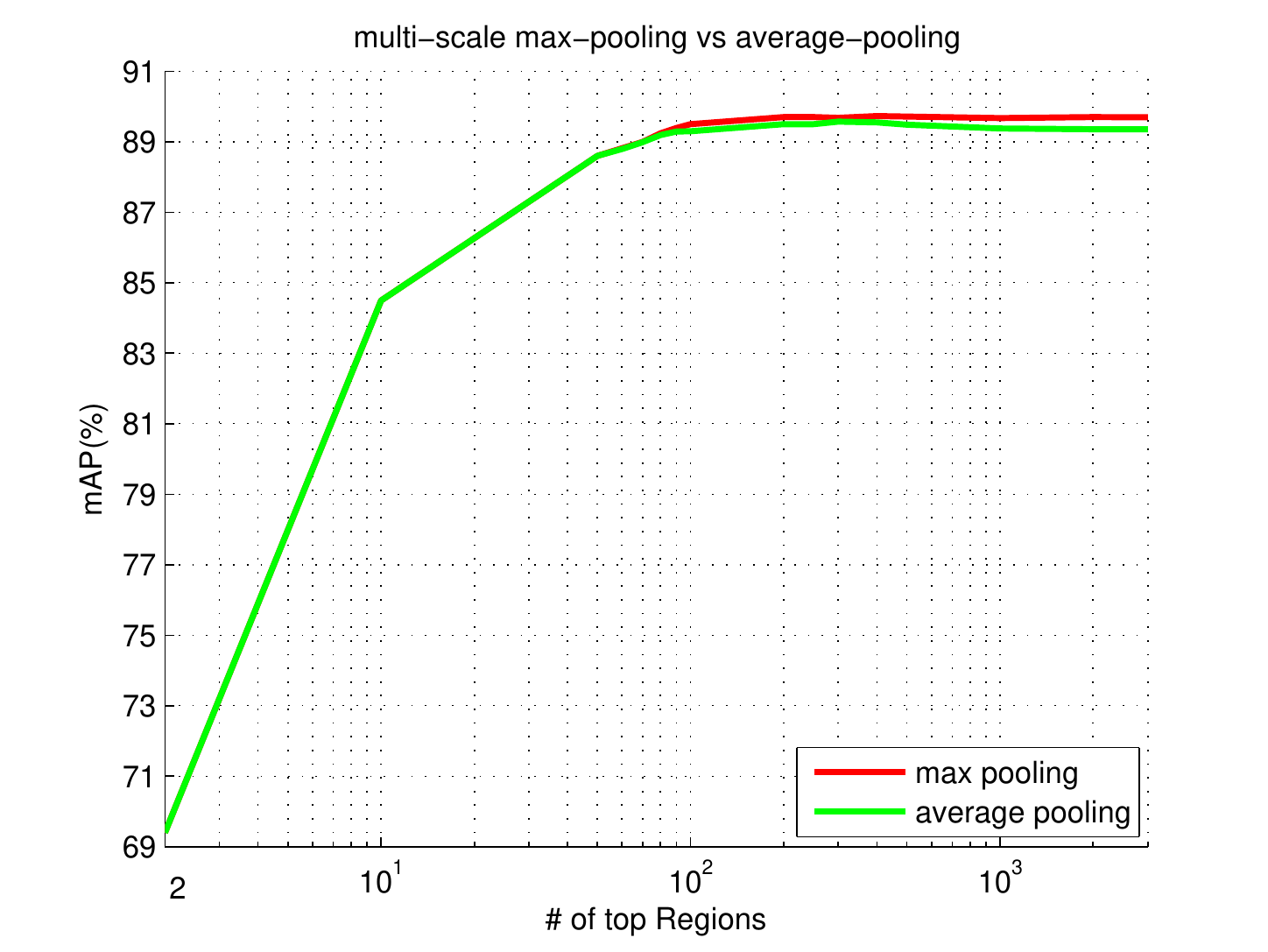}}
 \caption{(a) Recall rate vs number regions on VOC 2007 by edge-box and selective search. Here regions are opted in the order by their scores. (b) Comparison accuracy when different number of region proposals are used by the order of region score.}
 \label{fig5:region}
 \vspace{-0.13in}
\end{figure}

\subsection{Different CNN Layers}
We aggregate output from soft-max layer with cross-region-pooling as feature representations.
What if we used output of other CNN layers as features?
In this study, we compare 5 different CNN layers (pool5, fc1, fc2, fc3, soft-max) on their performance with cross-region-pooling.
Figure \ref{fig4:difflayer} illustrates the results on both single-scale case and multi-scale case. We can conclude that
(1) In multi-scale case, max-pooling consistently outperform averaging-pooling on different layers, while DA (soft-max) layer achieves 89.7\% mAP, which beats all the other layers on both max-pooling and average-pooling;
(2) In single-scale case, max-pooling outperforms averaging-pooling on all layers except for the soft-max layer.
Soft-max layer with average-pooling performs the best over all the other layers in this case.

As superb accuracy reached by soft-max layers and relative lower dimensionality (1000 for single scale, 5000 for multi-scale), we choose soft-max layer and multi-scale pooling in the proposed framework.

\begin{figure}[t]
\small
\centering
\includegraphics[width=0.495\linewidth]{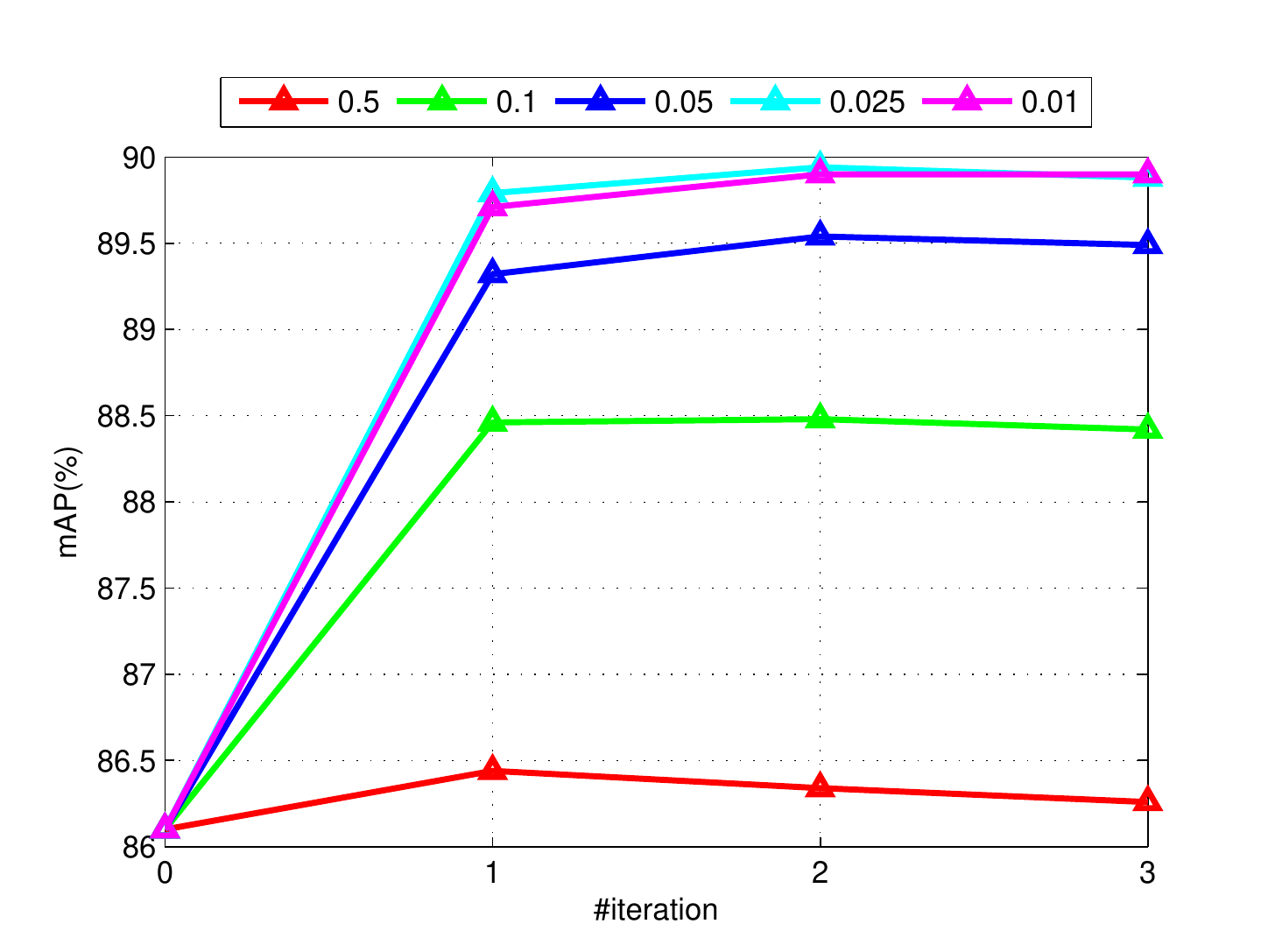}
\includegraphics[width=0.495\linewidth]{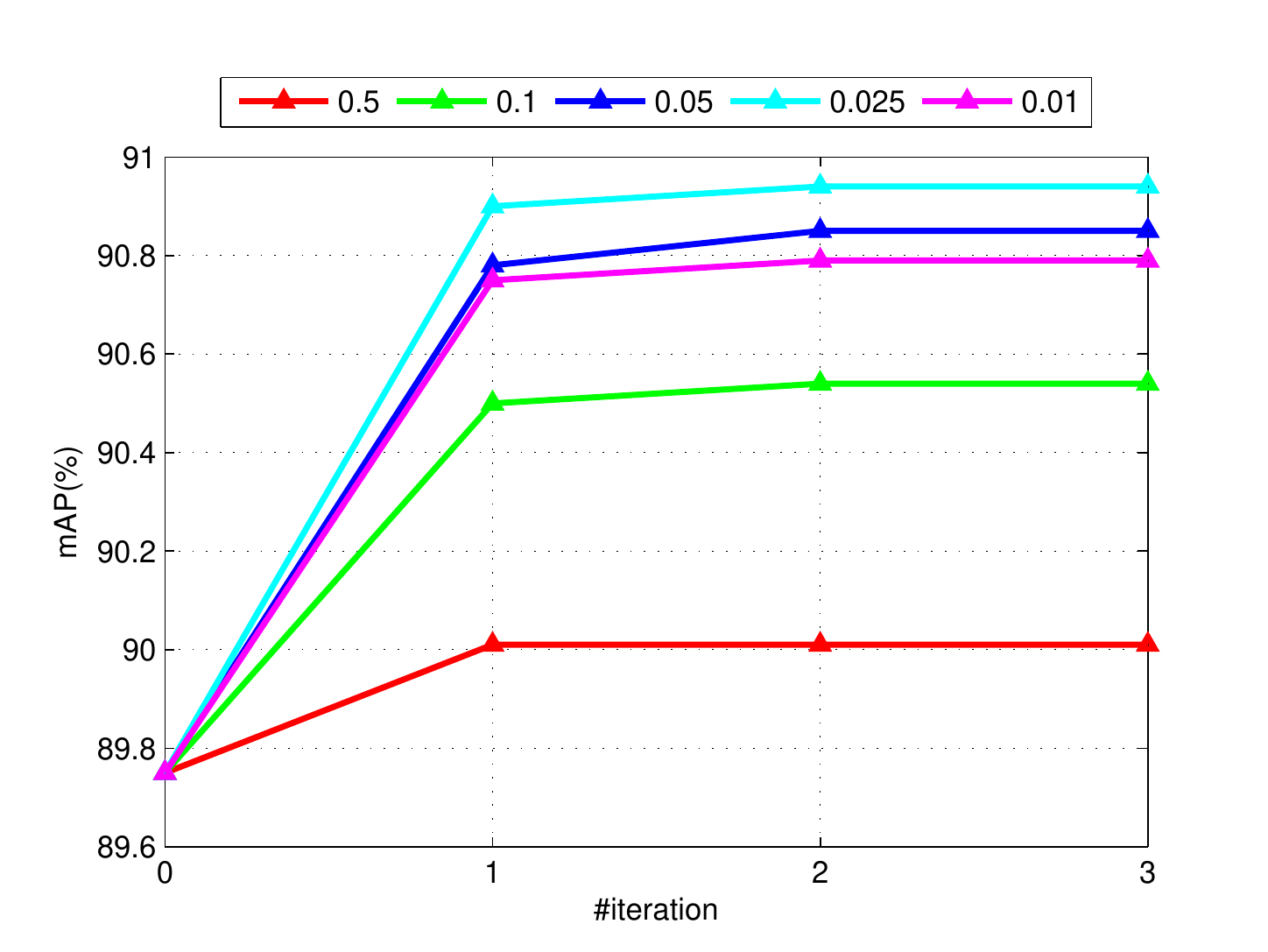}
\caption{The impact of \#iterations and the ratio of context regions ($\beta$ = $\frac{K}{N}$, where $N$ is total number region proposals) in CARR algorithm. The left one is results of single scale case, while the right one is the results of multi-scale case.}
\label{fig5:carr}
\vspace{-0.13in}
\end{figure}

\subsection{How Many Regions Are Required?}
People may doubt whether cross-proposal pooling are useful, and how many regions are enough?
This study will answer these two questions. We use edge-box to generate region proposals.
We pick top-$K$ regions according to their region score for the multi-scale cross-region-pooling.
For different $K$ value, we get the accuracy on VOC 2007. The results are shown in Figure \ref{fig5b}.
It shows that more regions is better, and when the number of regions exceed 500, the accuracy is saturated.
That means we can just feed top 500 regions rather than all 1500+ regions to CNN, which can save a lot of computation cost.

\begin{figure}[t]
\small
\centering
\includegraphics[width=2.7in, height=1.6in]{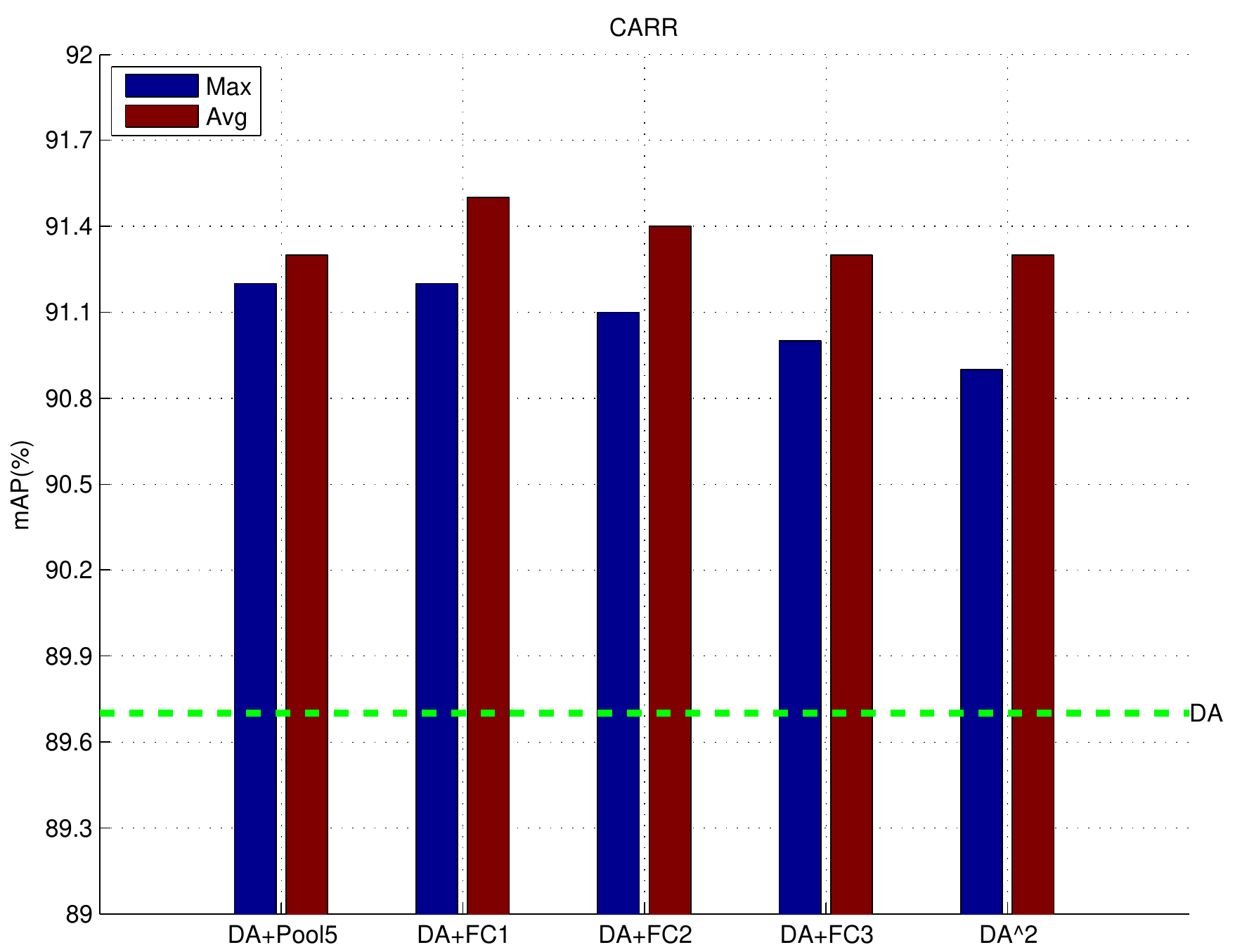}
\caption{Combining DA with different layers in refining step. The global CRP step adopts max-pooling DA features,
 while the refining step may be DA, pool5, FC1, FC2, FC3 features with both max-pooling and average-pooling. The green horizontal line indicate result by multi-scale deep attributes without region refining.}
\label{fig6:carr}
\vspace{-0.13in}
\end{figure}

\begin{table*}[t]
\caption{Classification results (AP in $\%$) comparison with state-of-the-art approaches on VOC 2007 (trainval/test).
$^\ast$ indicates methods fine-tuning on the new dataset. DA+FC1$^\ast$ is trained on VOC 2007 and VOC 2012 combined training set.}
\footnotesize
\begin{tabularx}{\textwidth} {cXXXXXXXXXXXXXXXXXXXXX} \hline
  & plane & bike & bird & boat & bottle & bus & car & cat & chair & cow & table & dog & horse & motor & person & plant & sheep & sofa & train & tv & mAP \\ \hline
INRIA \cite{harzallah2009combining}& 77.2 & 69.3 & 56.2 & 66.6 & 45.5 & 68.1 & 83.4 & 53.6 & 58.3 & 51.1 & 62.2 & 45.2 & 78.4 & 69.7 & 86.1 & 52.4 & 54.4 & 54.3 & 75.8 & 62.1 & 63.5 \\
CNN\ S$^\ast$ \cite{Chatfield14} & 95.3 & 90.4 & 92.5 & 89.6 & 54.4 & 81.9 & 91.5 & 91.9 & {64.1} & 76.3 & 74.9 & 89.7 & 92.2 & 86.9 & {95.2} & 60.7 & 82.9 & 68.0 & 95.5 & 74.4 & 82.4 \\
CNNaug-SVM \cite{razavian2014cnn} & 90.1 & 84.4 & 86.5 & 84.1 & 48.4 & 73.4 & 86.7 & 85.4 & 61.3 & 67.6 & 69.6 & 84.0 & 85.4 & 80.0 & 92.0 & 56.9 & 76.7 & 67.3 & 89.1 & 74.9 & 77.2 \\
HCP-1000C$^\ast$ \cite{WeiXHNDZY14} & 95.1 & 90.1 & 92.8 & 89.9 & 51.5 & 80.0 & 91.7 & 91.6 & 57.7 & 77.8 & 70.9 & 89.3 & 89.3 & 85.2 & 93.0 & 64.0 & 85.7 & 62.7 & 94.4 & 78.3 & 81.5 \\
HCP-2000C$^\ast$ \cite{WeiXHNDZY14} & 96.0 & 92.1 & 93.7 & 93.4 & 58.7 & 84.0 & {93.4} & 92.0 & 62.8 & 89.1 & 76.3 & 91.4 & 95.0 & 87.8 & 93.1 & 69.9 & 90.3 & 68.0 & 96.8 & 80.6 & 85.2 \\
VGG-16-19-Fusion \cite{Simonyan14c}    &  98.9 & 95.0 & 96.8 & 95.4 & 69.7 & 90.4 & 93.5 & 96.0 & 74.2 & 86.6 & \textbf{87.8} & \textbf{96.0} & 96.3 & 93.1 & 97.2 & 70.0 & 92.1 & \textbf{80.3} & 98.1 & 87.0 & 89.7 \\
FV+LV-20-VD \cite{yang2015can}&  97.9 & 97.0 & 96.6 & 94.6 & 73.6 & \textbf{93.9} & 96.5 & 95.5 & 73.7 & 90.3 & 82.8 & 95.4 & \textbf{97.7} & 95.9 & \textbf{98.6} & 77.6 & 88.7 & 78.0 & 98.3 & 89.0 & 90.6 \\ \hline
DA$^2$     &  99.1 & 96.2 & 96.3 & 96.0 & 79.2 & 91.9 & 95.8 & 95.9 & 74.0 & 87.9 & 83.1 & 94.4 & 95.5 & 94.8 & 98.1 & 77.1 & 94.7 & 77.3 & 98.5 & 90.4 & 90.8 \\
DA+FC1    &  \textbf{99.4} & 96.9 & \textbf{96.9} & 95.6 & 78.9 & 92.4 & 96.5 & 96.6 & 74.3 & 89.5 & 82.7 & 95.5 & 95.5 & 95.4 & 98.4 & 78.2 & 95.3 & 78.8 & 98.6 & \textbf{92.2} & 91.4 \\
DA+FC1$^\ast$    & \textbf{99.4} & \textbf{97.5} & 96.8 & \textbf{96.6} & \textbf{81.3} & 92.9& \textbf{96.8} & \textbf{97.1} & \textbf{75.6} & \textbf{93.7} & 84.5 & 95.8 & 96.8 & \textbf{96.0} & \textbf{98.6} & \textbf{81.9} & \textbf{97.7} & 80.2 & \textbf{99.0} & 91.5 & \textbf{92.5}\\ \hline

\end{tabularx}
\label{table:pascal07:result}
\vspace {-0.1in}
\end{table*}

\begin{table*}[t]
\caption{Classification results (AP in $\%$) comparison with state-of-the-art approaches on VOC 2012 (trainval/test).}
\footnotesize
\begin{tabularx}{\textwidth} {cXXXXXXXXXXXXXXXXXXXXX} \hline
  & plane & bike & bird & boat & bottle & bus & car & cat & chair & cow & table & dog & horse & motor & person & plant & sheep & sofa & train & tv & mAP \\ \hline
HCP-1000C~\cite{WeiXHNDZY14} &    97.5 & 84.3 & 93.0 & 89.4 & 62.5 & 90.2 & 84.6 & 94.8 & 69.7 & 90.2 & 74.1 & 93.4 & 93.7 & 88.8 & 93.2 & 59.7 & 90.3 & 61.8 & 94.4 & 78.0 & 81.7 \\
HCP-2000C~\cite{WeiXHNDZY14} &    97.5 & 84.3 & 93.0 & 89.4 & 62.5 & 90.2 & 84.6 & 94.8 & 69.7 & 90.2 & 74.1 & 93.4 & 93.7 & 88.8 & 93.2 & 59.7 & 90.3 & 61.8 & 94.4 & 78.0 & 84.2 \\
VGG-16~\cite{Simonyan14c} &      99.0 & 88.8 & 95.9 & 93.8 & 73.1 & 92.1 & 85.1 & 97.8 & 79.5 & 91.1 & 83.3 & 97.2 & 96.3 & 94.5 & 96.9 & 63.1 & 93.4 & 75.0 & 97.1 & 87.1 & 89.0 \\
VGG-16-19-Fusion~\cite{Simonyan14c}&   99.1 & 89.1 & \textbf{96.0} & 94.1 & 74.1 & 92.2 & 85.3 & 97.9 & 79.9 & 92.0 & \textbf{83.7} & \textbf{97.5} & 96.5 & 94.7 & 97.1 & 63.7 & \textbf{93.6} & 75.2 & 97.4 & 87.8 & 89.3 \\
FV+LV-20-VD~\cite{yang2015can} &   98.9 & 93.1 & \textbf{96.0} & 94.1 & 76.4 & 93.5 & 90.8 & 97.9 & 80.2 & \textbf{92.1} & 82.4 & 97.2 & \textbf{96.8} & 95.7 & 98.1 & 73.9 & \textbf{93.6} & \textbf{76.8} & 97.5 & 89.0 & 90.7 \\ \hline
DA$^2$&   99.0 & 92.0 & 95.1 & 93.9 & 80.0 & 93.1 & 89.7 & 97.6 & 80.8 & 88.0 & 83.1 & 95.7 & 94.3 & 94.8 & 97.8 & 74.4 & 92.3 & 74.5 & 97.3 & 90.8 & 90.2\\
DA+FC1&   \textbf{99.2} & \textbf{93.7} & 95.5 & 94.8 & \textbf{81.8} & 93.3 & 91.1 & \textbf{98.1} & 81.5 & 91.1 & 82.6 & 95.9 & 95.5 & 95.9 & 98.0 & 76.9 & 93.4 & 75.8 & \textbf{97.7} & 91.6 & 91.2\\
DA+FC1$^\ast$  & \textbf{99.2} & \textbf{93.7} & \textbf{96.0} & \textbf{95.2} & 81.7 & \textbf{94.3} & \textbf{91.6} & \textbf{98.1} & \textbf{81.9} & 91.7 & 83.5 & 96.3 & 95.6 & \textbf{96.0} & \textbf{98.2} & \textbf{77.8} & \textbf{93.6} & 74.7 & 97.6 & \textbf{91.9} & \textbf{91.4} \\\hline
\end{tabularx}
\label{table:pascal12:result}
\vspace {-0.1in}
\end{table*}

\subsection{Parameters in CARR}
In the context-aware region refining, we need to determine parameter $K$ and number of iterations.
Instead of directly set $K$, we define the context region ratio $\beta=\frac{K}{N}$, where $N$ is the total number of region proposals extracted.
Figure \ref{fig5:carr} illustrates the mAP curves on VOC 2007 with different $\beta$ and different iteration number.
It shows that $\beta$ = 0.025 gives the best results. As the number of total regions $N$ usually larger than 1500,
that means there are usually about 40 context regions picked out for each image and each category.
Besides, we observed that only one iteration of context region refining will bring $3.6\%$ accuracy improvement for single scale deep attributes (from 86.1\% to 89.7\%), while bring 1.1\% accuracy improvement for the multi-scale case (from 89.7\% to 90.8\%).
Further iterations do not bring notable gains.
Therefore, we set $\beta$=0.025, iteration number = 2 for all the remaining experiments.

\subsection{How About using Other Layers in Refining?}
In section 4.4, we showed that soft-max layer (DA) is the best with cross-region-pooling.
We therefore fixed the first step (global CRP step) with DA features.
In section 4.6, we adopted DA features in the region refining step.
Here we study other CNN layers in the refining step.
We compared DA+DA (aka, DA$^2$) to DA+Pool5, DA+FC1, DA+FC2, DA+FC3 on VOC 2007, in which
the first DA indicates the global CRP step, while the 2nd item (after `+' sign) indicates the layer used in the region refining step.
We also study the impact from max-pooling and average-pooling in the refining step.
Figure \ref{fig6:carr} illustrates the comparison results.

It is obvious that average-pooling works better than max-pooling in the refining step.
This is different from the global CRP step.
The reason is that the global CRP step contains all the regions which may contain many background clutter noise regions;
while in the refining step, most noise regions are suppressed.
Also we find that DA+FC1 works the best (91.4\% mAP). For the fusion of two steps, we still adopted the rule by
Eq.\ref{eq2} with $\alpha_t$ given by Eq.\ref{eq3}.

\section{Experimental Results}\label{sec:experiments}
We thoroughly evaluate the proposed approach on three vision tasks, \ie, image classification, fine-grained object recognition and visual instance retrieval.

According to previous study, we fixed parameters of the proposed framework as VGG-16 CNN model, edge-box for region extraction, multi-scale pooling layout,
and one CARR step with $\beta$=0.025. We report results both by DA$^2$ and DA+FC1 for all the recognition tasks.

\begin{table*}[th]
\centering
\small
\caption{Method comparison on PASCAL VOC 2007 \& 2012 benchmarks with detailed CNN settings. $\sharp$ indicates that methods bounding box information. Specially, HCP-2000C is trained from 2000 categories dataset from ImageNet. DA+FC1$^\ast$ is trained on VOC 2007 and VOC 2012 combined training set.}
\begin{tabular} {c|cccc|c c} \hline
Method & CNN Arch & Fine-tuned &  Augmentation & Features & 2007 mAP (\%) & 2012 mAP (\%)\\ \hline
CNNaug-SVM \cite{razavian2014cnn} & OverFeat \cite{sermanet-iclr-14} & No & Yes & FC & 77.2 & NA\\
CNN S  \cite{Chatfield14} & CNN-S \cite{Chatfield14} & Yes & Yes & FC & 82.4 & 83.2\\
HCP-1000C \cite{WeiXHNDZY14}& Alex's \cite{krizhevsky2012imagenet} & Yes & No & FC & 81.5 & 81.7\\
HCP-2000C \cite{WeiXHNDZY14}& Alex's \cite{krizhevsky2012imagenet} & Yes & Yes & FC & 85.2 & 84.2\\
VGG-16-19-Fusion \cite{Simonyan14c} & VGG-16+19 \cite{Simonyan14c} & No & Yes & FC & {89.7} & 89.3\\
FV+LV-20-VD$^\sharp$ \cite{yang2015can} & CNN-S/M \cite{Chatfield14}  & Yes & Yes & FC & {90.6} & 90.7\\ \hline
DA$^2$ & VGG-16 \cite{Simonyan14c}  & No & No & Soft-max & 90.8 & 90.2\\
DA+FC1 & VGG-16  \cite{Simonyan14c} & No & No & Soft-max+FC & 91.4 & 91.2\\
DA+FC1$^\ast$ & VGG-16  \cite{Simonyan14c} & No & No & Soft-max+FC & 92.5 & 91.4\\ \hline
\end{tabular}
\label{table:info}
\vspace{-0.1in}
\end{table*}

\subsection{Image Classification Task}
The image classification task is evaluated on the PASCAL VOC 2007 and 2012 benchmarks.

We first report the mean Average Precision (mAP) for the PASCAL VOC 2007 dataset with per-category results.
Table \ref{table:pascal07:result} shows per-category results in comparison to the state-of-the-art methods w.r.t each category on VOC 2007,
while Table \ref{table:pascal12:result} shows per-category results on VOC 2012.
Note that we listed one results by DA+FC1$^\ast$, which is trained on a combination training set from VOC 2007 and 2012.
We further listed CNN-related methods on their training configurations at Table \ref{table:info} on two benchmarks.

We can see that our method is fairly simple, without fine-tuning and data augmentation.
It outperforms current state-of-the-art method with a large margin.
Specially, we should mention two methods here.
First is the very-deep CNN method (aka, VGG-16-19 fusion).
The result of very-deep is by fusion both VGG-16 model and VGG-19 models with sophisticated multi-scale and multi-crop data augmentation.
The proposed approach is based on VGG-16 model with just center-crop region proposal inputs, yielding a 2.8\% margin over very-deep.
Second is the multi-view multi-instance framework by \cite{yang2015can} (FV+LV-20-VD), which takes FC layers output from region proposals as feature view and ground truth bounding box as label view, and combine them under a Fisher-vector framework. This method achieves 90.7\% accuracy. In comparison, the proposed approach outperforms FV+LV-20-VD with a margin 1.8\% with out using ground-truth bounding-box information and without fine-tuning.

\subsection{Fine-Grained Recognition Task}
The fine-grained recognition task is evaluated on the Oxford flower dataset,
which contains 102 categories of flowers. Each category contains 40 to 258 of images.
The flowers appear at different scales, pose and lighting conditions. The dataset provides segmentation for all the images.
However, we do not use this information in our experiment.

Our evaluation follows the standard protocol of this benchmark.
We report mean Accuracy on the Oxford 102 flowers dataset in Table \ref{table:flowers}.
DA$^2$ achieve $90.1\%$ accuracy, which is significantly higher than all existing methods.
When replacing refining step DA feature with FC1, it (DA+FC1) achieved 95.1\% accuracy.
Note that these results are obtained without using segmentation information, fine-tuning CNN networks.
More interestingly, the flower dataset has little concepts overlap with the pre-trained 1000-category CNN models.

\begin{table}[t]
\centering
\small
\caption{Accuracy comparison on fine-grained flower recognition. $w\slash o seg$ means those methods do not exploit the segmentation information for the task.}
\begin{tabular} {c|c} \hline
Method & mean Accuracy \\ \hline
Dense\ HOG+Coding+Pooling\ w\slash o seg \cite{angelova2013efficient}& 76.7 \\
Seg+Dense HOG+Coding+Pooling \cite{angelova2013efficient}& 80.7 \\
CNN-SVM w\slash o seg \cite{razavian2014cnn}& 74.7 \\
CNNaug-SVM w\slash o seg \cite{razavian2014cnn}& 86.8 \\ \hline
DA$^2$ w\slash o seg& {90.1} \\
DA+FC1 w\slash o seg& \textbf{95.1} \\ \hline
\end{tabular}
\label{table:flowers}
\vspace{-0.1in}
\end{table}

\subsection{Visual Instance Retrieval Task}
The visual instance retrieval task is evaluated on two datasets:
(1) Holidays \cite{jegou2008hamming} consists of 1491 vacation photographs distributed among 500 groups based on same object or scene. One image is selected from each group and is acted as a query image. This dataset is challenge due to viewpoint and scale variations.
(2) University of Kentucky Benchmark dataset \cite{nister2006scalable} (UKB) includes 10,200 indoor photographs uniformly from 2550 objects, and each image is used to query the rest. It is challenging due to viewpoint variations. In this experiment, we use the cosine similarity as the metric.

We report the mean Average Precision (mAP) for the Holidays dataset, and accuracy of top-4 retrieval results for the UKB dataset according to standard evaluation protocol on these two datasets.
Note that this task is for image retrieval, which does not require to train a classifier.
We just adopt single-scale and multi-scale max-pooling DA features for retrieval, without any special tricks.
Table \ref{table:retrieval} illustrates comparison results on these two datasets.
It shows that the proposed methods outperform the state-of-the-art methods with a notable margin.

\begin{table}[t]
\centering
\small
\caption{Accuracy comparison on visual instance retrieval. }
\begin{tabular}{c|c|c} \hline
Method  & Holidays & UKB \\ \hline
MOP-CNN \cite{gong2014multi} & 80.2 & -\\
Neural Codes \cite{babenko2014neural}  & 74.9 & 85.8\\
Neural Codes+ retrain \cite{babenko2014neural}  & 79.3& 82.3 \\
CNNaug-ss \cite{razavian2014cnn}   & 84.3 & 91.1 \\ \hline
DA (single-scale) & 85.1 & 93.5 \\
DA (multi-scale) & \textbf{86.1}&  \textbf{94.2}\\ \hline
\end{tabular}
\label{table:retrieval}
\vspace{-0.15in}
\end{table}

\begin{figure*}[th]
\centering
\small
\subfigure[Pascal VOC 2007]{
\begin{minipage}[b]{1.0\textwidth}
\centering
\includegraphics[height=0.09\linewidth,width=0.09\linewidth]{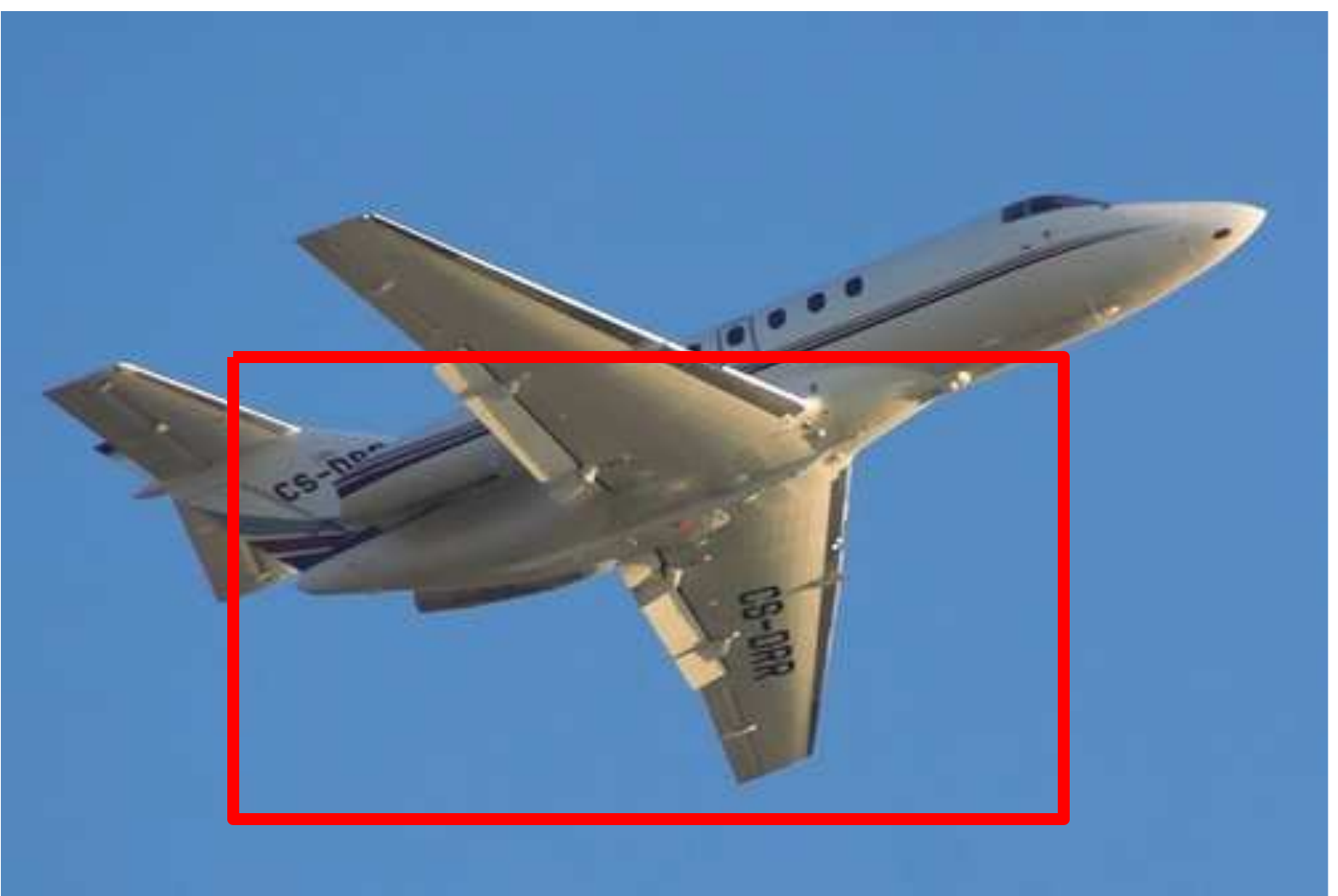}
\includegraphics[height=0.09\linewidth,width=0.09\linewidth]{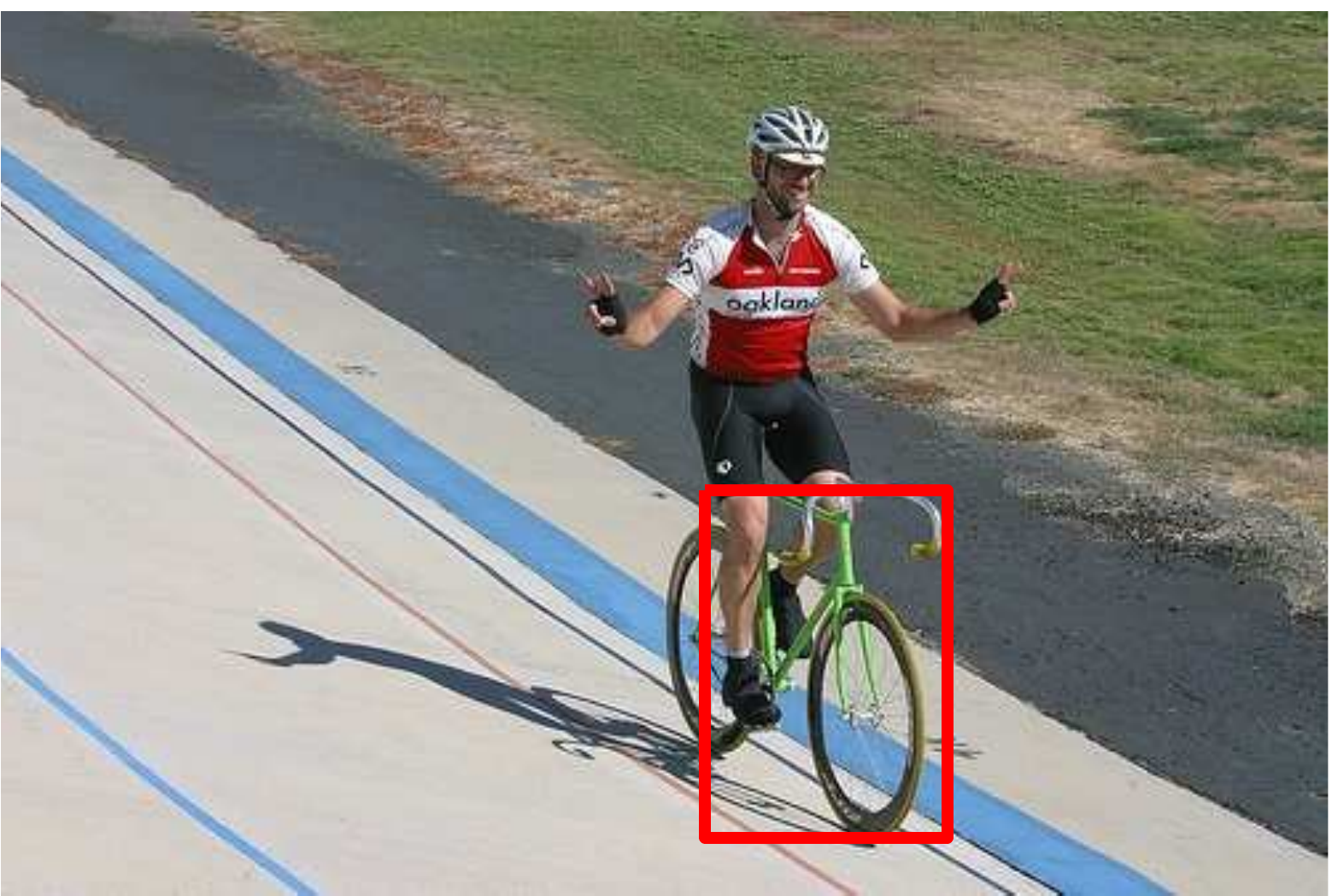}
\includegraphics[height=0.09\linewidth,width=0.09\linewidth]{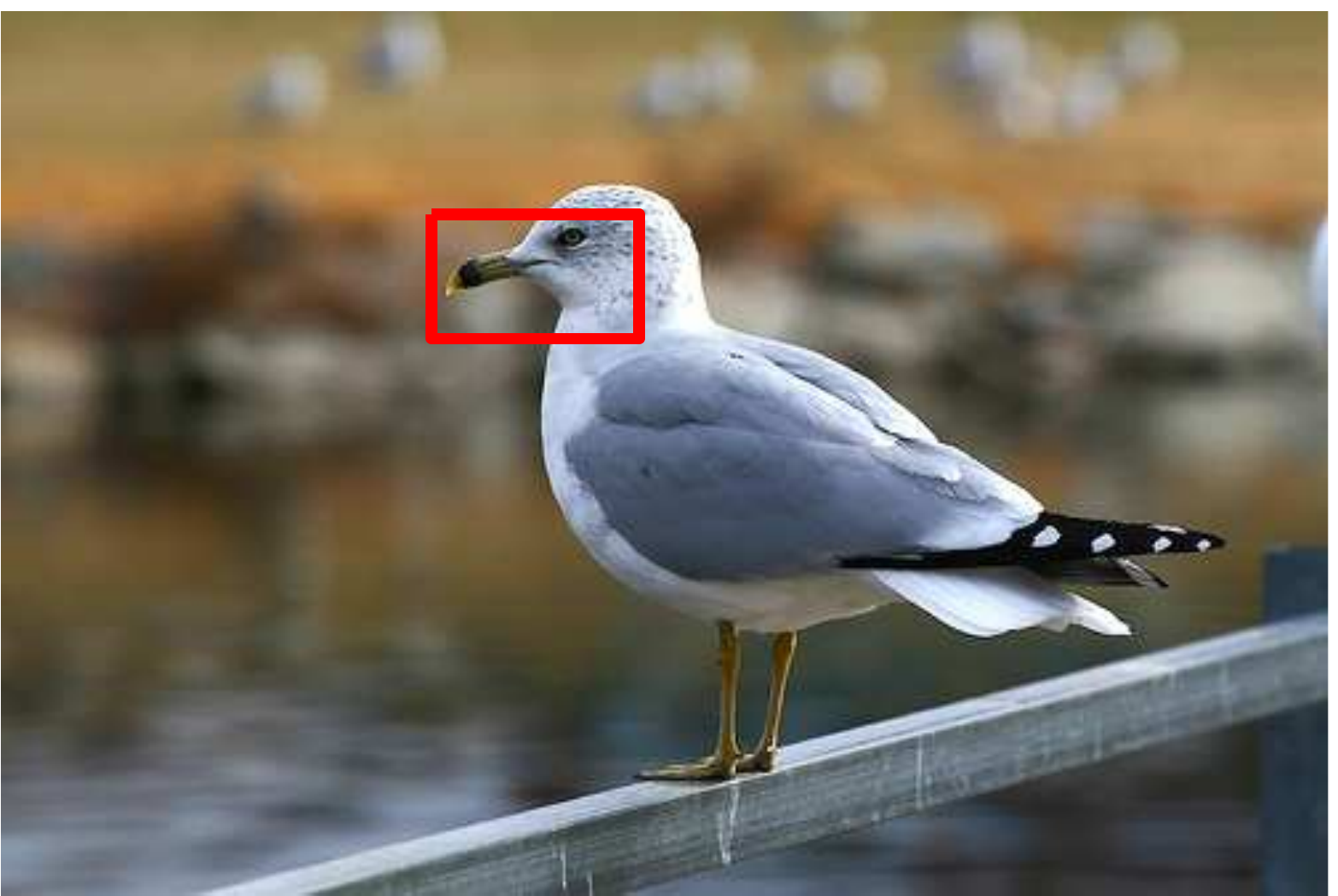}
\includegraphics[height=0.09\linewidth,width=0.09\linewidth]{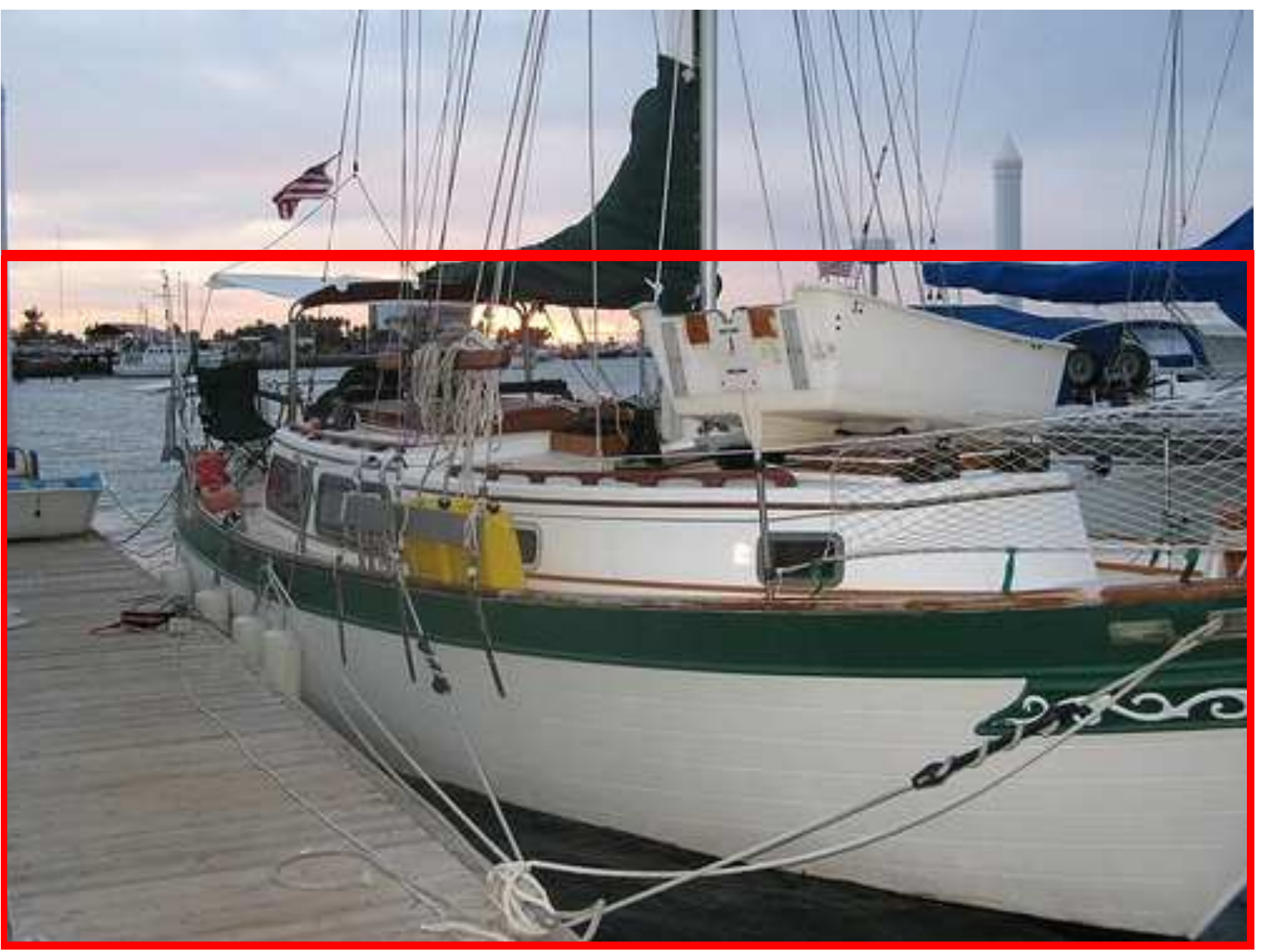}
\includegraphics[height=0.09\linewidth,width=0.09\linewidth]{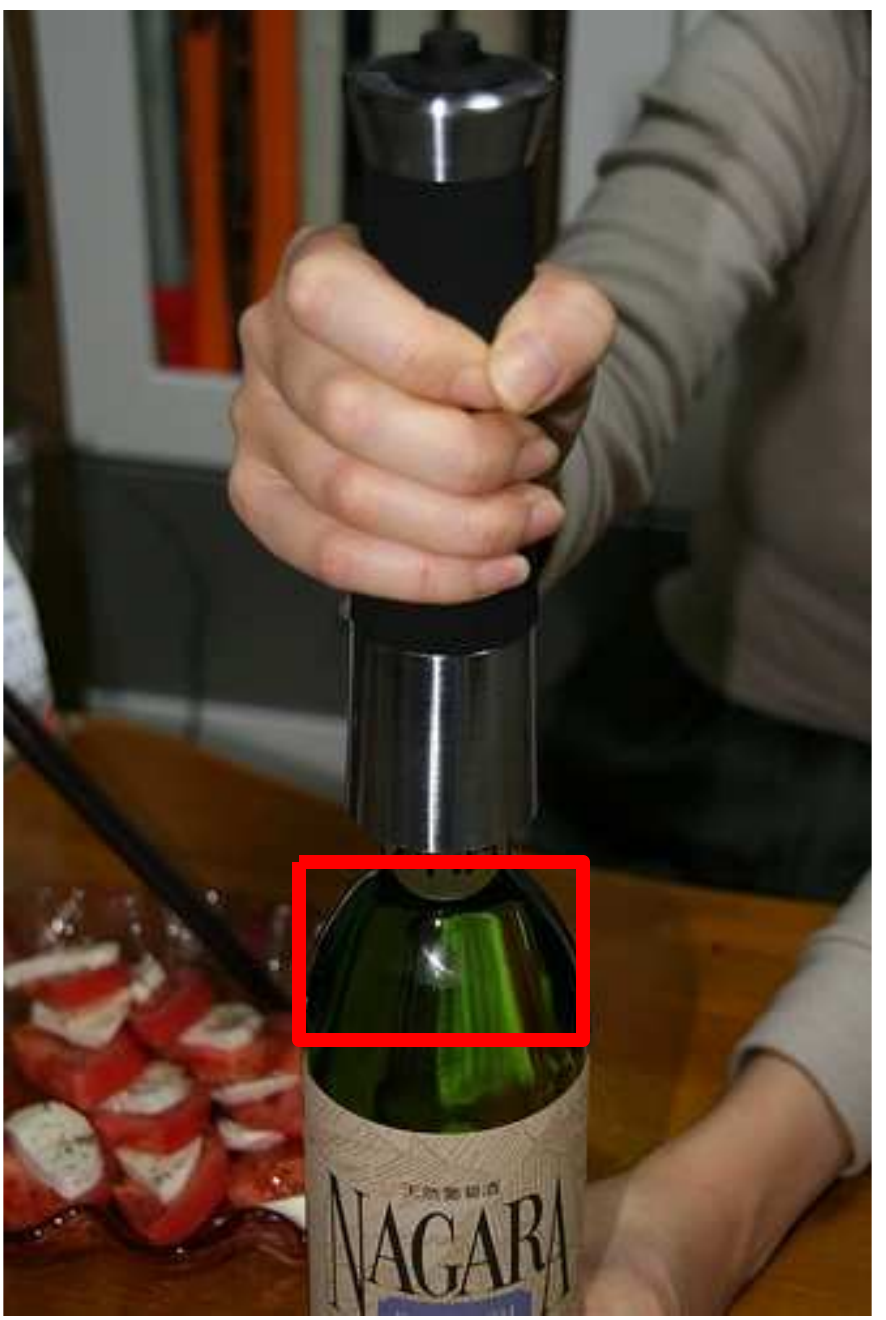}
\includegraphics[height=0.09\linewidth,width=0.09\linewidth]{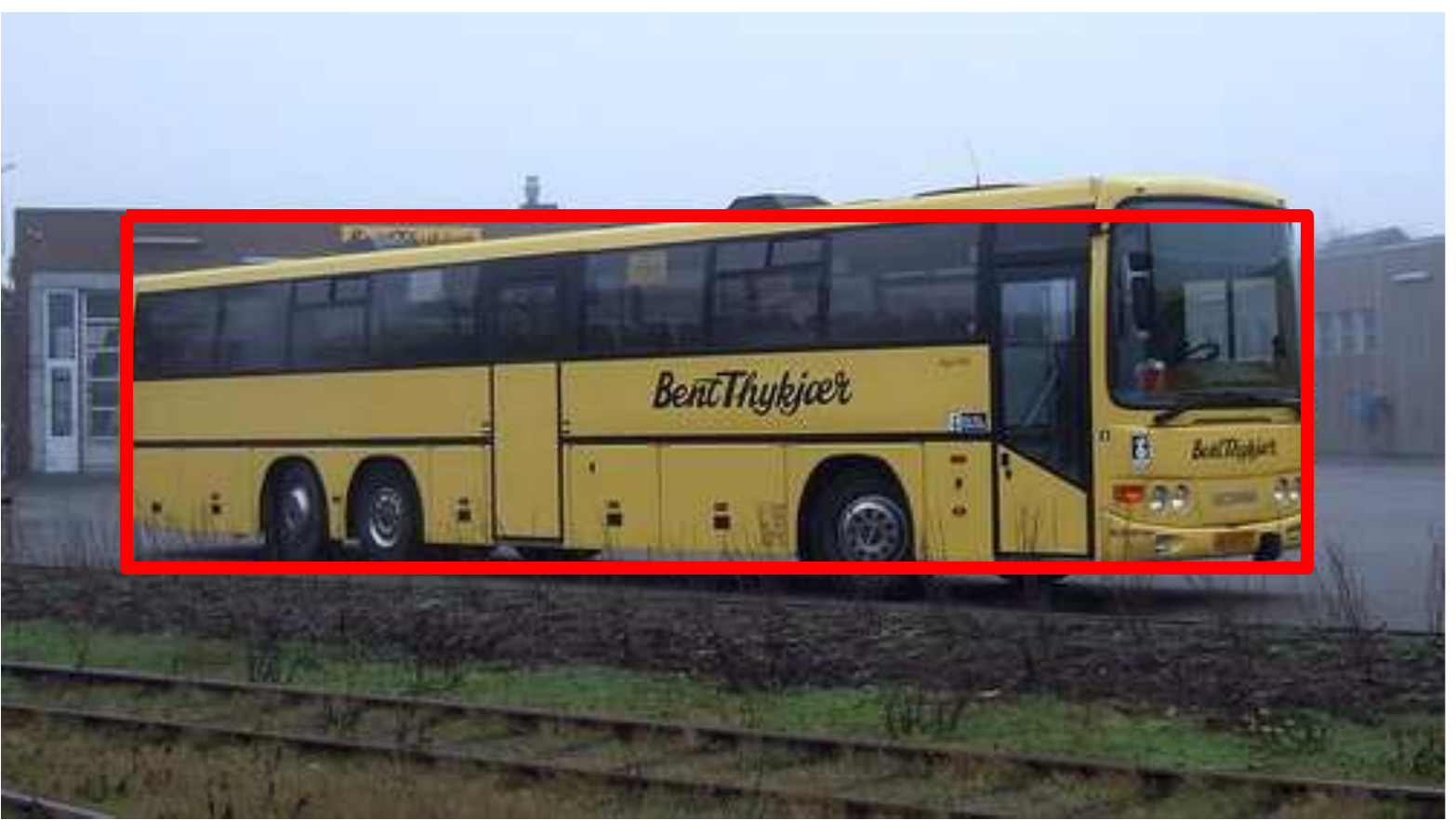}
\includegraphics[height=0.09\linewidth,width=0.09\linewidth]{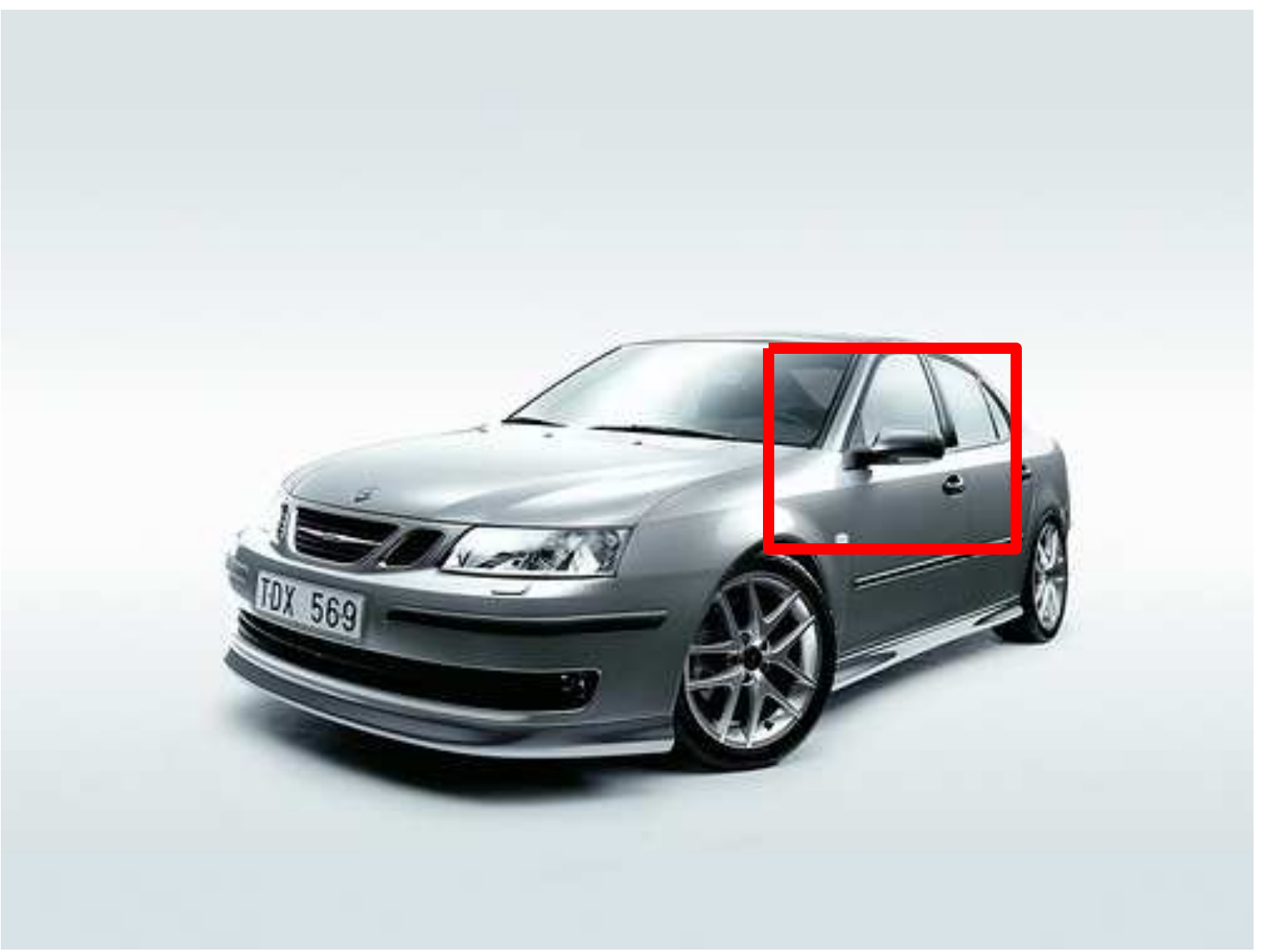}
\includegraphics[height=0.09\linewidth,width=0.09\linewidth]{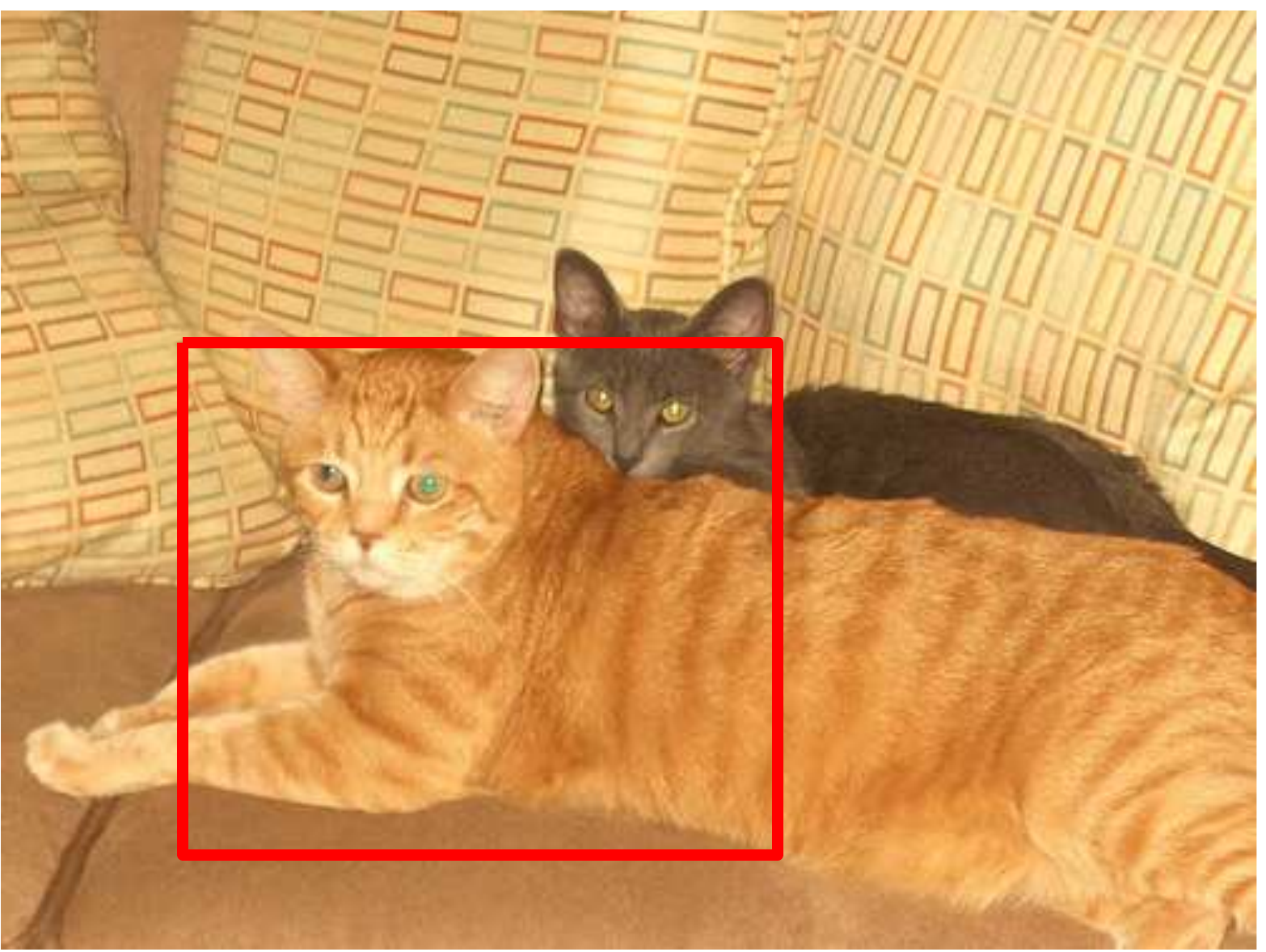}
\includegraphics[height=0.09\linewidth,width=0.09\linewidth]{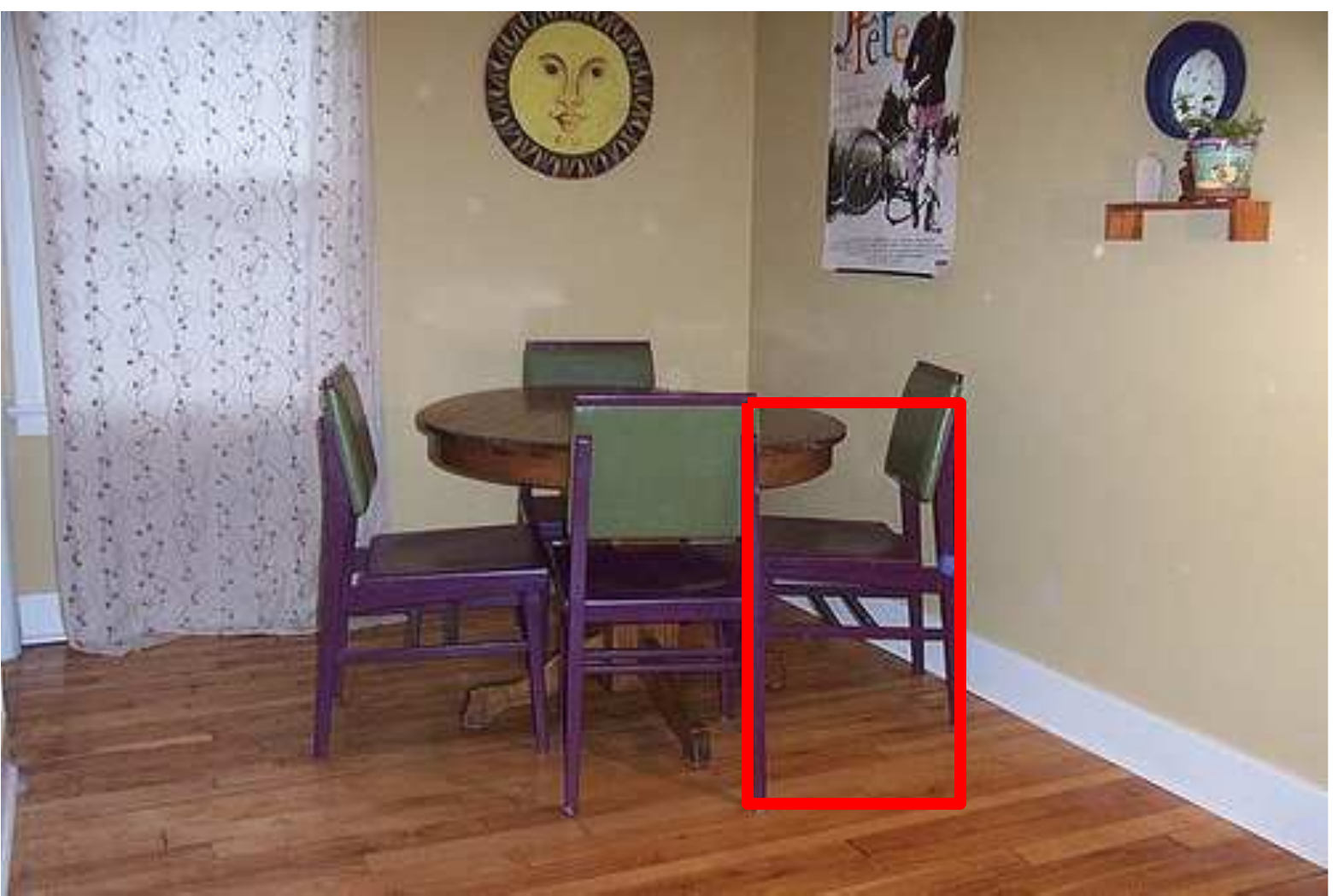}
\includegraphics[height=0.09\linewidth,width=0.09\linewidth]{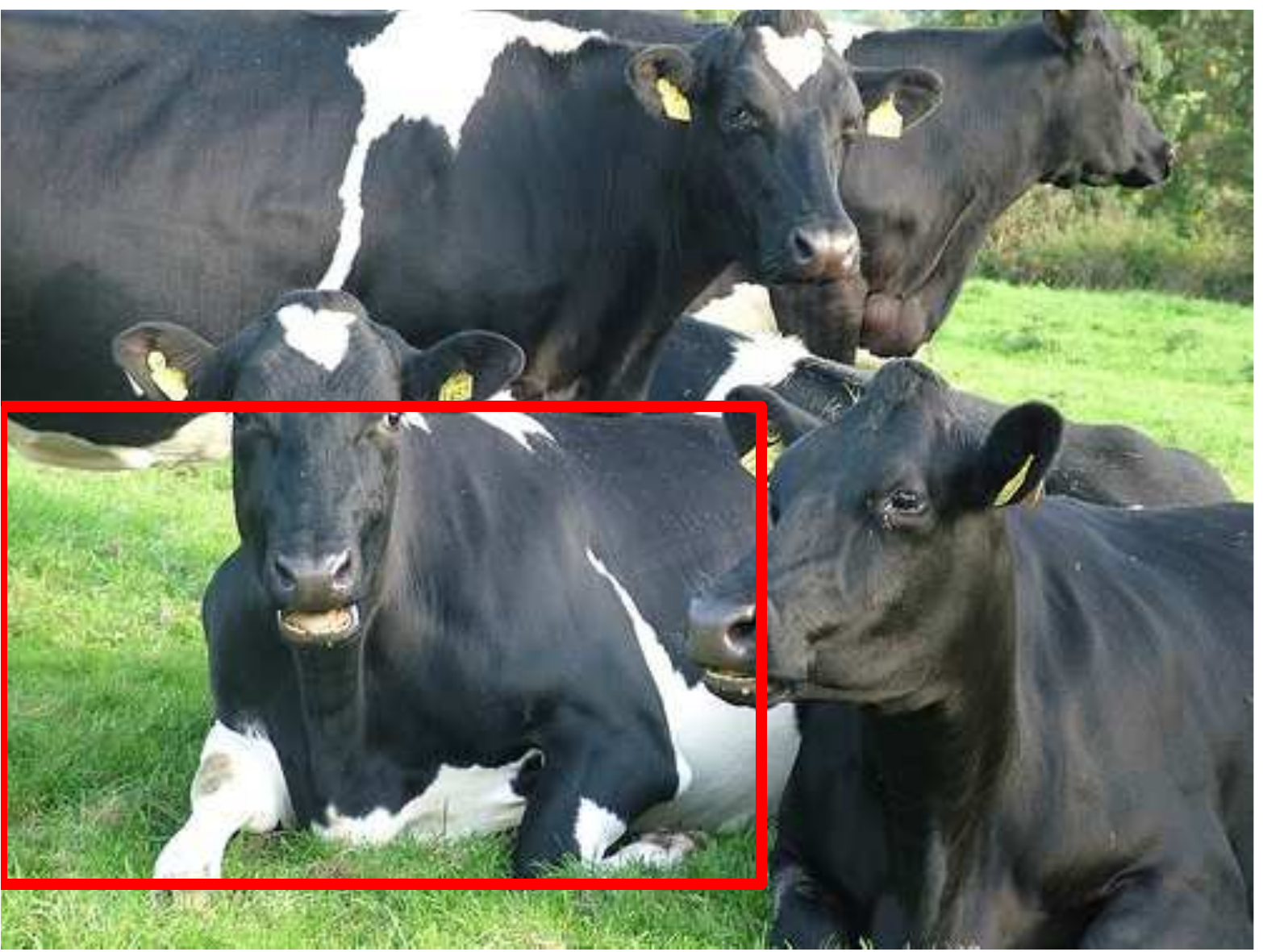}\\
\includegraphics[height=0.09\linewidth,width=0.09\linewidth]{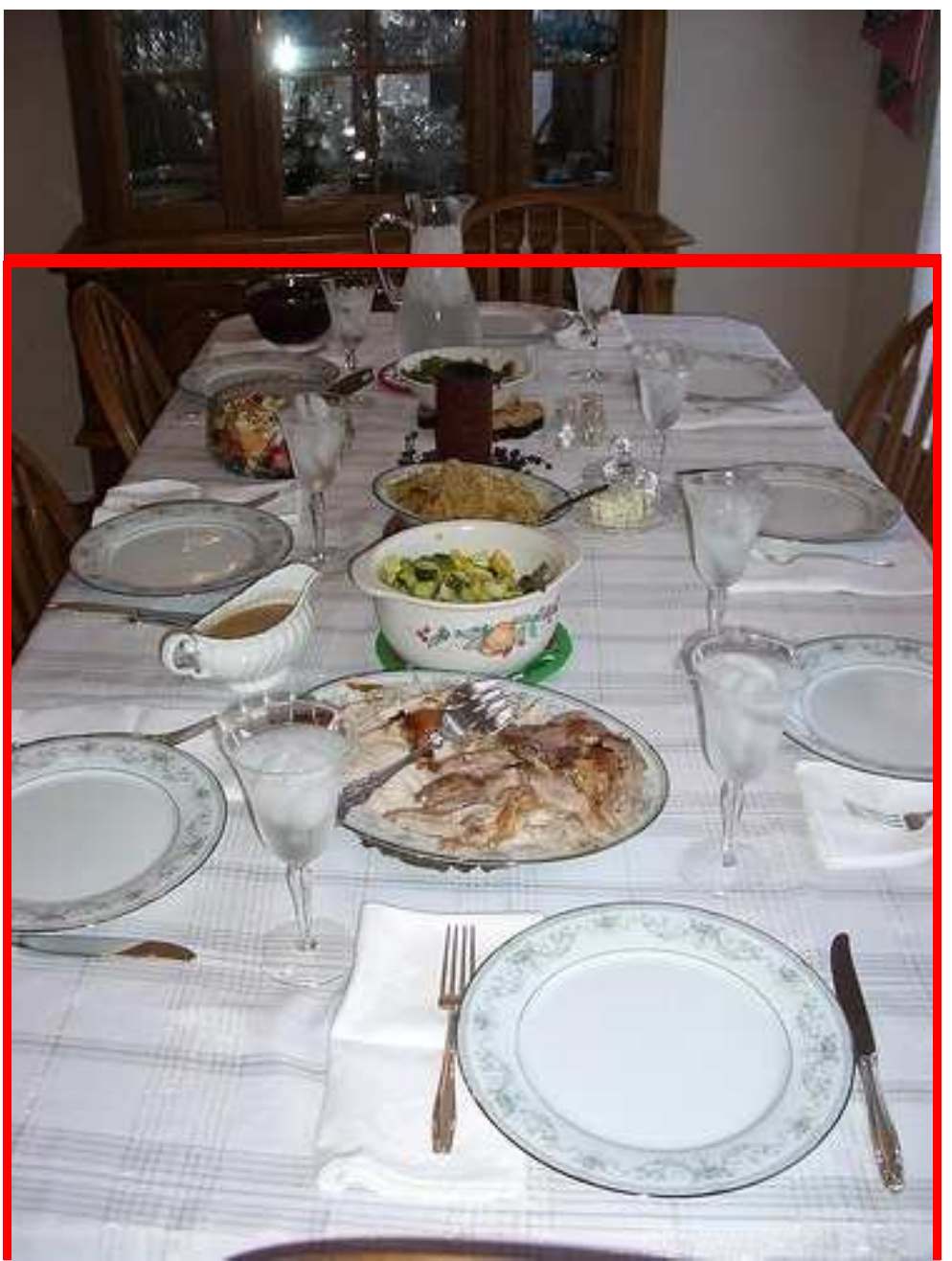}
\includegraphics[height=0.09\linewidth,width=0.09\linewidth]{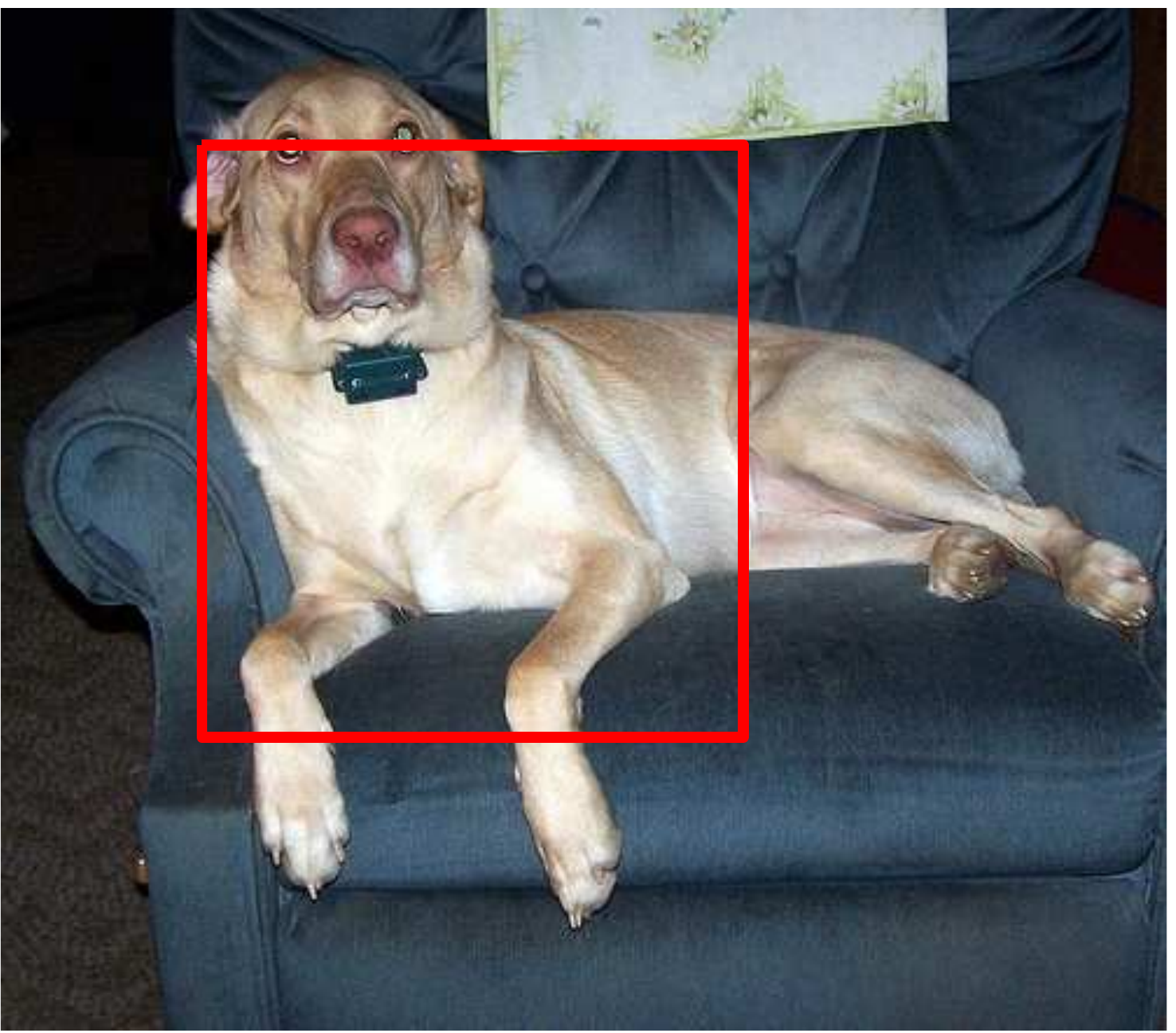}
\includegraphics[height=0.09\linewidth,width=0.09\linewidth]{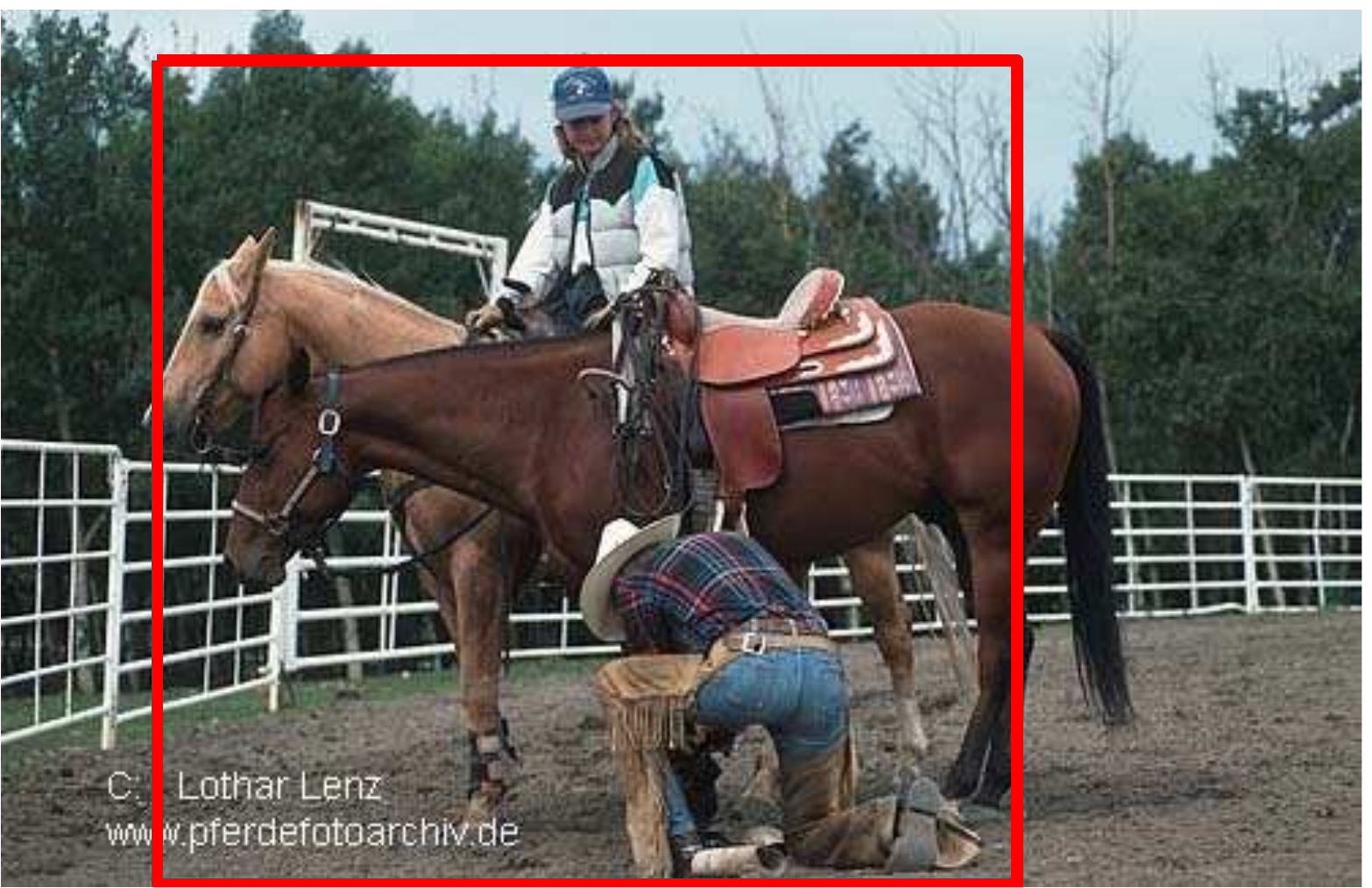}
\includegraphics[height=0.09\linewidth,width=0.09\linewidth]{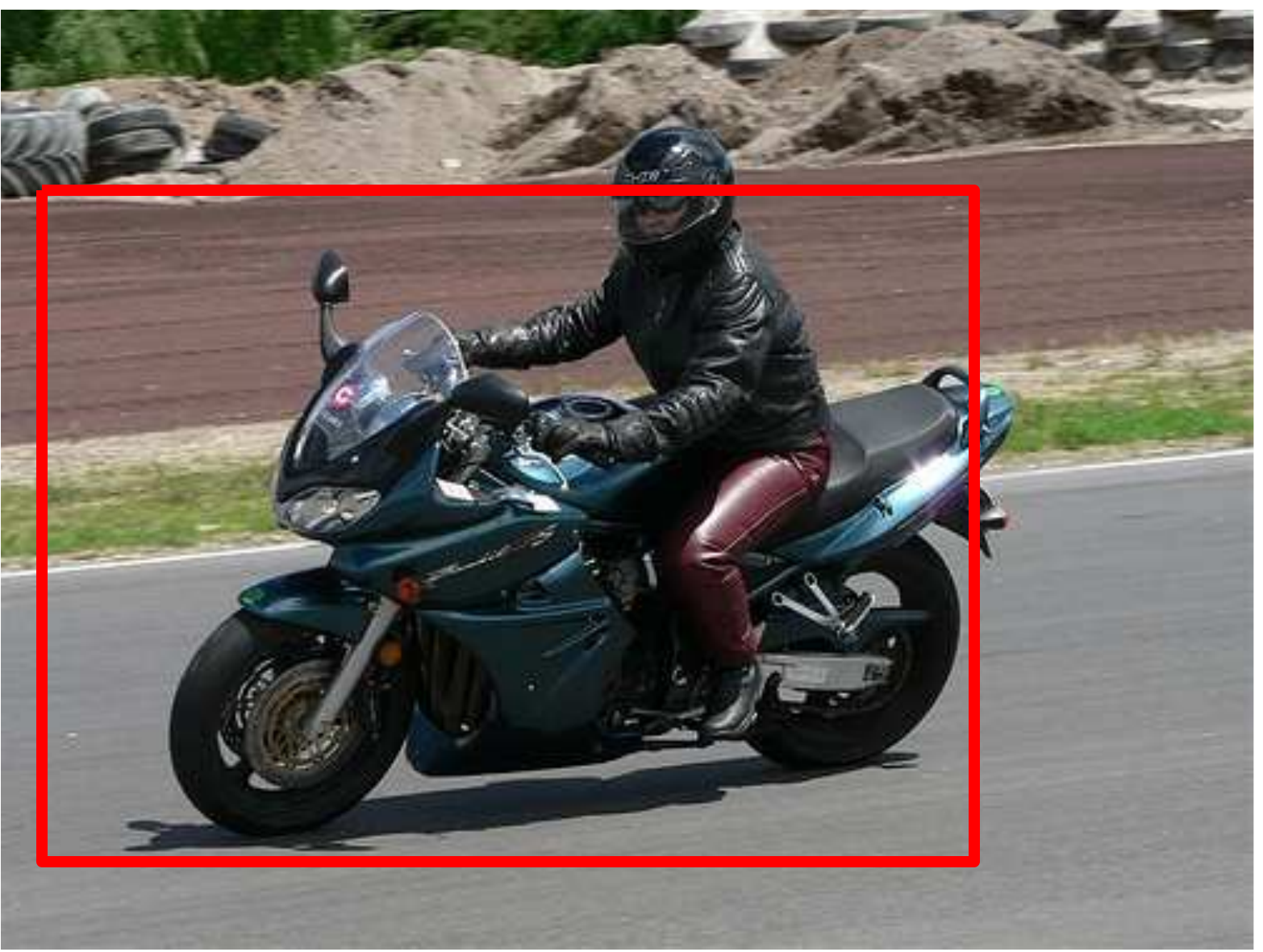}
\includegraphics[height=0.09\linewidth,width=0.09\linewidth]{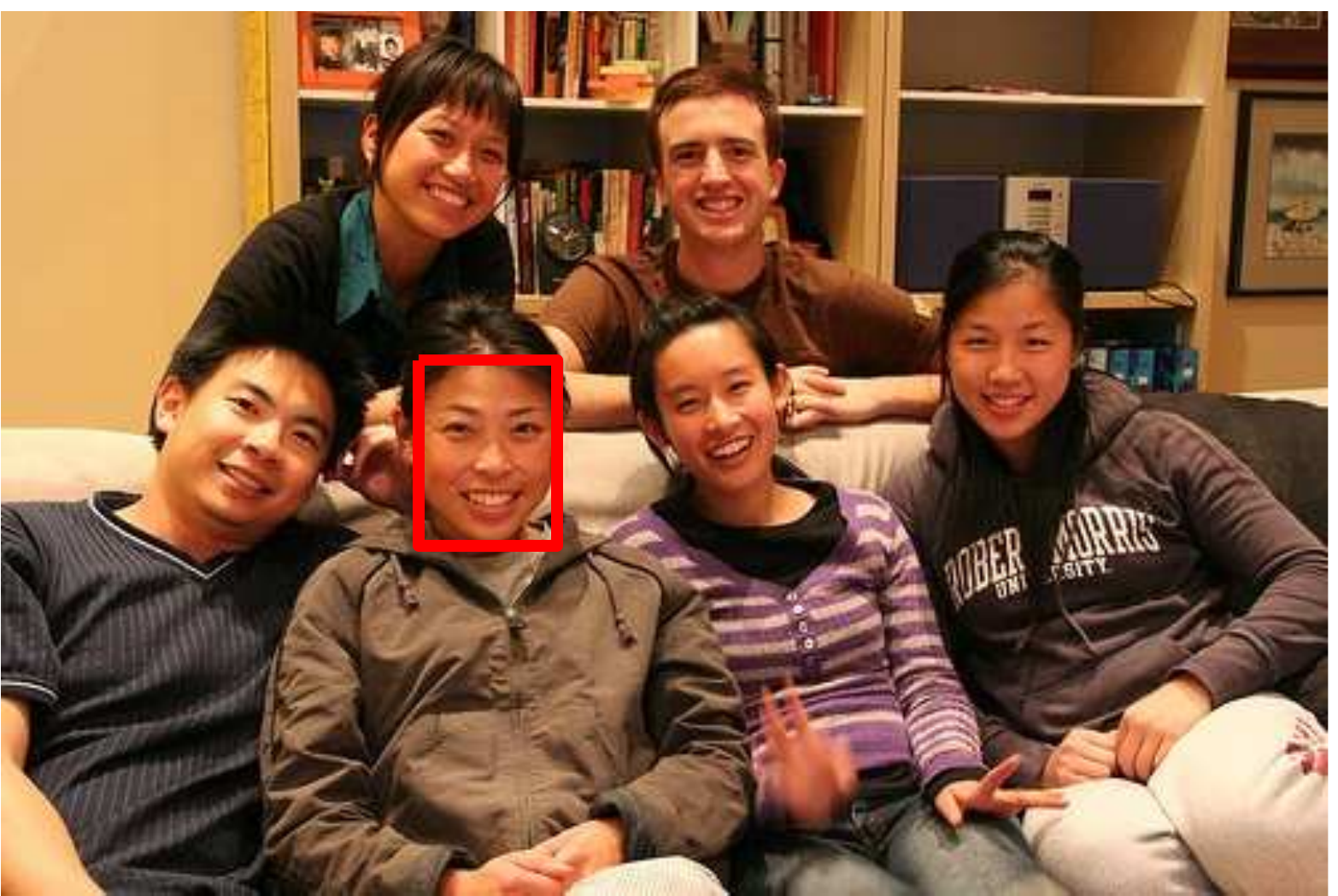}
\includegraphics[height=0.09\linewidth,width=0.09\linewidth]{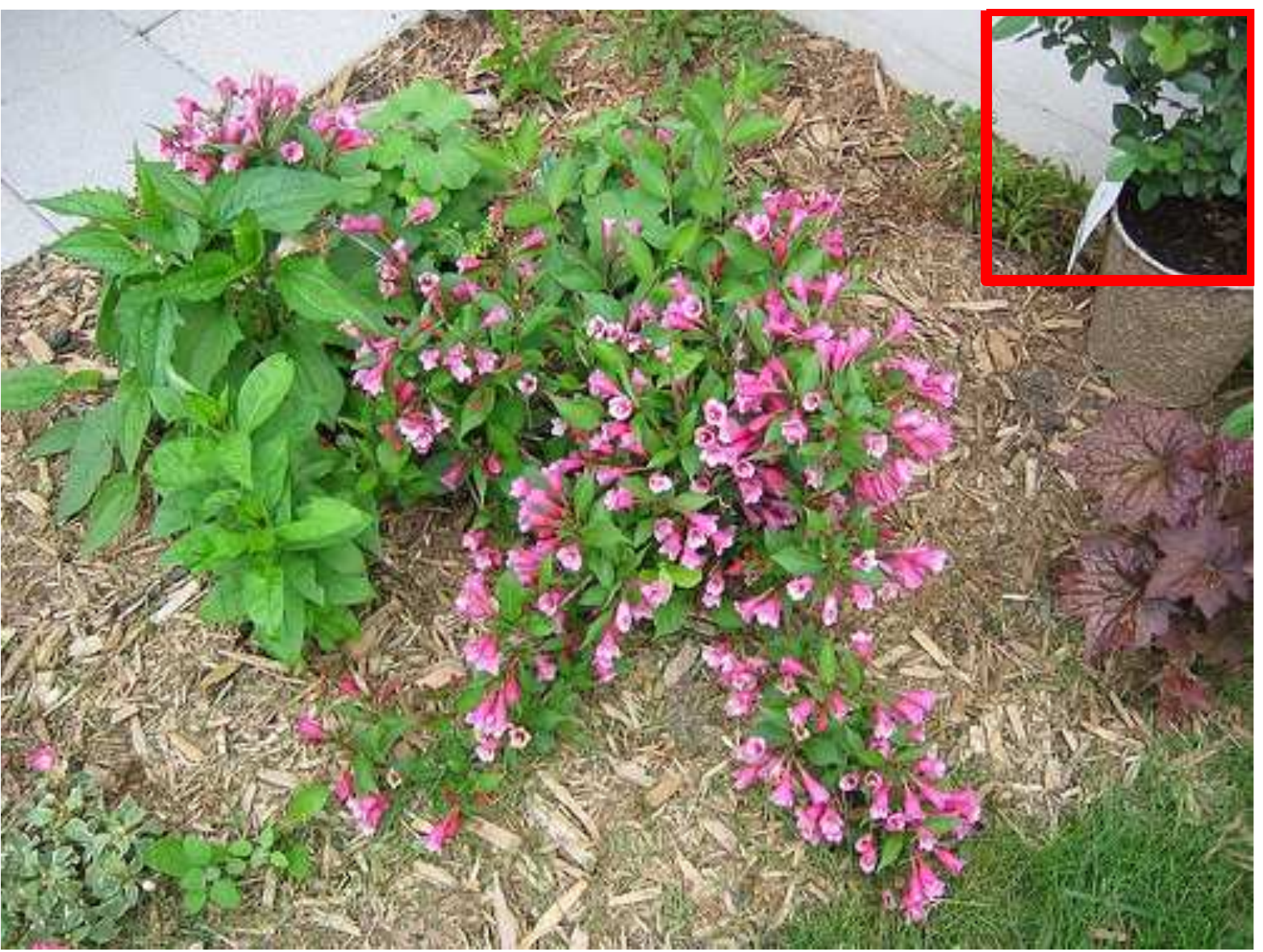}
\includegraphics[height=0.09\linewidth,width=0.09\linewidth]{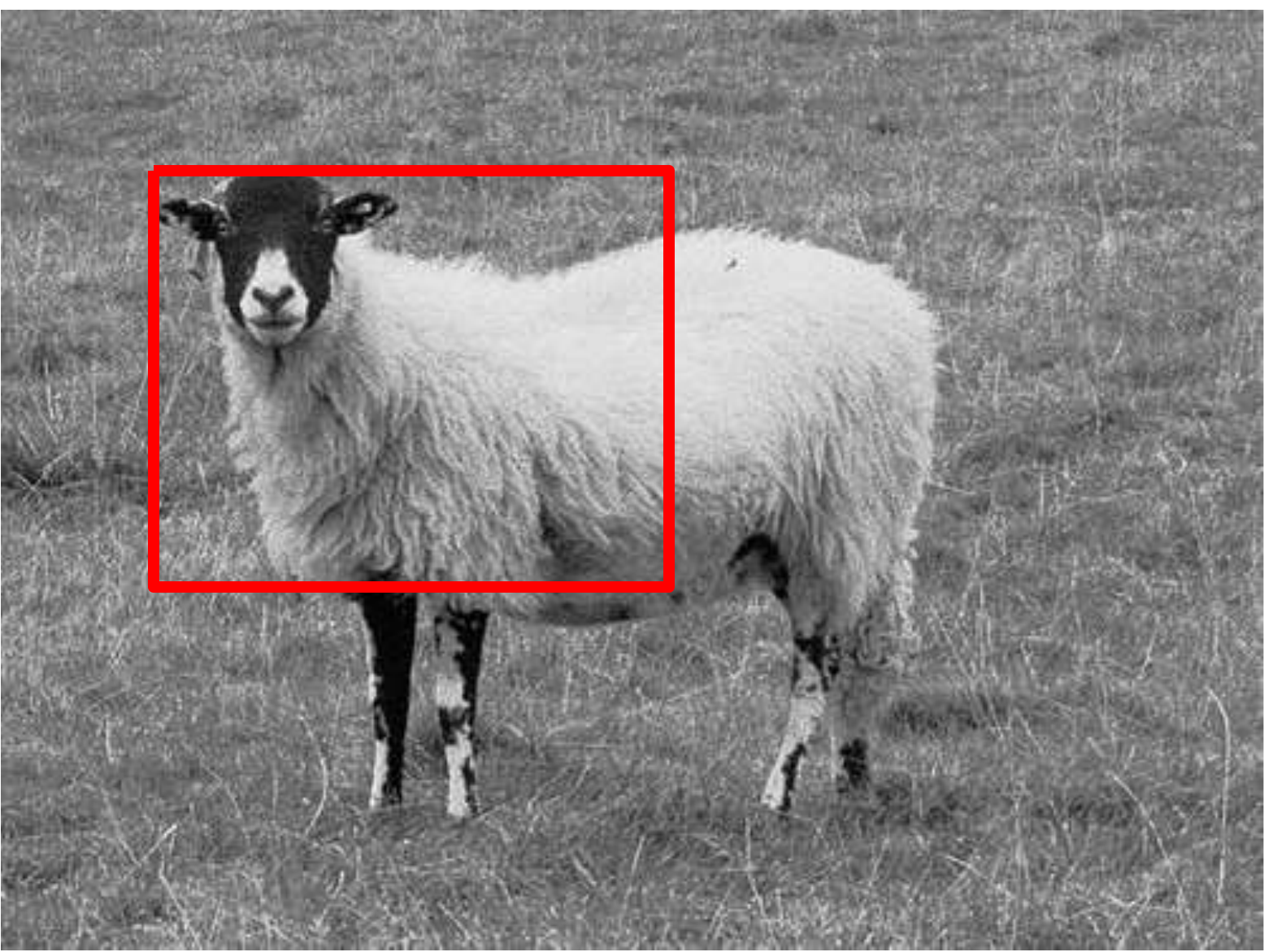}
\includegraphics[height=0.09\linewidth,width=0.09\linewidth]{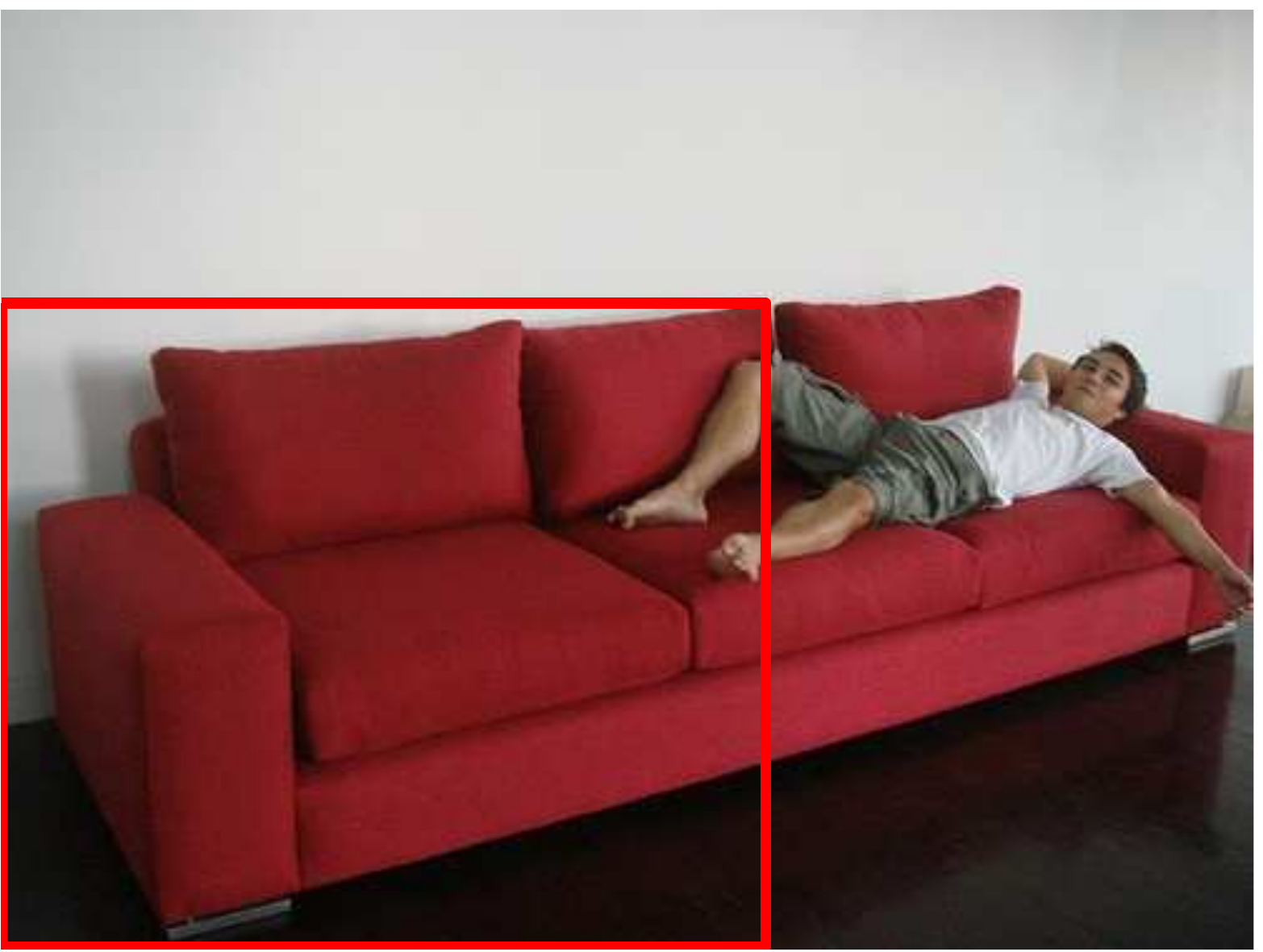}
\includegraphics[height=0.09\linewidth,width=0.09\linewidth]{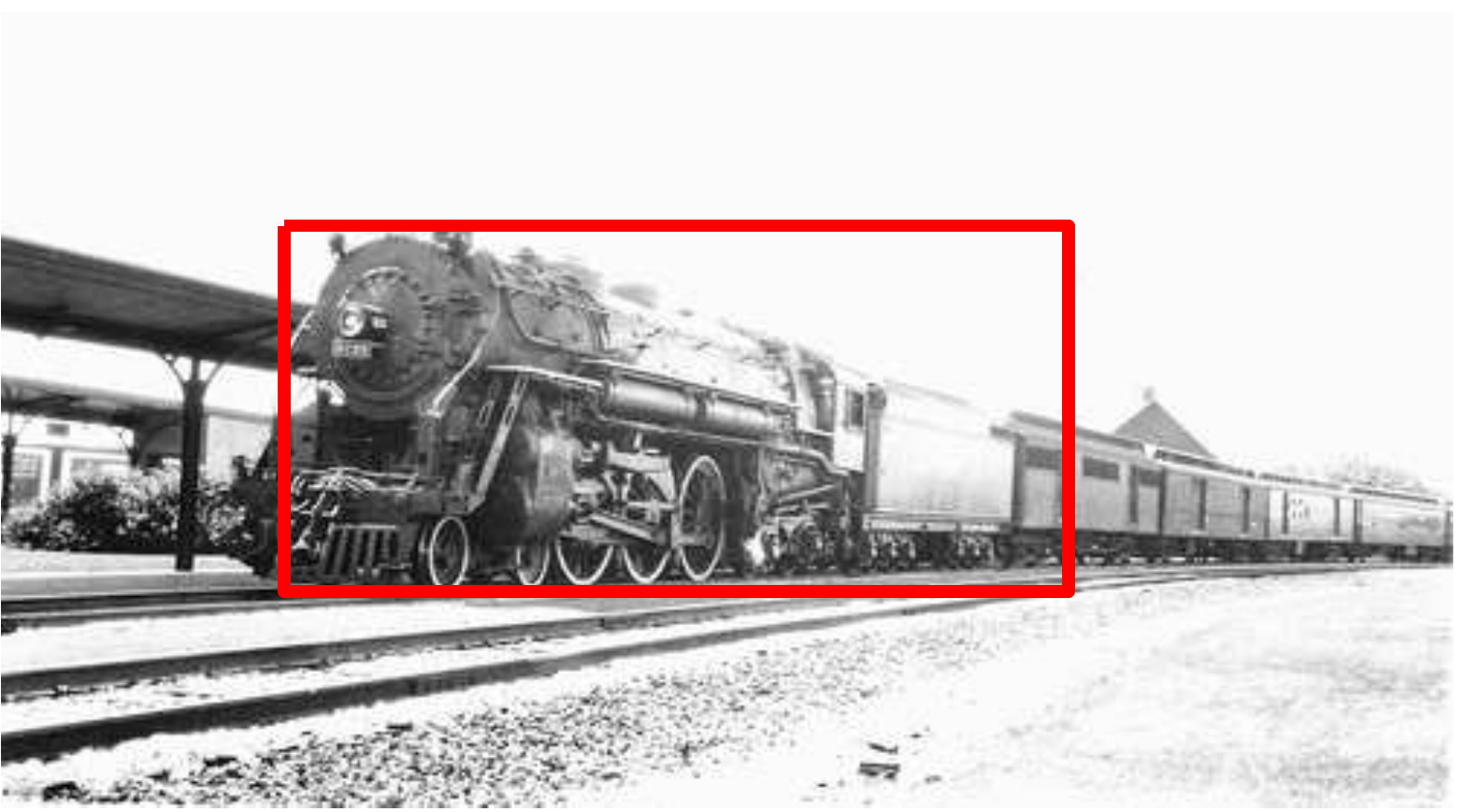}
\includegraphics[height=0.09\linewidth,width=0.09\linewidth]{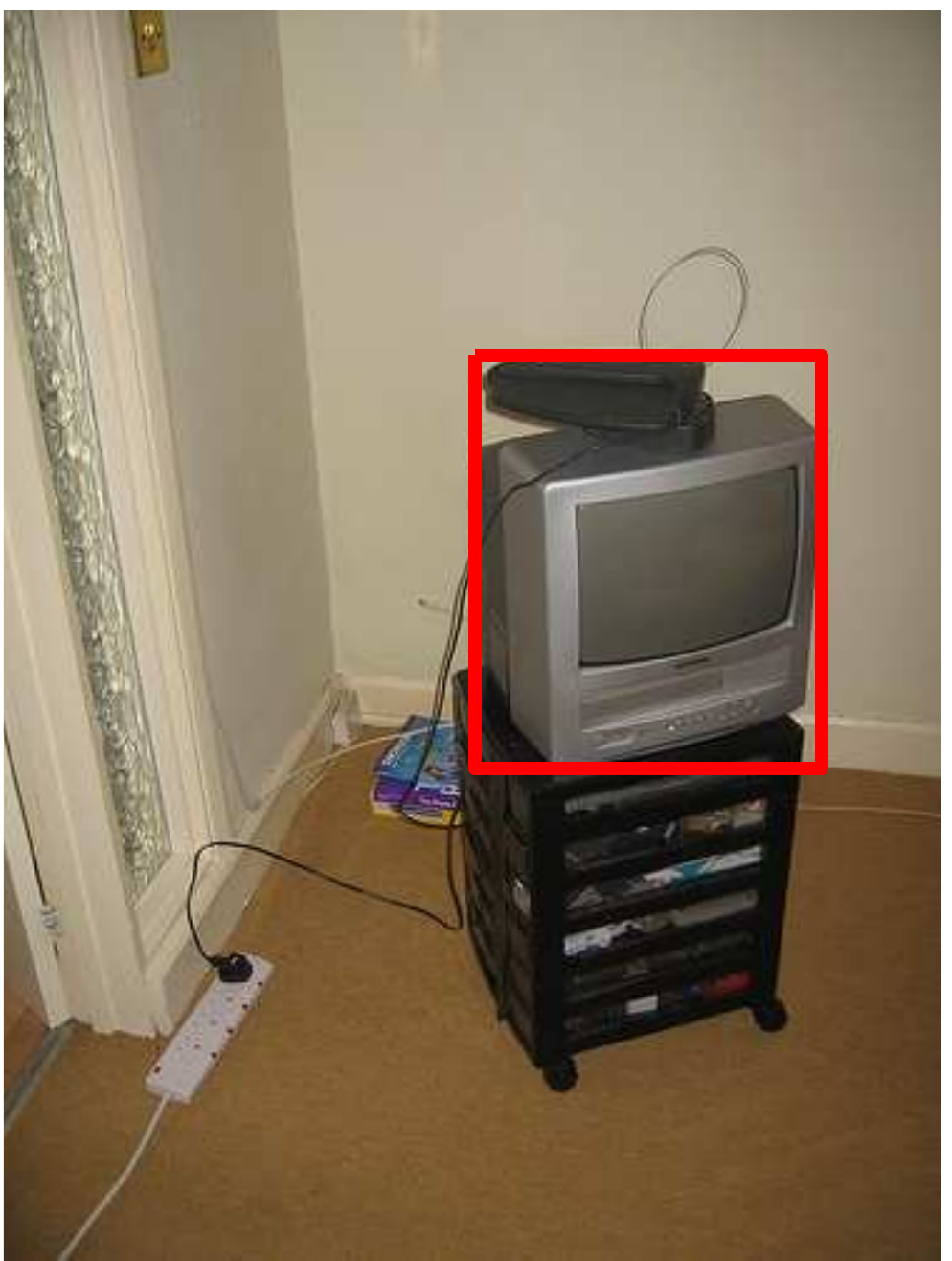}
\end{minipage}
}
\subfigure[Oxford 102 flowers]{
\centering
\begin{minipage}[b]{1.0\textwidth}
\centering
\includegraphics[height=0.09\linewidth,width=0.09\linewidth]{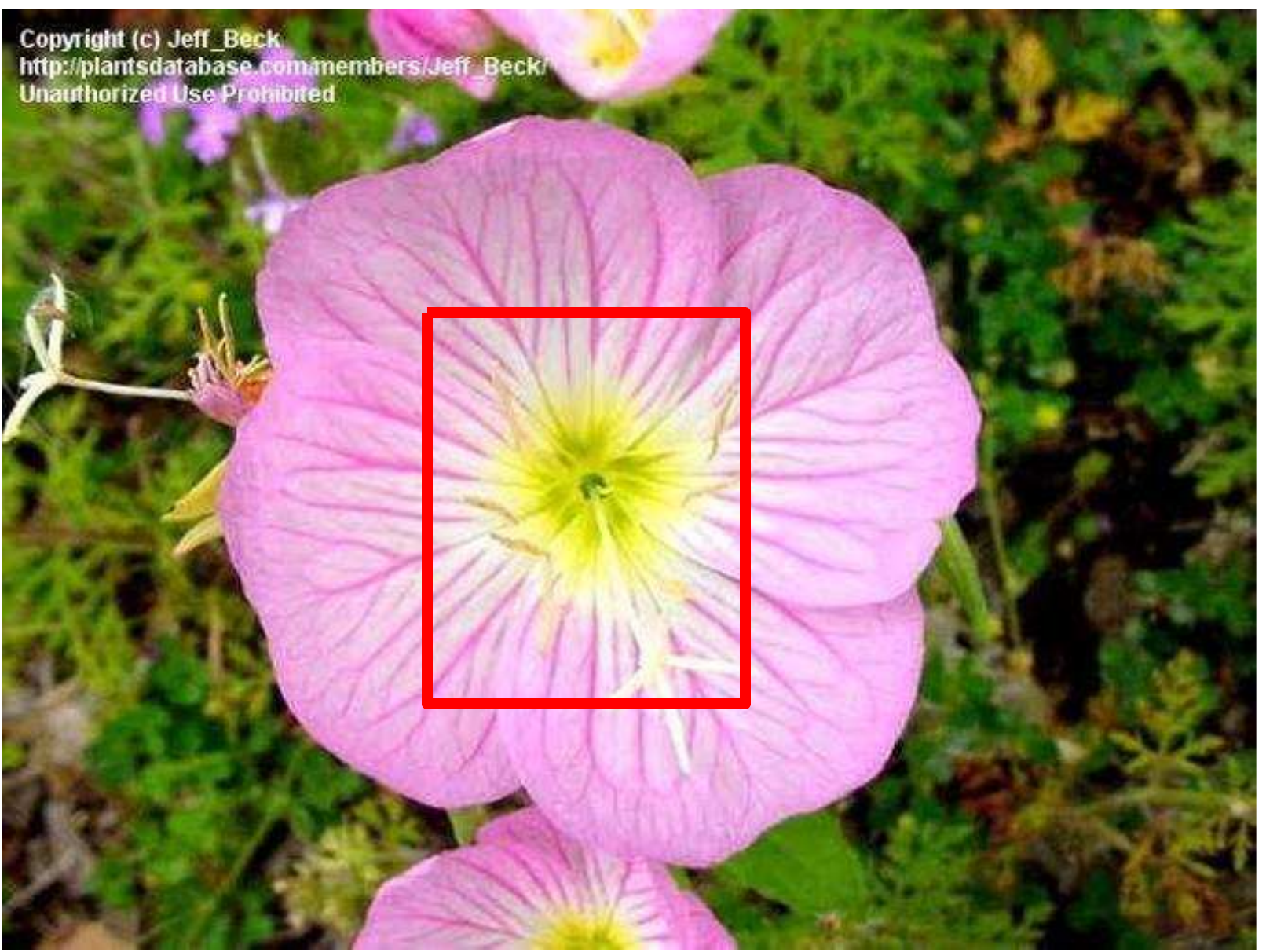}
\includegraphics[height=0.09\linewidth,width=0.09\linewidth]{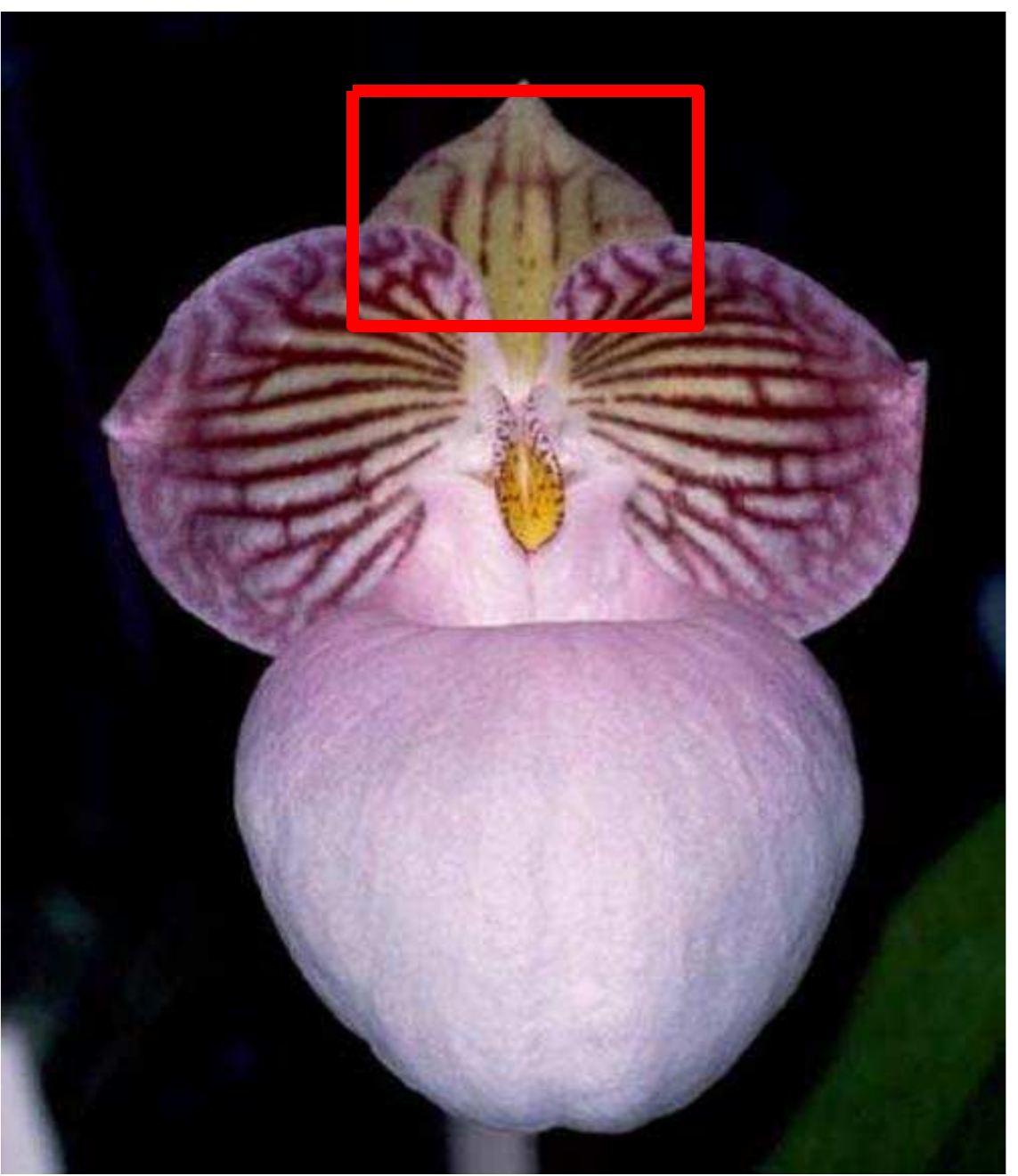}
\includegraphics[height=0.09\linewidth,width=0.09\linewidth]{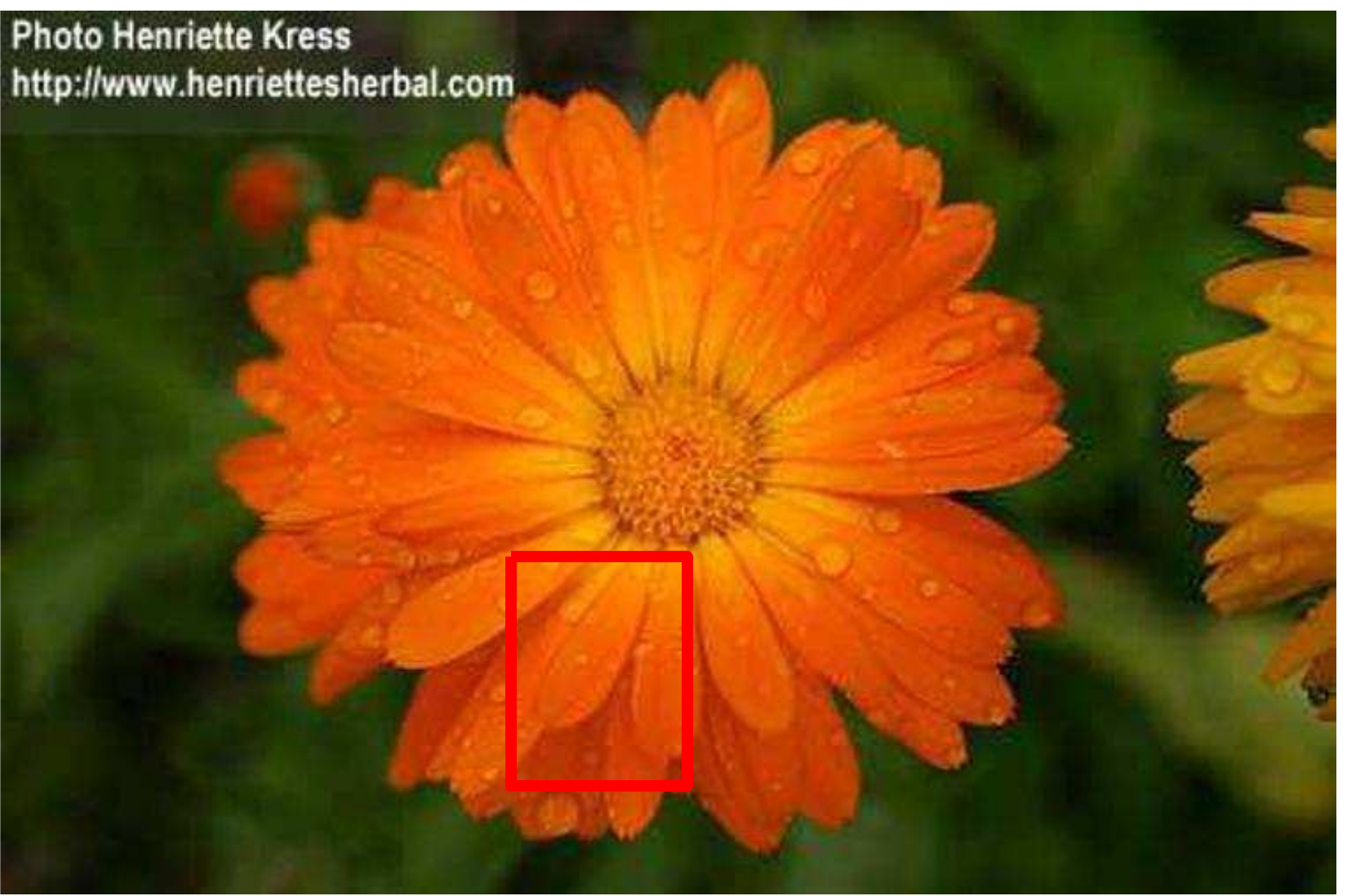}
\includegraphics[height=0.09\linewidth,width=0.09\linewidth]{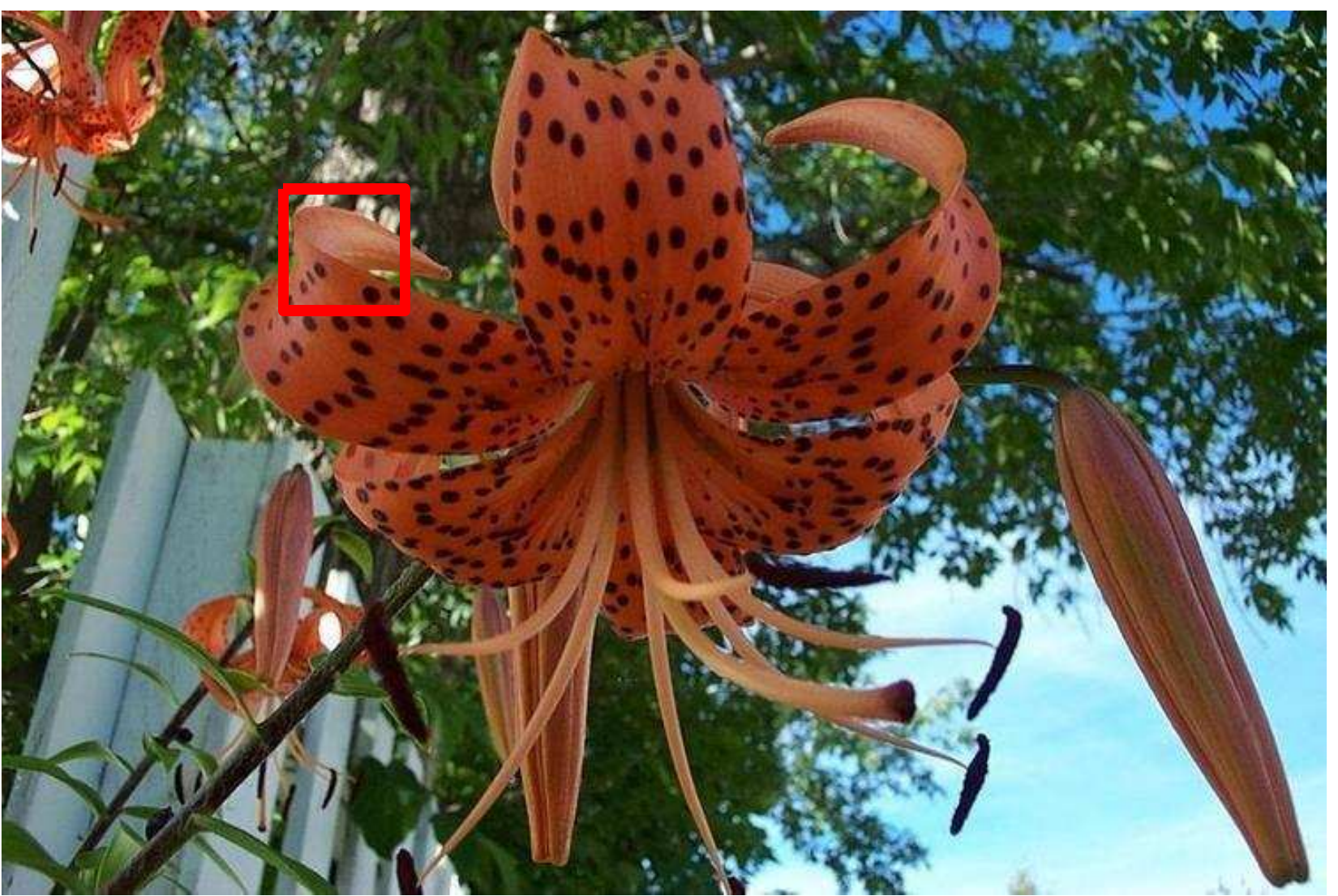}
\includegraphics[height=0.09\linewidth,width=0.09\linewidth]{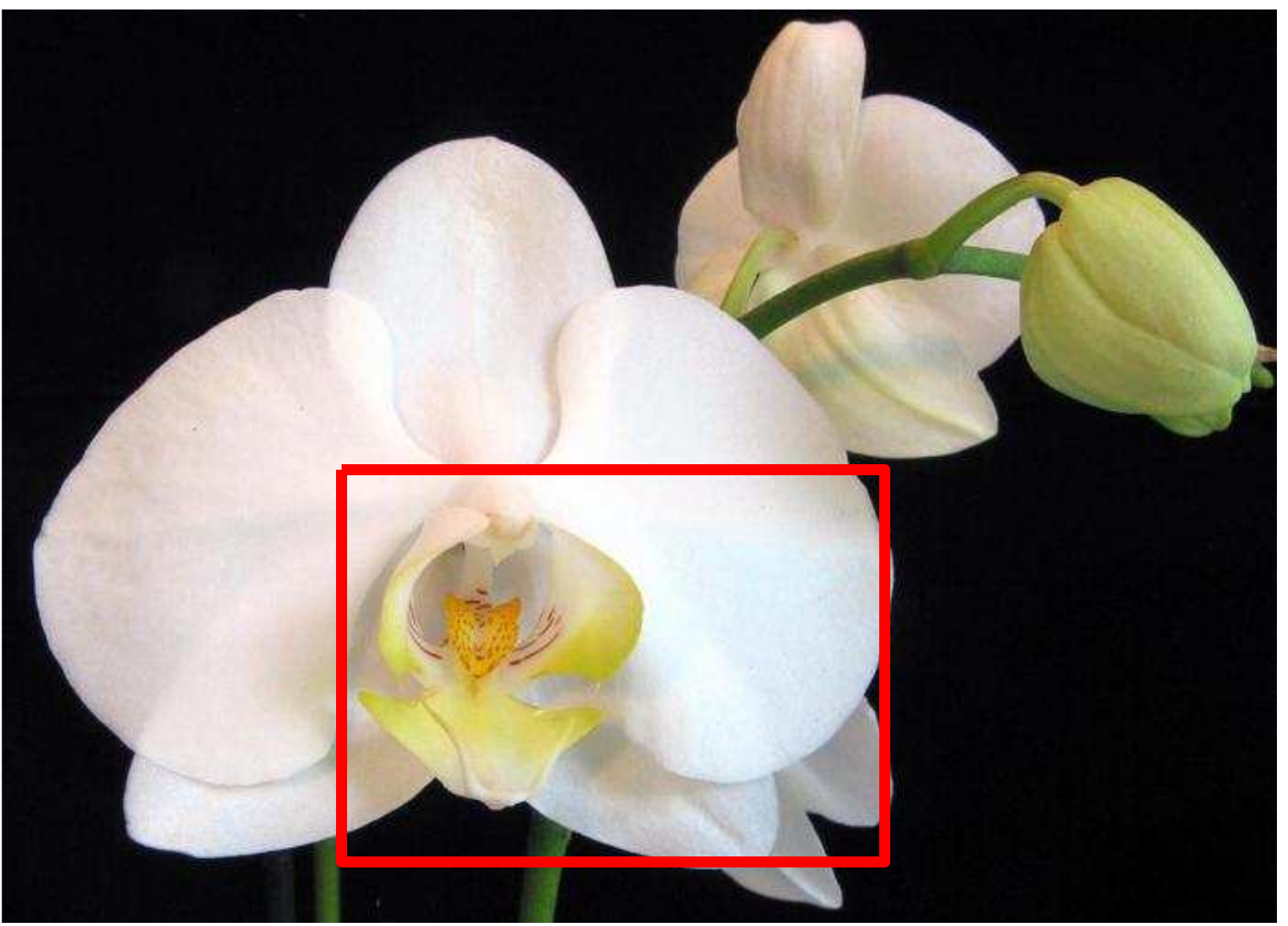}
\includegraphics[height=0.09\linewidth,width=0.09\linewidth]{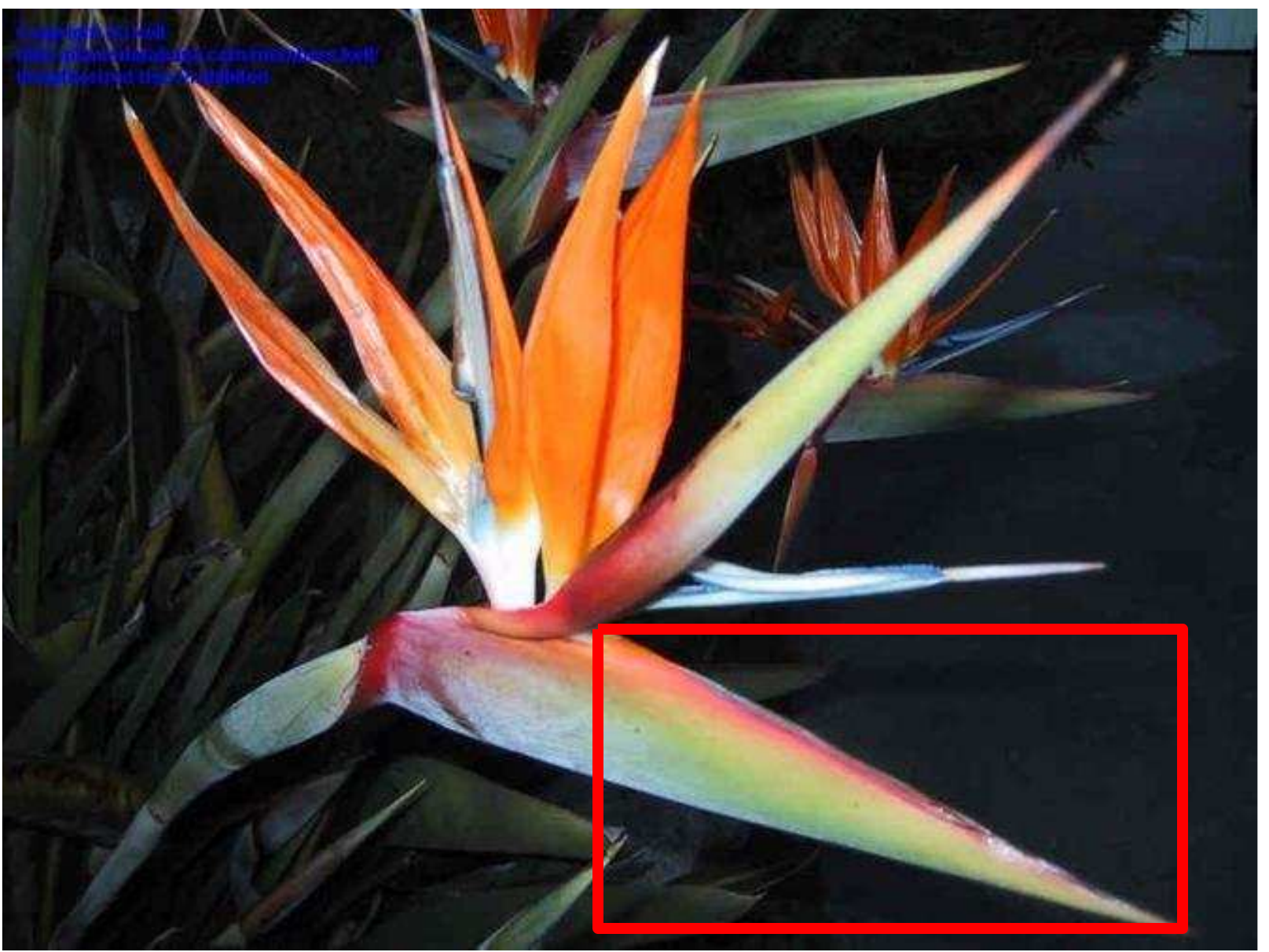}
\includegraphics[height=0.09\linewidth,width=0.09\linewidth]{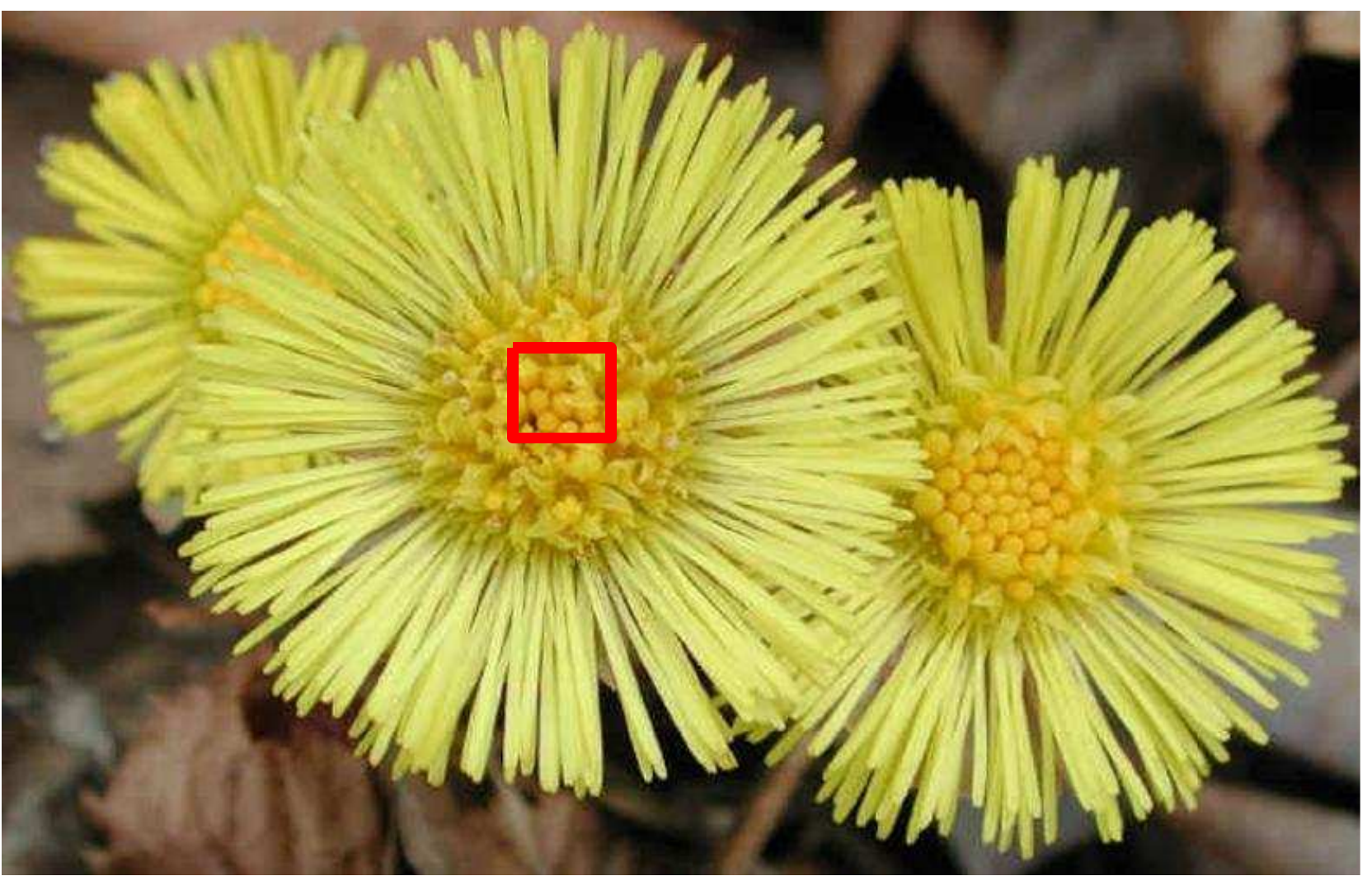}
\includegraphics[height=0.09\linewidth,width=0.09\linewidth]{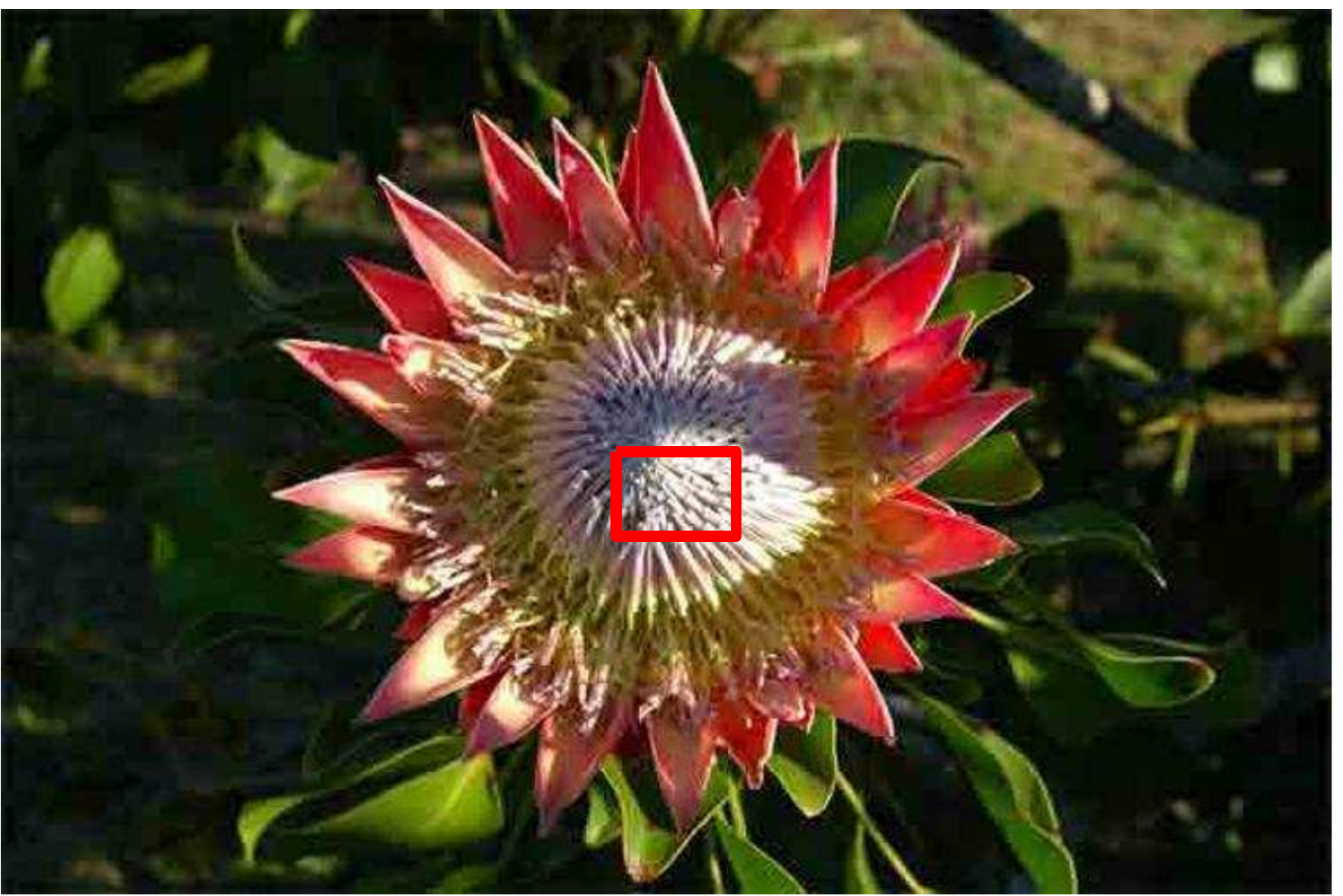}
\includegraphics[height=0.09\linewidth,width=0.09\linewidth]{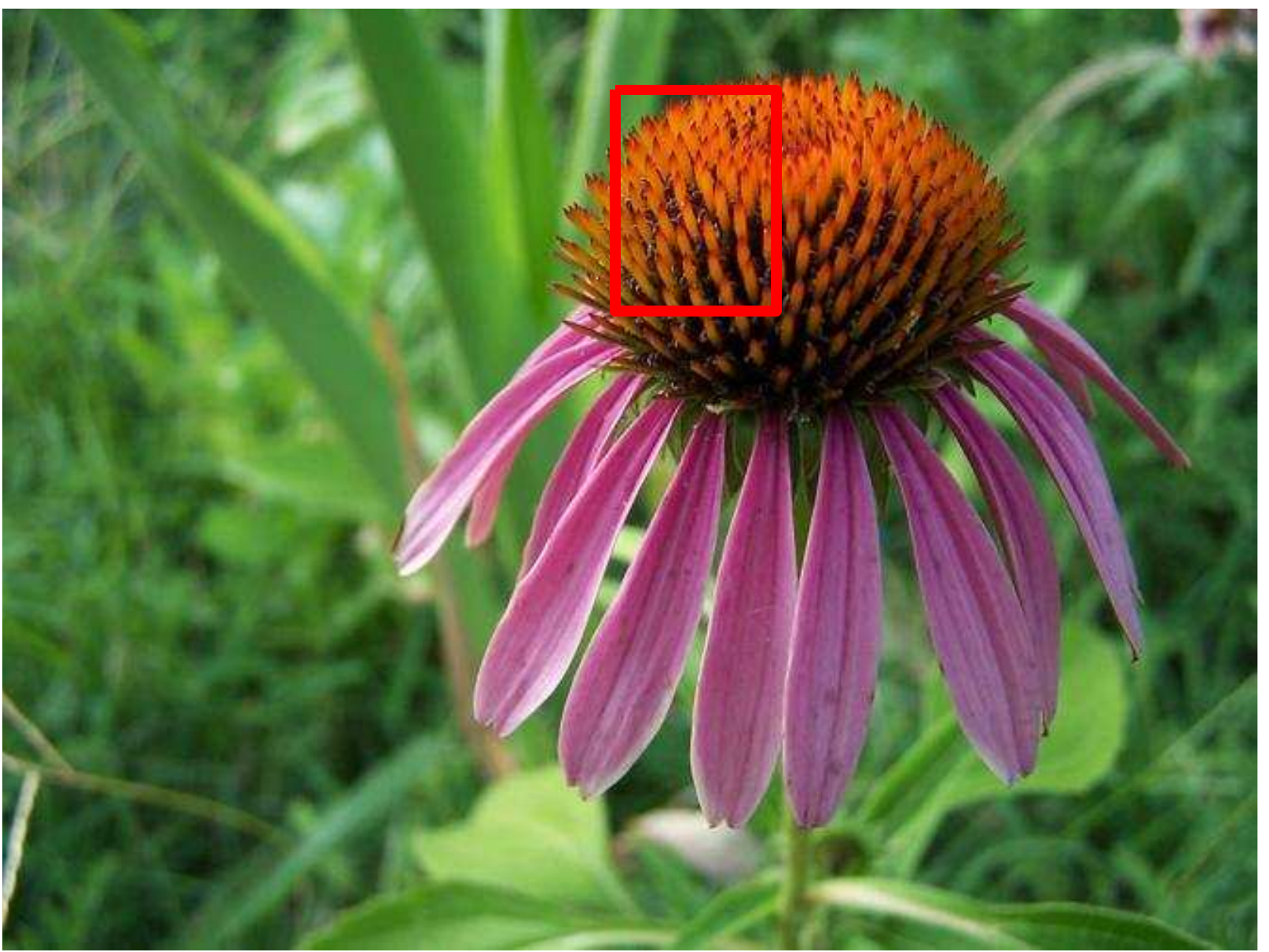}
\includegraphics[height=0.09\linewidth,width=0.09\linewidth]{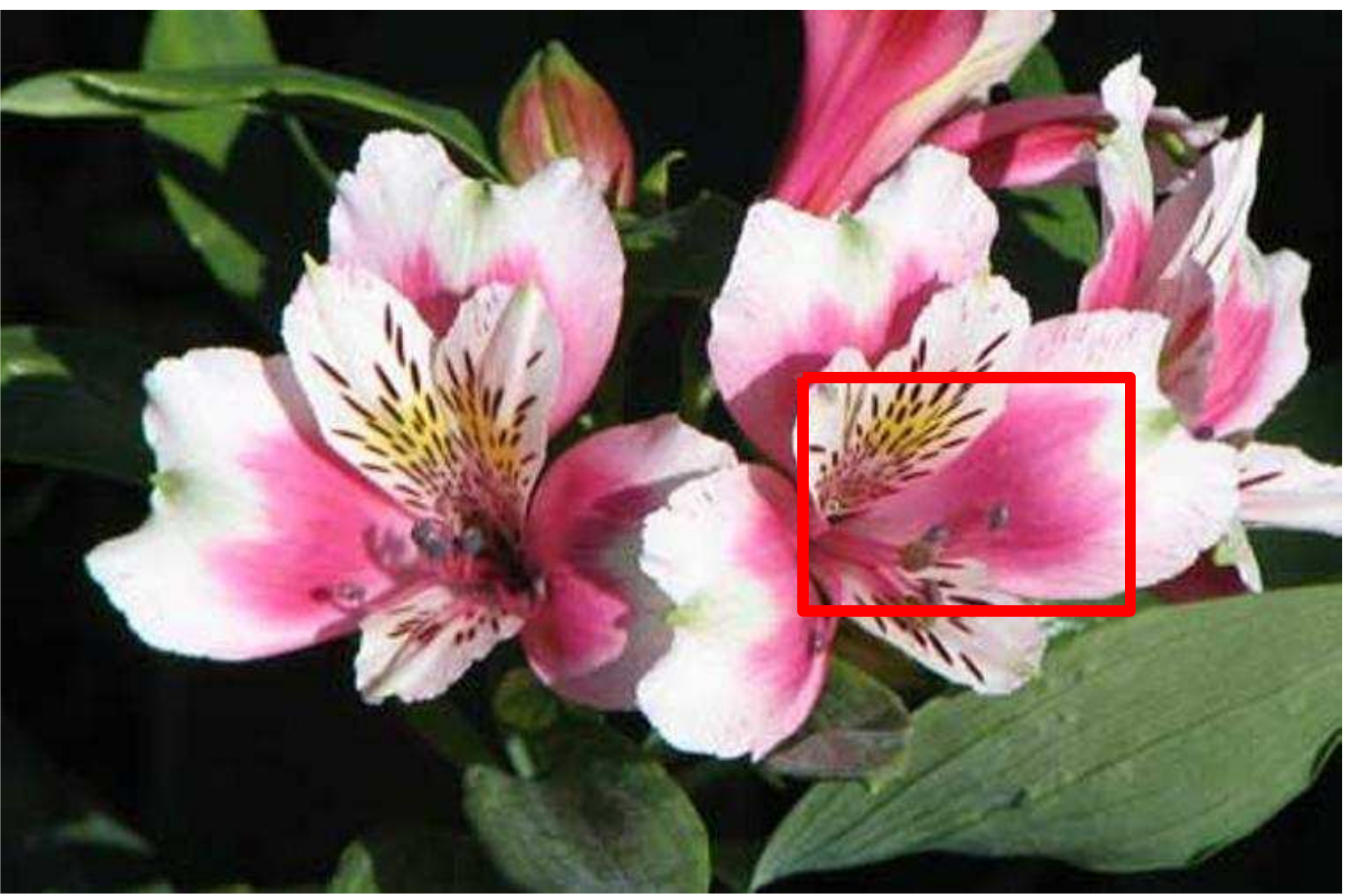}\\
\includegraphics[height=0.09\linewidth,width=0.09\linewidth]{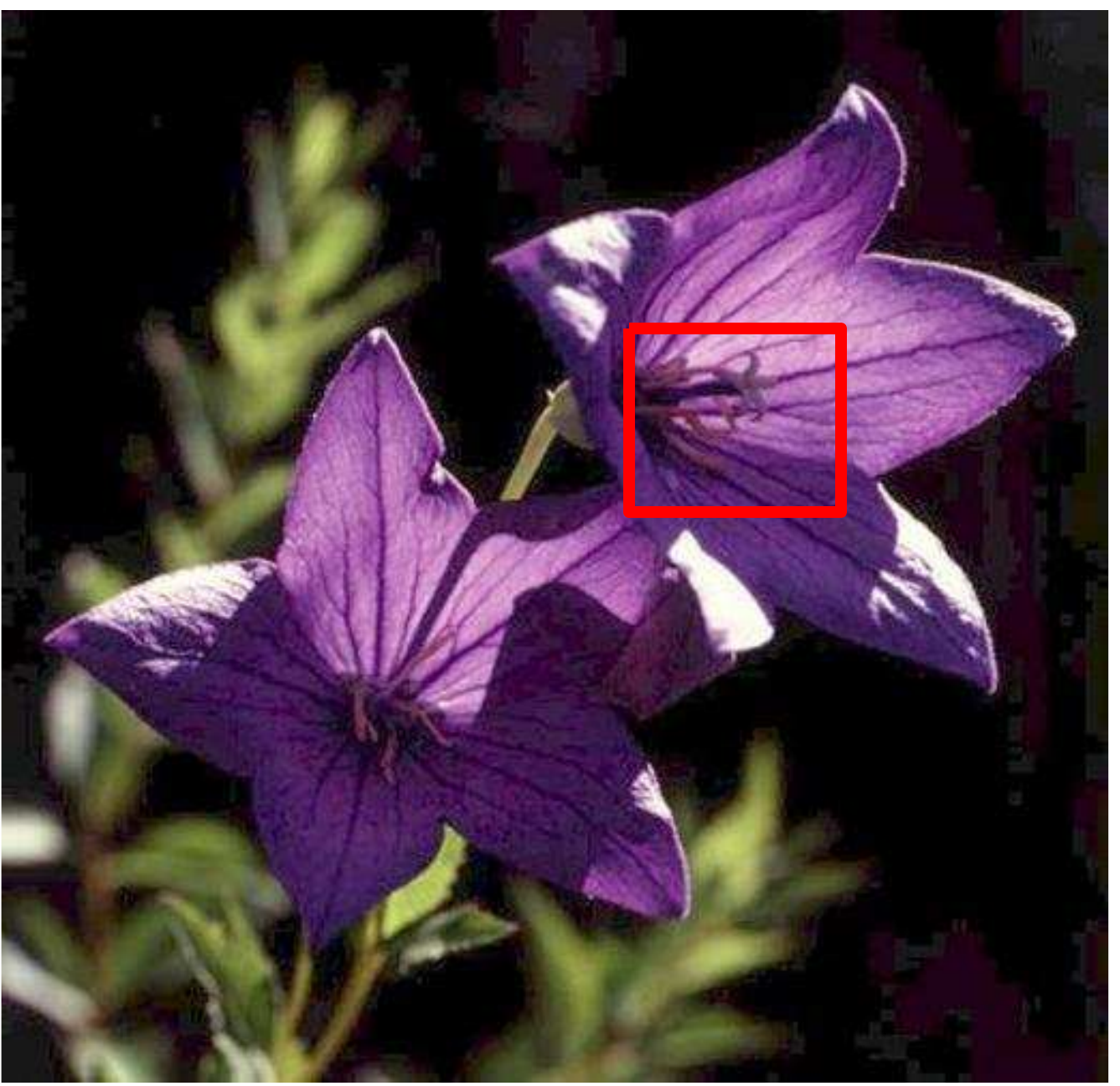}
\includegraphics[height=0.09\linewidth,width=0.09\linewidth]{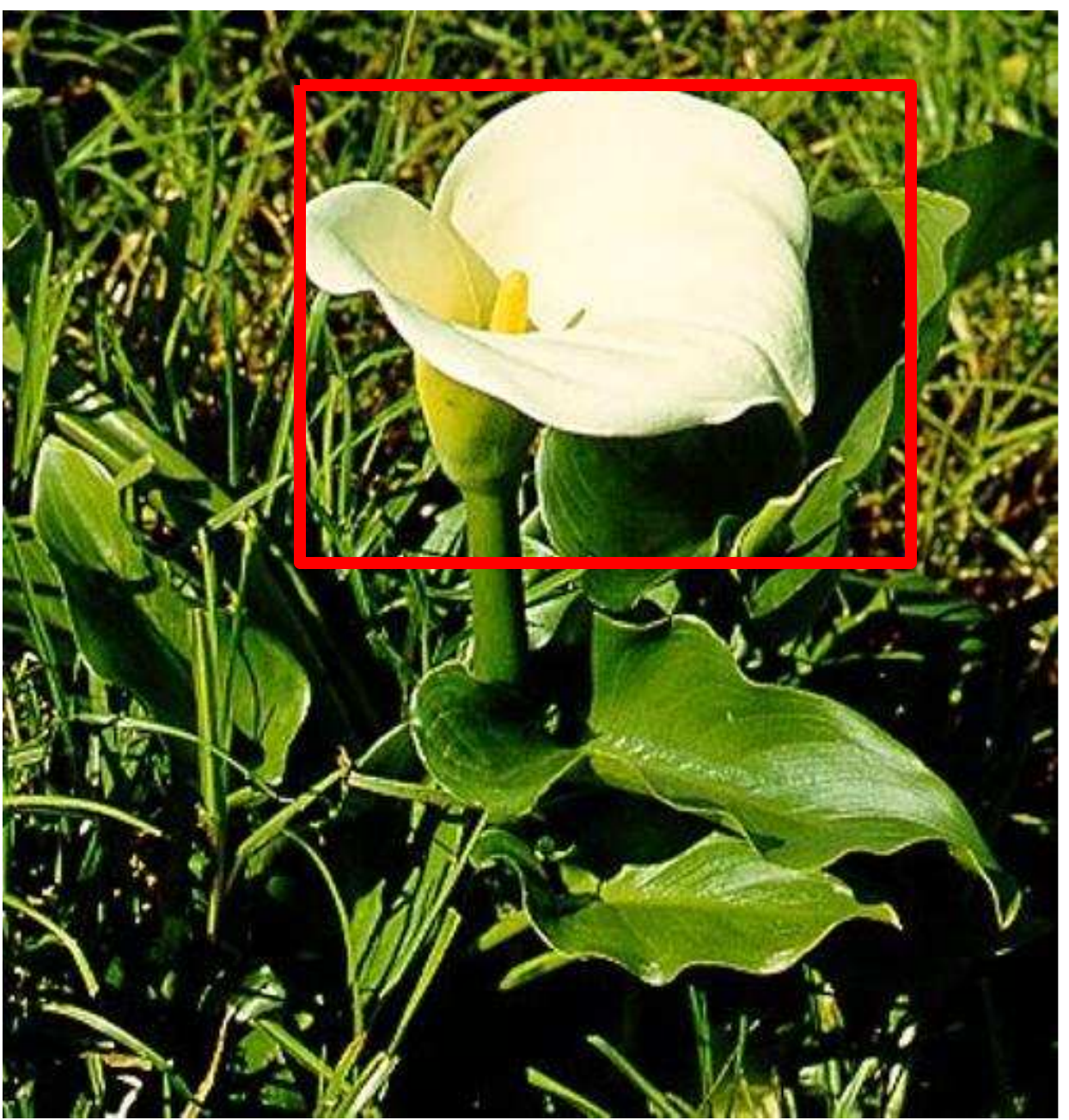}
\includegraphics[height=0.09\linewidth,width=0.09\linewidth]{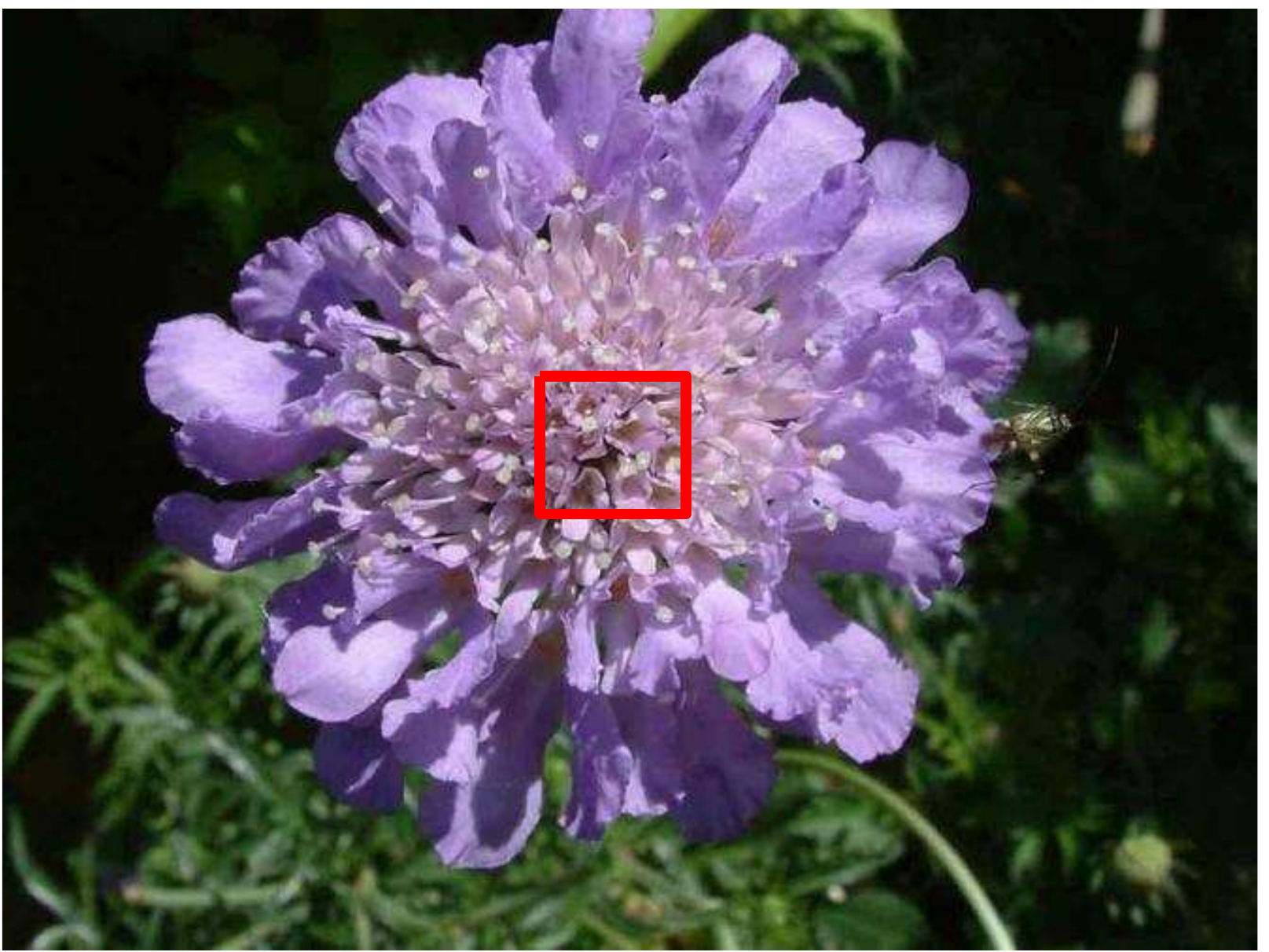}
\includegraphics[height=0.09\linewidth,width=0.09\linewidth]{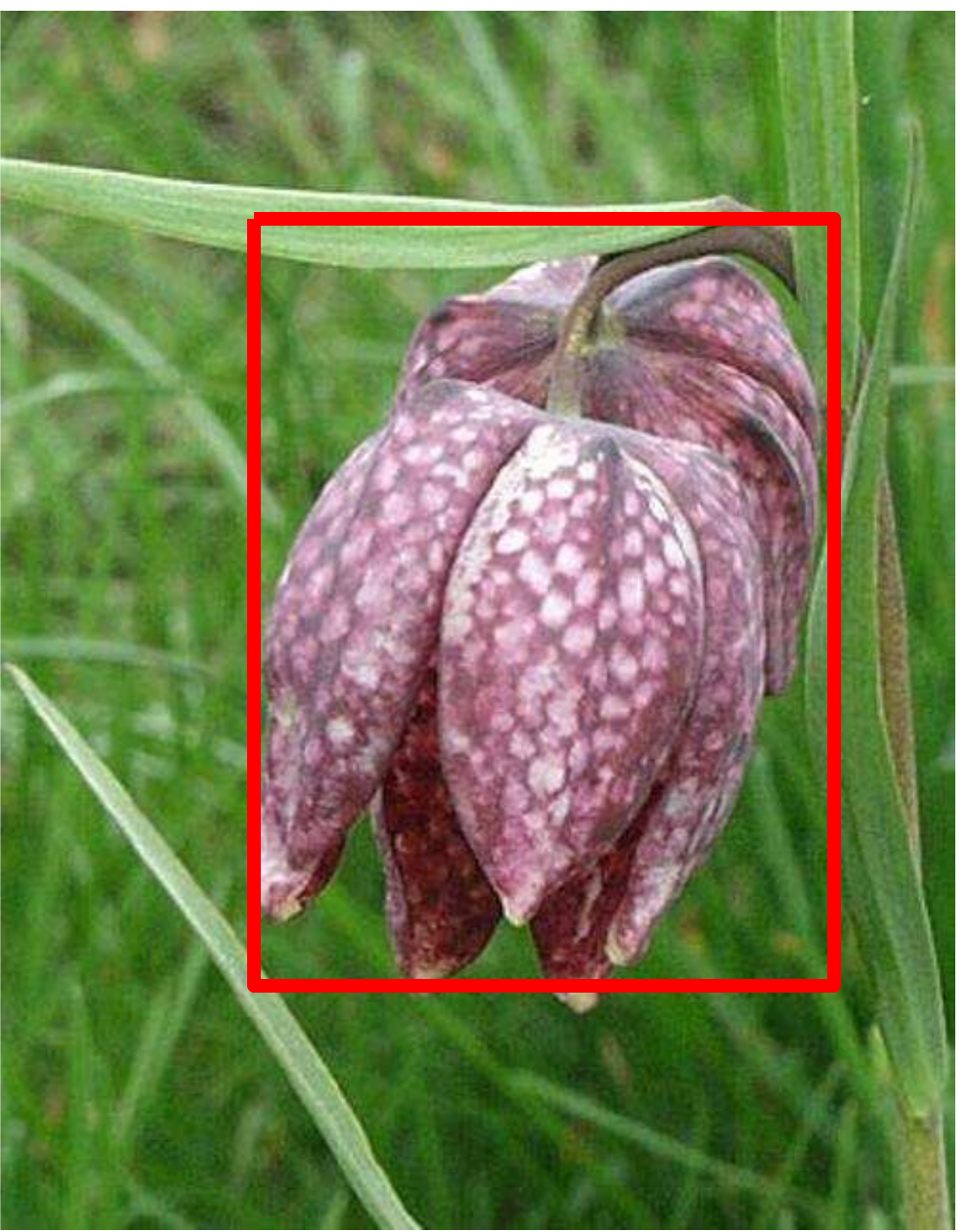}
\includegraphics[height=0.09\linewidth,width=0.09\linewidth]{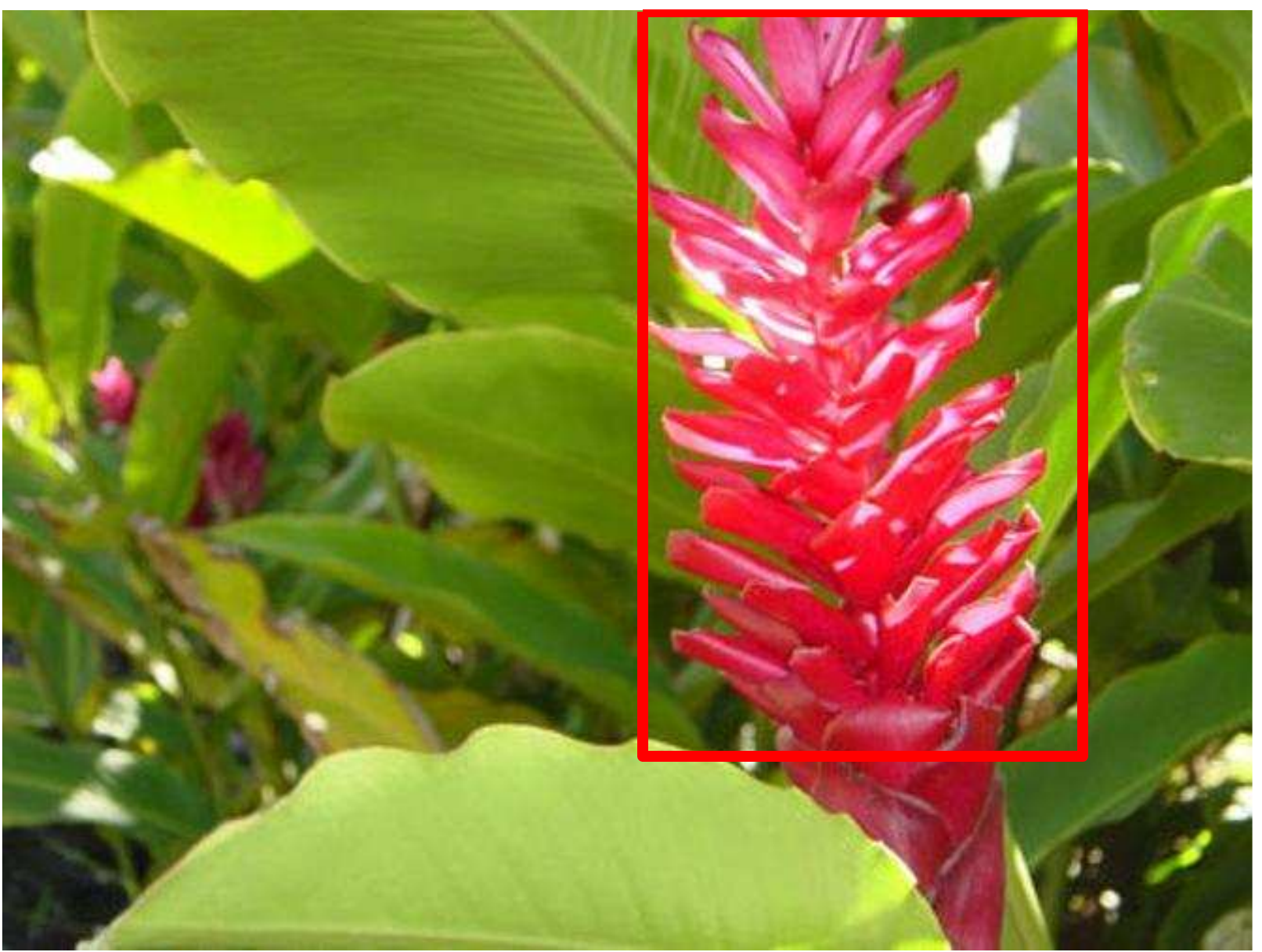}
\includegraphics[height=0.09\linewidth,width=0.09\linewidth]{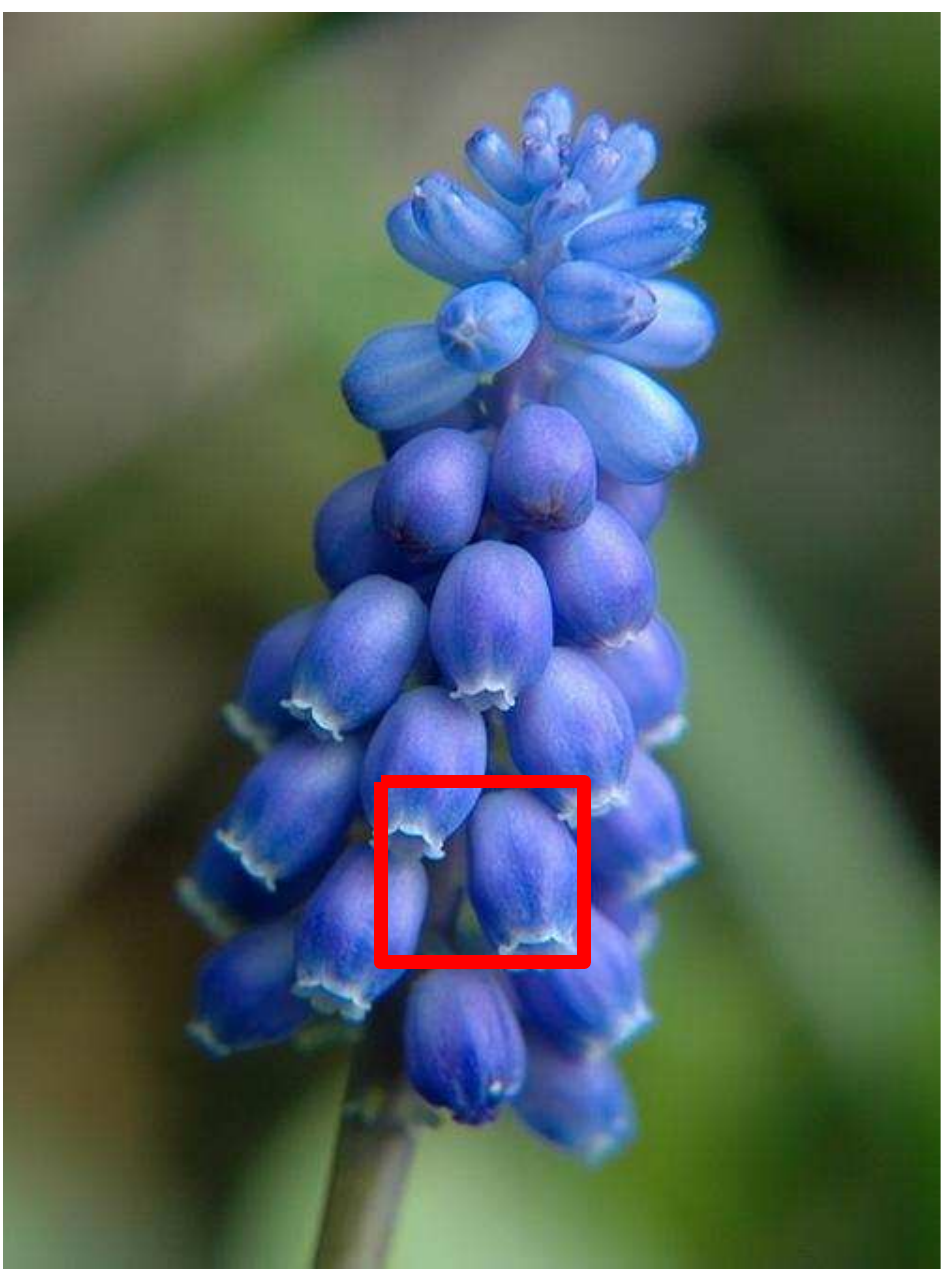}
\includegraphics[height=0.09\linewidth,width=0.09\linewidth]{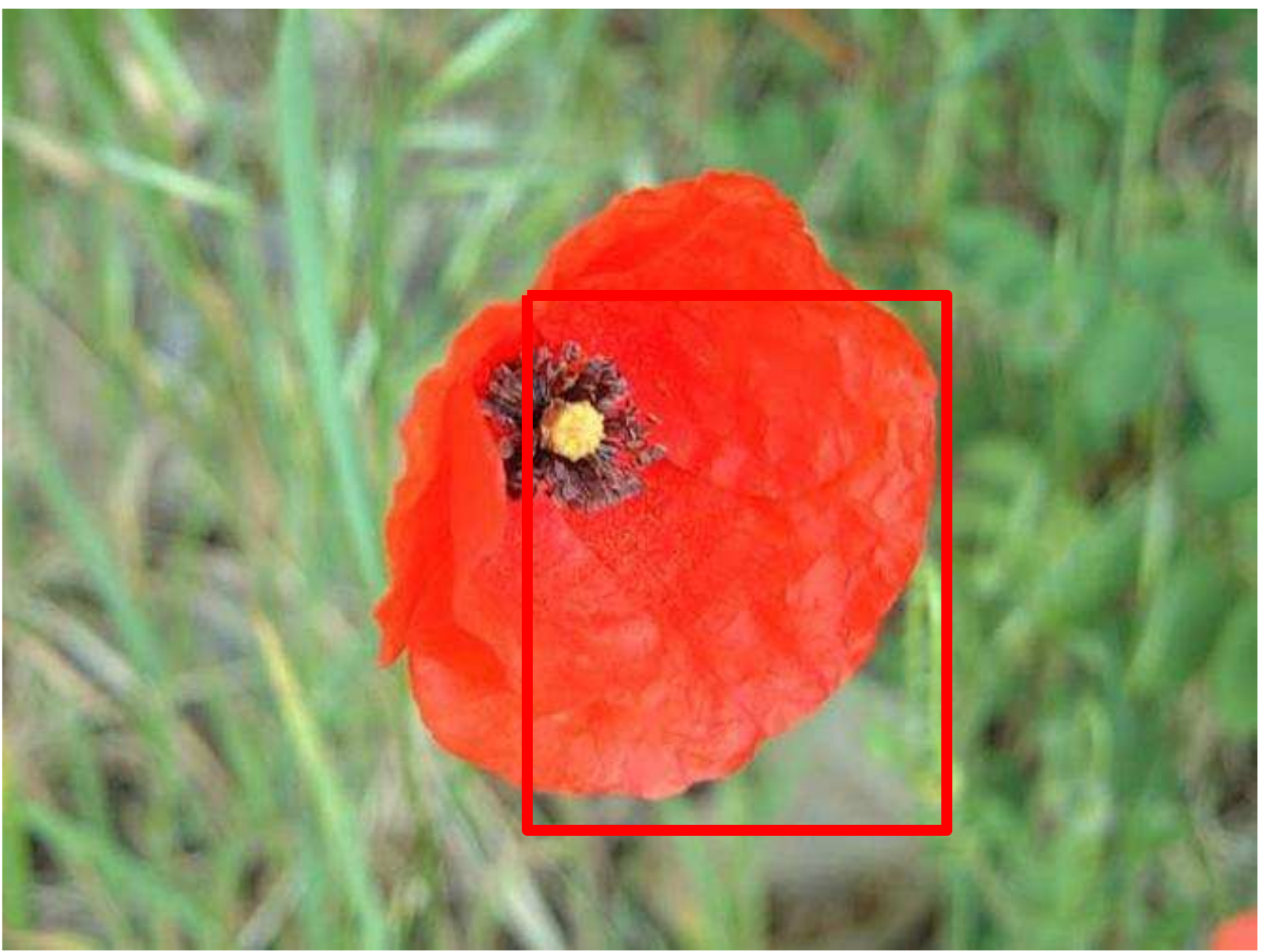}
\includegraphics[height=0.09\linewidth,width=0.09\linewidth]{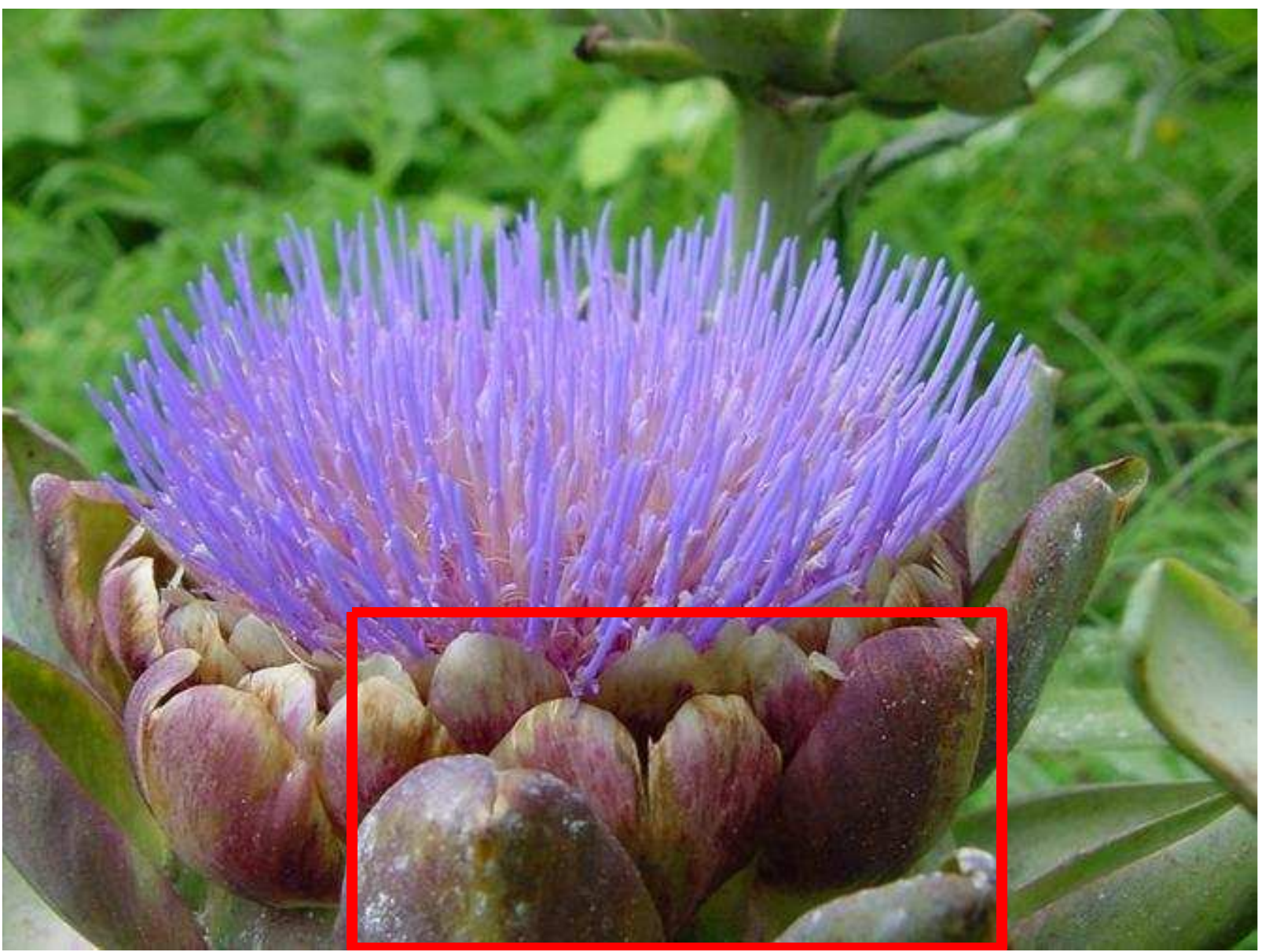}
\includegraphics[height=0.09\linewidth,width=0.09\linewidth]{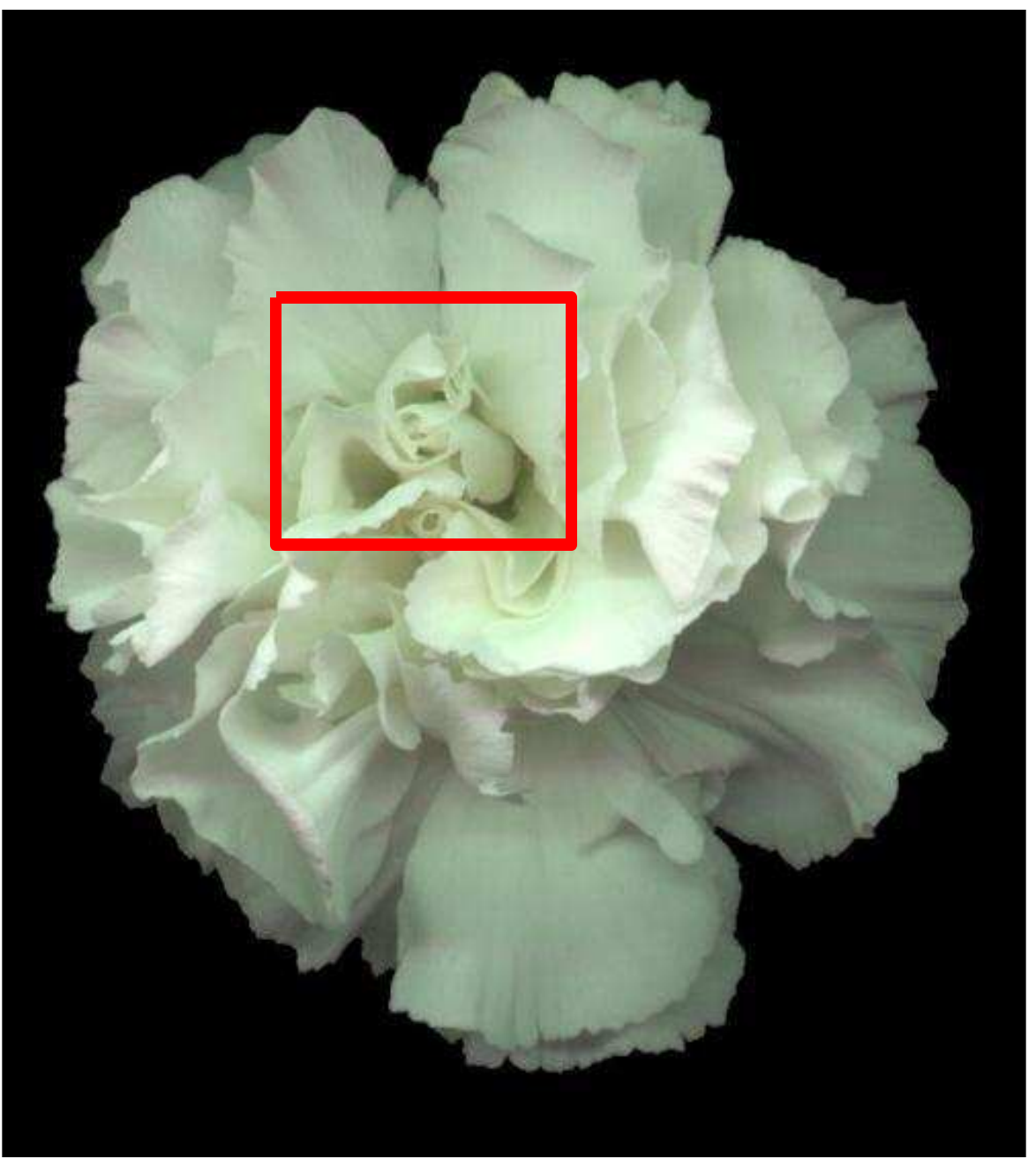}
\includegraphics[height=0.09\linewidth,width=0.09\linewidth]{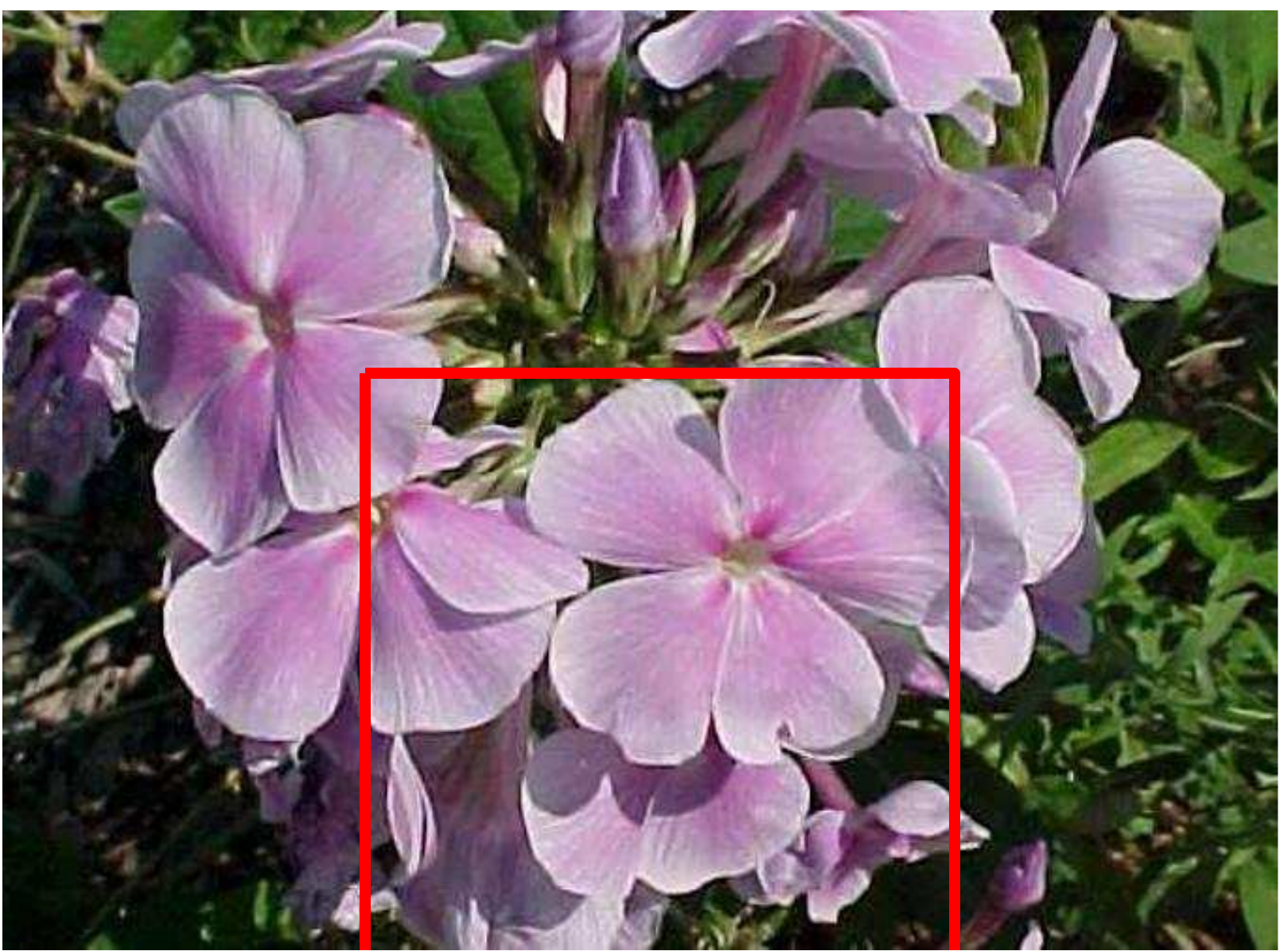}
\end{minipage}
}
\caption{Examples of discriminative part for each category shown in red box on PASCAL VOC 2007 dataset and Oxford flower dataset.
The discriminative part is the region which fires most on the classifier obtained by backtracking tricks.}
\label{fig:examples:part}
\vspace {-0.12in}
\end{figure*}

\section{Conclusions and Discussions}\label{sec:discussion}
To handle semantic gap of global CNN feature representation, this paper propose the deep attribute framework to alleviate the issue from three aspects. First, we introduce semantic region proposals as an intermedia to represent images.
Second, we show that aggregating soft-max output from region proposals with cross-region max-pooling yields best accuracy among all different CNN feature layers. The soft-max output (aka, deep attributes) is interpretable yet compact.
Third, we introduce context-aware region refining algorithm to pick out classification target related regions, and build context-aware classifiers.

To corroborate the effectiveness, we use the deep attribute as generic feature representation on various vision tasks.
Our empirical studies show that the proposed approach outperforms the competing methods with a large margin.
The reason for success is due to at least three factors. First, the region proposals are good alignment to target objects.
Second, cross-region max-pooling will suppress most noise regions, while keep most meaningful regions.
Third, the context region refining will keep only those those highly correlated regions while suppress most possible background noise regions.

Deep attributes have several good properties. Here we discussion some of them.
First, the deep attribute representation can be easily interpreted.
We can backtrack each attribute (per feature dimension) to region proposal which produce this attribute, as the aggregation is done by max-pooling.
We can also identify the most discriminant region for each image and category by sorting region score $S_k$ from Eq.\ref{eq1}.
Figure~\ref{fig:examples:part} shows more examples of the most discriminative region for each image in a red bounding box.

Second, deep attribute can be applied to a wide range of compute vision tasks. Especially, even the deep attribute is obtained from CNN model trained from ImageNet, it can be employed to those tasks which has fairly different object concepts from that of ImageNet.
For instance, in the experiment for fine grained flower recognition, the flower dataset has little category overlap with the ImageNet 1000 categories; as well as the Holiday and UKB dataset in image retrieval task.
As discussed above, Figure~\ref{fig:examples:part} illustrates the most discriminative region associated with each semantic category for the samples from both the PASCAL VOC and flower datasets. It is obvious that regions in PASCAL VOC samples reflect more semantic meanings, while this is not true for flower samples.
This is due to the fact that PASCAL VOC has more semantic concepts overlap with that of ImageNet than flowers dataset.
This phenomenon also inspires us to explore more things behind deep attributes. For instance, how many attributes are sufficient to support generic visual recognition tasks? Could we find a minimum supporting attribute set for visual recognition? Future works will make further exploration on it.

Third, in Figures~\ref{flowchart2:da}, we can see that the extracted deep attribute representation is somewhat sparse. We could make the representation even sparse by shrinking the deep attribute with a pre-defined threshold. Such a property is extremely useful for large-scale vision system, because a sparse representation can bring benefits in both storage and execution speed.

The proposed approach has several aspects to be improved.
First, there are large space for execution efficiency improvement. Currently, it runs about 5$\sim$10s per image on NVidia Titan X GPU.
Second, the accuracy and robustness could be further improved with fine-tuning on given dataset.
An interesting future direction is to design an unified framework to consider these two points together.

{\small
\bibliographystyle{ieee}
\bibliography{arxiv_ne_da_2015}

\begin{thebibliography}{10}\itemsep=-1pt

\bibitem{angelova2013efficient}
A.~Angelova and S.~Zhu.
\newblock Efficient object detection and segmentation for fine-grained
  recognition.
\newblock In {\em CVPR}, pages 811--818, 2013.

\bibitem{Arandjelovic12}
R.~Arandjelovi\'c and A.~Zisserman.
\newblock Three things everyone should know to improve object retrieval.
\newblock In {\em CVPR}, 2012.

\bibitem{arbelaez2009contours}
P.~Arbelaez, M.~Maire, C.~Fowlkes, and J.~Malik.
\newblock From contours to regions: An empirical evaluation.
\newblock In {\em CVPR}, pages 2294--2301, 2009.

\bibitem{mcg}
P.~Arbelaez, J.~Pont-Tuset, and et~al.
\newblock Multiscale combinatorial grouping.
\newblock In {\em CVPR}, 2014.

\bibitem{babenko2014neural}
A.~Babenko, A.~Slesarev, A.~Chigorin, and V.~Lempitsky.
\newblock Neural codes for image retrieval.
\newblock {\em arXiv:1404.1777}, 2014.

\bibitem{Chatfield14}
K.~Chatfield, K.~Simonyan, A.~Vedaldi, and A.~Zisserman.
\newblock Return of the devil in the details: Delving deep into convolutional
  nets.
\newblock In {\em BMVC}, 2014.

\bibitem{cheng2014bing}
M.-M. Cheng, Z.~Zhang, W.-Y. Lin, and P.~Torr.
\newblock Bing: Binarized normed gradients for objectness estimation at 300fps.
\newblock In {\em CVPR}, 2014.

\bibitem{deng2009imagenet}
J.~Deng, W.~Dong, R.~Socher, L.-J. Li, K.~Li, and L.~Fei-Fei.
\newblock Imagenet: A large-scale hierarchical image database.
\newblock In {\em CVPR}, pages 248--255, 2009.

\bibitem{DeCaf}
J.~Donahue, Y.~Jia, O.~Vinyals, and et~al.
\newblock Decaf: A deep convolutional activation feature for generic visual
  recognition.
\newblock In {\em ICML}, 2014.

\bibitem{everingham2010pascal}
M.~Everingham, L.~Van~Gool, C.~K. Williams, J.~Winn, and A.~Zisserman.
\newblock The pascal visual object classes (voc) challenge.
\newblock {\em IJCV}, 88(2):303--338, 2010.

\bibitem{farhadi2009describing}
A.~Farhadi, I.~Endres, D.~Hoiem, and D.~Forsyth.
\newblock Describing objects by their attributes.
\newblock In {\em CVPR}, pages 1778--1785, 2009.

\bibitem{girshick2013rich}
R.~Girshick, J.~Donahue, T.~Darrell, and J.~Malik.
\newblock Rich feature hierarchies for accurate object detection and semantic
  segmentation.
\newblock {\em arXiv:1311.2524}, 2013.

\bibitem{gong2014multi}
Y.~Gong, L.~Wang, R.~Guo, and S.~Lazebnik.
\newblock Multi-scale orderless pooling of deep convolutional activation
  features.
\newblock {\em arXiv:1403.1840}, 2014.

\bibitem{gu2009recognition}
C.~Gu, J.~J. Lim, P.~Arbelaez, and J.~Malik.
\newblock Recognition using regions.
\newblock In {\em CVPR}, pages 1030--1037, 2009.

\bibitem{harzallah2009combining}
H.~Harzallah, F.~Jurie, and C.~Schmid.
\newblock Combining efficient object localization and image classification.
\newblock In {\em CVPR}, pages 237--244, 2009.

\bibitem{hosang2014good}
J.~Hosang, R.~Benenson, and B.~Schiele.
\newblock How good are detection proposals, really?
\newblock {\em BMVC}, 2014.

\bibitem{jegou2008hamming}
H.~Jegou, M.~Douze, and C.~Schmid.
\newblock Hamming embedding and weak geometric consistency for large scale
  image search.
\newblock In {\em ECCV}, pages 304--317. 2008.

\bibitem{JDSP10}
H.~J\'egou, M.~Douze, C.~Schmid, and P.~P\'erez.
\newblock Aggregating local descriptors into a compact image representation.
\newblock In {\em CVPR}, pages 3304--3311, jun 2010.

\bibitem{jia2014caffe}
Y.~Jia, E.~Shelhamer, J.~Donahue, S.~Karayev, J.~Long, R.~Girshick,
  S.~Guadarrama, and T.~Darrell.
\newblock Caffe: Convolutional architecture for fast feature embedding.
\newblock {\em arXiv:1408.5093}, 2014.

\bibitem{krizhevsky2012imagenet}
A.~Krizhevsky, I.~Sutskever, and G.~E. Hinton.
\newblock Imagenet classification with deep convolutional neural networks.
\newblock In {\em NIPS}, pages 1097--1105, 2012.

\bibitem{kumar2009attribute}
N.~Kumar, A.~C. Berg, P.~N. Belhumeur, and S.~K. Nayar.
\newblock Attribute and simile classifiers for face verification.
\newblock In {\em CVPR}, pages 365--372, 2009.

\bibitem{lazebnik2006beyond}
S.~Lazebnik, C.~Schmid, and J.~Ponce.
\newblock Beyond bags of features: Spatial pyramid matching for recognizing
  natural scene categories.
\newblock In {\em CVPR}, volume~2, pages 2169--2178, 2006.

\bibitem{li2014attributes}
Z.~Li, E.~Gavves, T.~Mensink, and C.~G. Snoek.
\newblock Attributes make sense on segmented objects.
\newblock In {\em ECCV}, pages 350--365. 2014.

\bibitem{nister2006scalable}
D.~Nister and H.~Stewenius.
\newblock Scalable recognition with a vocabulary tree.
\newblock In {\em CVPR}, volume~2, pages 2161--2168, 2006.

\bibitem{razavian2014cnn}
A.~S. Razavian, H.~Azizpour, J.~Sullivan, and S.~Carlsson.
\newblock Cnn features off-the-shelf: an astounding baseline for recognition.
\newblock {\em arXiv:1403.6382}, 2014.

\bibitem{ILSVRCarxiv14}
O.~Russakovsky, J.~Deng, H.~Su, J.~Krause, S.~Satheesh, S.~Ma, Z.~Huang,
  A.~Karpathy, A.~Khosla, M.~Bernstein, A.~C. Berg, and L.~Fei-Fei.
\newblock {ImageNet Large Scale Visual Recognition Challenge}.
\newblock {\em CoRR}, abs/1409.0575, 2014.

\bibitem{sermanet-iclr-14}
P.~Sermanet, D.~Eigen, X.~Zhang, M.~Mathieu, R.~Fergus, and Y.~LeCun.
\newblock Overfeat: Integrated recognition, localization and detection using
  convolutional networks.
\newblock In {\em International Conference on Learning Representations}, 2014.

\bibitem{Simonyan14c}
K.~Simonyan and A.~Zisserman.
\newblock Very deep convolutional networks for large-scale image recognition.
\newblock {\em CoRR}, abs/1409.1556, 2014.

\bibitem{sivic2003video}
J.~Sivic and A.~Zisserman.
\newblock Video google: A text retrieval approach to object matching in videos.
\newblock In {\em CVPR}, pages 1470--1477. IEEE, 2003.

\bibitem{Smeulders:2000PAMI}
A.~W. Smeulders, M.~Worring, S.~Santini, A.~Gupta, and R.~Jain.
\newblock Content-based image retrieval at the end of the early years.
\newblock {\em Pattern Analysis and Machine Intelligence, IEEE Transactions
  on}, 22(12):1349--1380, 2000.

\bibitem{uijlings2013selective}
J.~R. Uijlings, K.~E. van~de Sande, T.~Gevers, and A.~W. Smeulders.
\newblock Selective search for object recognition.
\newblock {\em IJCV}, 104(2):154--171, 2013.

\bibitem{WeiXHNDZY14}
Y.~Wei, W.~Xia, J.~Huang, B.~Ni, J.~Dong, Y.~Zhao, and S.~Yan.
\newblock {CNN:} single-label to multi-label.
\newblock {\em CoRR}, abs/1406.5726, 2014.

\bibitem{yang2015can}
H.~Yang, J.~T. Zhou, Y.~Zhang, B.-B. Gao, J.~Wu, and J.~Cai.
\newblock Can partial strong labels boost multi-label object recognition?
\newblock {\em arXiv preprint arXiv:1504.05843}, 2015.

\bibitem{yang2009linear}
J.~Yang, K.~Yu, Y.~Gong, and T.~Huang.
\newblock Linear spatial pyramid matching using sparse coding for image
  classification.
\newblock In {\em CVPR}, pages 1794--1801, 2009.

\bibitem{edgebox}
C.~L. Zitnick and P.~Dollar.
\newblock Edge boxes: Locating object proposals from edges.
\newblock In {\em ECCV}, 2014.

\end{thebibliography}
}

\end{document}